\documentclass{article}

\usepackage[preprint]{corl_2024} 

\usepackage{amsmath,amsfonts,bm}

\def\eqref#1{equation~\ref{#1}}

\def\1{\bm{1}}

\DeclareMathAlphabet{\mathsfit}{\encodingdefault}{\sfdefault}{m}{sl}
\SetMathAlphabet{\mathsfit}{bold}{\encodingdefault}{\sfdefault}{bx}{n}

\usepackage{amsmath}
\usepackage{mathtools}
\usepackage{amsthm}
\usepackage{bbm}
\usepackage{amssymb}
\usepackage[capitalize,noabbrev]{cleveref}
\usepackage[ruled, vlined]{algorithm2e}
\usepackage{algpseudocode} 
\usepackage{float}
\usepackage{caption}
\usepackage{subcaption}
\usepackage{wrapfig}
\usepackage{array}
\usepackage{makecell}
\usepackage{bbm}
\usepackage{enumitem}
\usepackage{multirow}
\usepackage[export]{adjustbox}
\usepackage{mdframed}
\usepackage{tocloft}
\usepackage{colortbl}
\usepackage[frozencache=true,cachedir=.]{minted}
\usepackage{xcolor}
\definecolor{shadecolor}{RGB}{245,245,245} 
\definecolor{mylightblue}{rgb}{0.78,0.90, 0.96}
\definecolor{myskyblue}{rgb}{0.53, 0.81, 0.98}
\definecolor{mydarkblue}{rgb}{0,0.08,0.45}
\definecolor{myred}{rgb}{0.84,0.17,0.11}
\definecolor{myyellow}{rgb}{0.95,0.69,0.10}
\definecolor{mygreen}{rgb}{0.17,0.63,0.17}
\definecolor{myorange}{rgb}{0.8,0.40,0.05}
\definecolor{myblue}{rgb}{0.07,0.43,0.69}
\definecolor{mypurple}{rgb}{0.58,0.40,0.71}
\definecolor{mypink}{rgb}{0.85, 0.38, 0.69}
\definecolor{mygray}{rgb}{0.45,0.45, 0.45}

\theoremstyle{plain}

\theoremstyle{definition}

\theoremstyle{remark}

\newcommand{\ourshort}{\textbf{RoboGolf}}

\title{\textbf{RoboGolf}: Mastering Real-World Minigolf with a Reflective Multi-Modality Vision-Language Model}

\author{
    Hantao Zhou$^{\textbf{1*}}$\\
    \And
    Tianying Ji$^{\textbf{2*}}$\\
    \And
    Lukas Sommerhalder$^{\textbf{1}}$\\\
    \And
    Michael Görner$^{\textbf{1}}$\\\
    \And
    Norman Hendrich$^{\textbf{1}}$\\
    \And
    Jianwei Zhang$^{\textbf{1}}$\\
    \And
    Fuchun Sun$^{\textbf{2}}$\\
    \And
    Huazhe Xu$^{\textbf{2,3}}$\\
    \AND
    $^{1}$Universität Hamburg\\
    \And
    $^{2}$Tsinghua University\\
    \And
    $^{3}$ Shanghai Qi Zhi Institute
    \And
    *Equal contribution\\
    \AND  \url{https://jity16.github.io/RoboGolf/}
}

\begin{document}
\maketitle


\begin{abstract}
Minigolf is an exemplary real-world game for examining embodied intelligence, requiring challenging spatial and kinodynamic understanding to putt the ball. Additionally, reflective reasoning is required if the feasibility of a challenge is not ensured.
We introduce \ourshort, a VLM-based framework that combines
dual-camera perception with closed-loop action refinement,
augmented by a reflective equilibrium loop.
The core of both loops is powered by finetuned VLMs.
We analyze the capabilities of the framework in an offline inference setting, relying on an extensive set of recorded trajectories.
Exemplary demonstrations of the analyzed problem domain are available at \url{https://robogolfvlm.github.io/}.
\end{abstract}


\addtocontents{toc}{\protect\setcounter{tocdepth}{-1}}
\section{Introduction}\label{sec:introduction}

Recent advancements in foundation models~\citep{yang2023dawn, Math-LLM-ahn2024large, telecomLLM-maatouk2023large} offer a welcoming avenue for augmenting robot intelligence. Many of them are achieved by the integration of large language models~(LLMs) and vision-language models~(VLMs), leveraging their world knowledge~\citep{yu2023kola}, API use~\citep{qin2023toolllm}, information retrieval~\citep{nakano2021webgpt}, and more recently spatial understanding~\citep{chen2024spatialvlm} abilities.

Pioneering projects integrating these techniques in robotic manipulation~\citep{saycanahn2022can, codeaspoliciesliang2023code, hu2023look, huang2023voxposer} demonstrate impressive behavior.
Most of these projects are confined to tabletop manipulation tasks and often fail when applied to more complex kinematics-related tasks.
Recent efforts have focused on developing loop-closing paradigms, such as failure recovery~\citep{zhi2024closed} or human feedback for ambiguous queries~\citep{huang2022innermonologue}. However, without kinodynamic understanding, repeated feedback cycles in VLM inference will still fail to cope with the task dynamics,  failing to address the inherent temporal and physical complexities of continuous-time nonlinear systems~\citep{LI20141507}.  Equipping a VLM with kinodynamic understanding is crucial for extending its application in robotics beyond pick-and-place tasks.

\textcolor{mydarkblue}{\textit{To effectively handle complex, open-ended scenarios, what more do robots need beyond kinodynamic understanding?}}
We consider higher-level reflective reasoning mechanisms inspired by the philosophical ``reflective equilibrium'' method~\citep{daniels1979wide, cath2016reflective}.
The method proposes an iterative process of reflecting upon and refining one's beliefs by enhancing consistency between assumed general principles and concrete statements or observations.
Applied to robotic action inference, intelligent robots should not only understand tasks and scenarios and then design possible solutions, but also proactively suggest adjustments to make tasks feasible if they exceed the robot's observed capabilities. This ability to proactively propose practical modifications to tasks can avoid repeated trial and error under infeasible tasks.

\textit{\textcolor{mygreen}{Real-world minigolf}} stands out as an exemplary setting for examining the advanced intelligence of robots. Its highly variable course settings, with numerous endpoints and obstacles, create countless potential courses grounded in a well-defined action space.
Robots are challenged to identify feasible routes and generate appropriate club strikes. 
The spatial understanding required is considerable, as the court features various objects, many of which are compound, along with holes and passages underneath objects that may also be navigated. Each element presents unique challenges for the robot in reasoning and planning.
Moreover, the game allows for the challenge of designing feasible court layouts, challenging the systems intelligence beyond plain action inference.

We propose \ourshort, a framework that is comprised of dual-camera perception, and nested closed-loops for both refinement and reflective equilibrium.
The perception module captures detailed court information and tracks the high-speed motion of golf balls. The inner closed-loop refinement focuses on reasoning and planning feasible routes, generating and iteratively refining hitting actions, mainly powered by a kinodynamic reasoning VLM. A counterfactual VLM evaluates the feasibility and underpins the outer closed-loop reflective equilibrium, designed to actively modify erroneously designed courts or generate more evolved court variants.

A qualitative evaluation on a challenging offline dataset of robot strike attempts — in courts with complex obstacles, multiple holes, and even a billiard-style task bumping another ball in the course — showcases the system's performance in multi-trial inference through spatial and kinodynamic reasoning.
Additionally, the outer reflective loop can be utilized to effectively alter court layouts, turning infeasible tasks into achievable ones, or providing more evolved court variants.

\section{Related Works}\label{sec:relatedworks}

\noindent\textbf{Kinodynamic understanding.}\quad
Training systems to understand kinodynamics has been a longstanding challenge in applied machine learning. Various approaches have been explored, including trajectory prediction~\citep{westny2024diffusion-trajectory, billiards-fragkiadaki2015learning, ehrhardt2017learning}, and physical reasoning via training specific deep neural networks. These networks either build data-driven models of intuitive physics~\citep{image-boundary-bhattacharyya2016long, chang2016compositional} or use graph neural networks to capture interactions~\citep{battaglia2018relational, qi2020learning}. 
These methods often require large-scale ground-truth data, which limits their application in real-world scenarios without a simulator.
While learning forward dynamics models for each specific task has gained popularity in robotics~\citep{hafner2023mastering, hansen2023td, ji2022update, janner2019trust}, it still falls short in terms of generalizability and sample efficiency.
In contrast, exploring the kinodynamic capacity of VLMs through finetuning on small data regimes presents an appealing alternative. Our work demonstrates this by fine-tuning VLMs on a small scale of auto-labeled data.

\begin{figure}[t]
    \centering
    \includegraphics[width=0.95\linewidth]{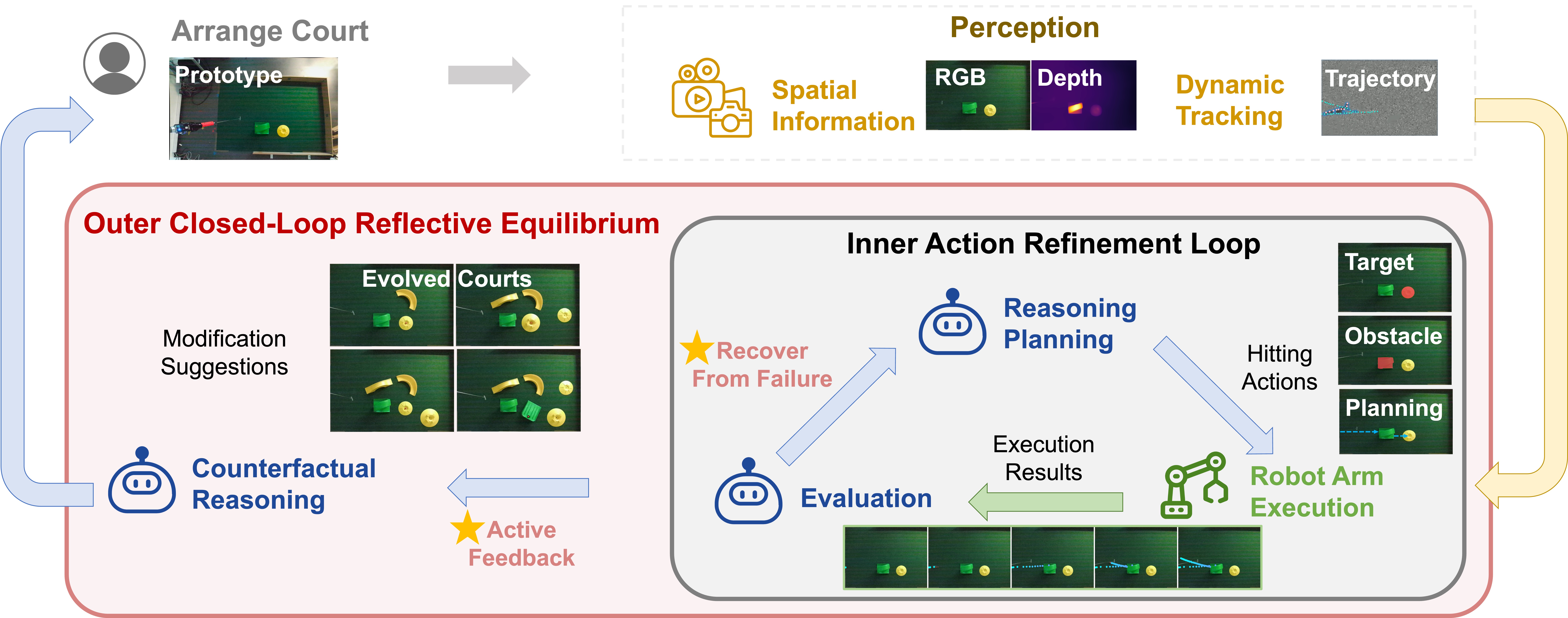} 
    \caption{\small \textbf{Conceptual Overview.} Our system integrates dual-camera scene perception, an inner action refinement loop predicting hit parameters, and an outer reasoning loop assessing course feasibility.
   An RGB-D camera captures the arranged spatial scene and ball trajectories are tracked using an event camera. The inner loop derives hitting parameters, adjusting them through evaluation of failed attempts. The outer reflective loop uses counterfactual reasoning to suggest course modifications in unsolvable scenarios.
    }
    \label{fig:simpleflow}
    \vspace{-5mm}
\end{figure}

\noindent\textbf{LLMs and VLMs for robotics.}\quad
LLMs~\citep{llm-robotics-zeng2023large, saycanahn2022can} and VLMs~\citep{huang2023voxposer, hu2023look, durante2024interactive-foundation, brohan2023rt-2, liu2024moka} have shown great potential in advancing robotic scenario understanding from textual or visual input and enhancing decision-making capabilities. 
Existing works can be broadly categorized as follows, with LLMs and VLMs primarily serving as:
1) High-level planners~\citep{saycanahn2022can, huang2022innermonologue, huang2022languagerobot, shah2023lm-nav-vlm, song2023llm}, leveraging the logical capabilities of LLMs/VLMs to focus on high-level reasoning for task or subgoal composition and the generalizability of VLMs across various tasks;
2) Code composers~\citep{liang2023codeaspolicies, yu2023language2rewards, huang2023voxposer}, which utilizes the programming techniques of LLMs/VLMs to act as low-level controllers for robots or provide optimization objective functions, for instance, generating execution code or reward functions;
3) Hybrid systems~\citep{liu2024moka, nasiriany2024pivot, huang2024copa}, combining both high-level planners and low-level code composers powered by LLMs/VLMs.
However, most previous works only allows robots to attempt tasks once without addressing failures. Recently, interest has emerged in devising VLMs with closed-loop for robotics tasks~\citep{zhi2024closed}, enabling robots to recover from failures. Despite the progress, these robots still passively try to complete tasks without questioning or modifying impractical scenarios caused by human errors.

\section{Method}\label{sec:method}
Our system, conceptualized in Figure~\ref{fig:simpleflow}, encompasses three major components embedded in a robotic paradigm: scene perception, action execution with closed-loop refinement, and an outer reflective reasoning loop to resolve impractical courses.
To capture fast golf ball trajectories and accurately perceive the geometric court layout, we use a dual-camera system.
The inner closed loop refines action parameters based on evaluated previous attempts.
It plans likely-feasible ball trajectories given the court geometry and utilizes a kinodynamic reasoning VLM to predict action parameters.
The outer loop uses a fine-tuned VLM for counterfactual reasoning to assess court feasibility from gathered trials and the observed court, devising court modifications as necessary.

To support short episodic rollouts, we restrict our scenario to a single club motion where each attempt to putt the ball starts with a fixed initial ball position. The action $a = \langle{}v,\theta\rangle$ parameterizes the club motion in the speed of the club head $v$ and its planar hitting angle $\theta$ when hitting the ball.
This intuitive choice of parameters provides the basis for parameter inference through the VLM.

\subsection{Perception}
Accurate perception of court and the ball movement is essential for effective action inference.
Figure~\ref{fig:perception} illustrates the workflow of the perception module.
More details are provided in Appendix~\ref{app:perception}.

\noindent\textbf{\textcolor{myblue}{Spatial information.}}\quad 
We employ a calibrated RGB-D camera to observe the course layout from a top-down perspective, ensuring an unobstructed, complete view, and segment the scene through the Segment Anything Model~(SAM)~\citep{kirillov2023segment}.
Combining depth information with segmented masks, we additionally incorporate knowledge of different surfaces.

\begin{figure}[t]
    \centering
    \includegraphics[width=0.96\linewidth]{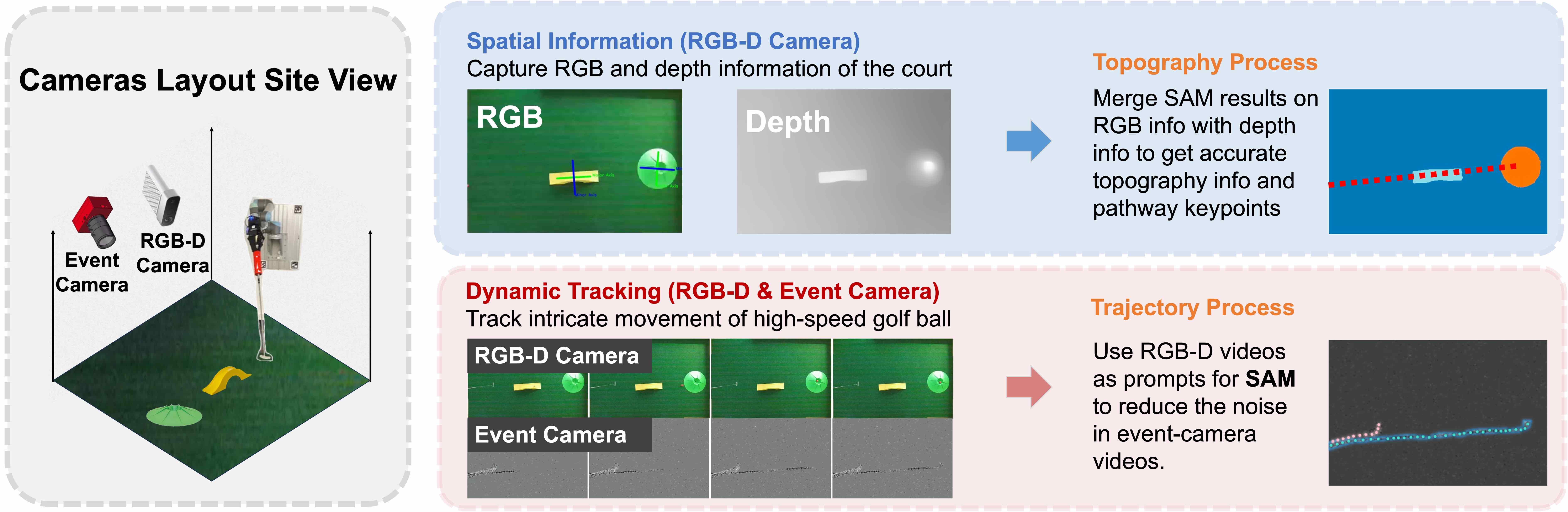} 
    \caption{\small \textbf{Perception module} with dual-camera setup.
    \textcolor{myblue}{$\bullet$} 
    \textit{\textcolor{myblue}{Spatial Information}}: RGB-D camera to capture details of the minigolf course. 
    \textcolor{myorange}{$\blacktriangle$}
    \textit{\textcolor{myorange}{Topography Process}}: use SAM to process RGB images, combine with depth information to generate accurate topography and court key points. The pathway is reconstructed by matching the positions of the segmented objects.
    \textcolor{myred}{$\bullet$} 
    \textit{\textcolor{myred}{Dynamic Tracking}}:  use RGB-D camera and event camera to record the movement of the high-speed golf ball. 
    \textcolor{myorange}{$\blacktriangle$} 
    \textit{\textcolor{myorange}{Trajectory Process}}: employ RGB-D videos as prompts for SAM to reduce noise in event-camera videos and to distinguish the trajectories of the golf club and golf balls.
    }
    \label{fig:perception}
    \vspace{-5mm}
\end{figure}

The segmentation mask, aligned with a pre-defined map, determines potential ball routes and critical points for each object, while the depth image provides additional information for route selection, feeding into the kinodynamic understanding of the VLM.
Finally, GPT-4V~\citep{openai2023gptv} selects and connects critical points to generate ball routes based on the depth and RGB images, assisted by visual lines.

\noindent\textbf{\textcolor{myred}{Dynamic tracking.}}\quad
However, the RGB-D camera alone is insufficient for kinodynamic perception, particularly in capturing the fast movement of a golf ball. While event cameras are optimized for fast motion, they are typically limited in their spatial resolution~\citep{rebecq2019event-highspeed, gallego2020event-survey}.
To address these challenges in kinodynamic tracking, we implemented a dual-camera setup to enhance overall perception quality.
For the fast-moving golf ball, we employed the high temporal-resolution event camera DAVIS 346~\citep{inivation2019davis346}.
We calibrate both the event camera and the RGB-D camera intrinsically as well as extrinsically to project observations into compatible spaces.

We employ RGB-D videos as prompts for SAM to filter noise in event-camera videos and distinguish the trajectories of the golf club and golf balls. Specifically, we record both raw and accumulated event information from the event camera and match this data to an RGB-D-based model of the golf court, using detected points from the RGB-D data as prompts for the event information.

\subsection{Inner Loop Action Refinement}

The inner action refinement loop involves the derivation of action parameters, their episode execution on the robot arm, as well as the evaluation of the observed ball trajectory.

\textcolor{myblue}{\textbf{Reasoning and planning}} involves mapping a route to the goal $\mathbf{g} \in \mathbb{R}^2$ and deriving hitting parameters to execute it. Specifically, the system reads the user's input $\mathcal{S}$ and uses LISA~\citep{lai2023lisa,yang2023improved} to identify the endpoint $\mathbf{g}$ from the court observation $\mathcal{OV}$: $\mathbf{g} = \mathcal{F}_{VLM}(\mathcal{S}; \mathcal{OV})$. Leveraging LISA, we thus identify a suitable endpoint in the visual field $\mathcal{V} \subseteq \mathbb{R}^2$.
All other recognized entities become obstacles $\mathcal{O} \subseteq \mathbb{R}^2$.
The model maps the obstacles onto pre-programmed route maps and modifies the image accordingly.
This image, now with assistive lines, is provided to GPT-4V~\citep{openai2023gptv} to connect key points and generate a plausible route $\mathbf{r}$ from start $\mathbf{s}$ to endpoint $\mathbf{g}$: $\mathbf{r} = \mathcal{F}_{VL}(\mathbf{s}, \mathbf{g}, \mathcal{O}; \mathcal{OV})$.
Finally, the model estimates the hitting speed $v$ and planar angle $\theta$ considering the obstacles and the distance to $\mathbf{g}$: $v = \mathcal{F}v(\mathbf{r}, \mathcal{O}, \mathbf{g}; \mathcal{OV})$ and $\theta = \mathcal{F}{\theta}(\mathbf{r}, \mathcal{O}, \mathbf{g}; \mathcal{OV})$. 
Given $v$ and $\theta$, our robot model including the golf club allows to readily compute a matching joint trajectory and execute it on the robot.
For more details on the trajectory derivation, please refer to Appendix~\ref{app:robotexecution}.

\textcolor{myblue}{\textbf{Failure analysis through evaluation}} involves assessing execution results and refining parameters accordingly. Specifically, the system evaluates outcomes using $\mathbf{e} = \mathcal{F}_e(\mathbf{r}, v, \theta, \mathbf{g}; \mathcal{OV})$, where $\mathbf{e}$ represents the evaluation result. If the ball misses the endpoint, it identifies failure reasons such as speed or angle deviations: $\mathcal{D}_v = \mathcal{F}_{VLM}(\mathbf{e}, \mathcal{S}; \mathcal{OV})$ and $\mathcal{D}_{\theta} = \mathcal{F}_{VLM}(\mathbf{e}, \mathcal{S}; \mathcal{OV})$.
Using these evaluations, the model refines hitting parameters with $\mathbf{a}' = \mathcal{F}_r(\mathbf{a}, \mathbf{e}, \mathcal{D}_v, \mathcal{D}_{\theta}; \mathcal{OV})$. This iterative process improves the system's precision and accuracy based on previous attempts.

\noindent\textbf{Spatial and kinodynamic understanding VLM.}\quad 
This step of the pipeline was motivated by spatialVLM~\citep{chen2024spatialvlm}, a successful attempt to enable the VLM to understand spatial relationships through fine-tuning. To enhance the specific reasoning capabilities of the VLM through fine-tuning, we collected 500 video episodes of the robot exploring actions in various courses. The action parameters were uniformly generated within thresholds around successful hits.
The important part of constructing the kinodynamic understanding of VLM is the construction of the VLM's training data. General-purpose VLMs often lack precise descriptive capabilities even in simple scenarios; while they can identify the number and locations of obstacles, they fail to provide accurate information and can produce hallucinations when faced with detailed questions. Therefore, it was infeasible to purely rely on GPT-4V to label the data.

Instead, we adopt several preprocessing steps including the semantic segmentation, obstacle pose detection models based on RGB-D information, and custom rules to segment the videos into different stages and generate related question-answer pairs. We employed a combined approach to automatically label the data step-by-step using GPT-4V and Claude3~\citep{anthropic2024claude}, applying rule-based methods.
Subsequently, we fine-tuned LLaVA~\citep{liu2023llava, liu2024llavanext, liu2023improvedllava, li2024llavanext-strong} with the auto-labeled data, which included question-answer pairs for each image. Our fine-tuned VLM successfully solves related spatial and kinodynamic questions concerning the minigolf game. For more details, please refer to Appendix~\ref{app:vlm-fintune}

\subsection{Outer closed-loop reflective equilibrium}
The outer reflective equilibrium loop acts as a higher-level loop beyond the episodic rollout. Specifically, counterfactual reasoning assesses the feasibility of the course based on the evaluation of previous rollouts. Upon identifying a court as infeasible, it actively suggests physical modifications. After the human operator rearranges the court as suggested, new episodes are rolled out.

In Figure~\ref{fig:outerclosed-loop}, to hit the golf ball into the yellow round target, it needs to traverse a roller coaster obstacle. The action reasoning determines that a very strong force is required. After applying the maximum force in the correct direction, the episode evaluation reveals that the ball cannot pass through the roller coaster and instead rolls backward. The counterfactual reasoning component identifies the endpoint as infeasible and actively suggests adding a redirecting surface, such as a yellow curved obstacle.

\noindent\textbf{Counterfactual reasoning VLM.}\quad
To enhance the counterfactual reasoning capability of a VLM, we collect images of 500 different impossible courts to fine-tune LLaVA. 
We label the data automatically using a rule-based method combined with GPT-4V.
The types of impossibilities are varied, including unreachable goals due to limitation of the robot arm, obstacles standing in the way, or slightly maladjusted obstacle angles.
It is straightforward to contract question-answer pairs for obstacles that obviously block the way based on a topographic analysis.
On the other hand, both --- arm limitations and maladjusted obstacles --- require detailed reasoning and the analysis of past episodes.
After several attempts the VLM can give correct evaluations.
For more details refer to Appendix~\ref{app:vlm-fintune}.

\begin{figure}[t]
    \centering
    \includegraphics[width=0.8\linewidth]{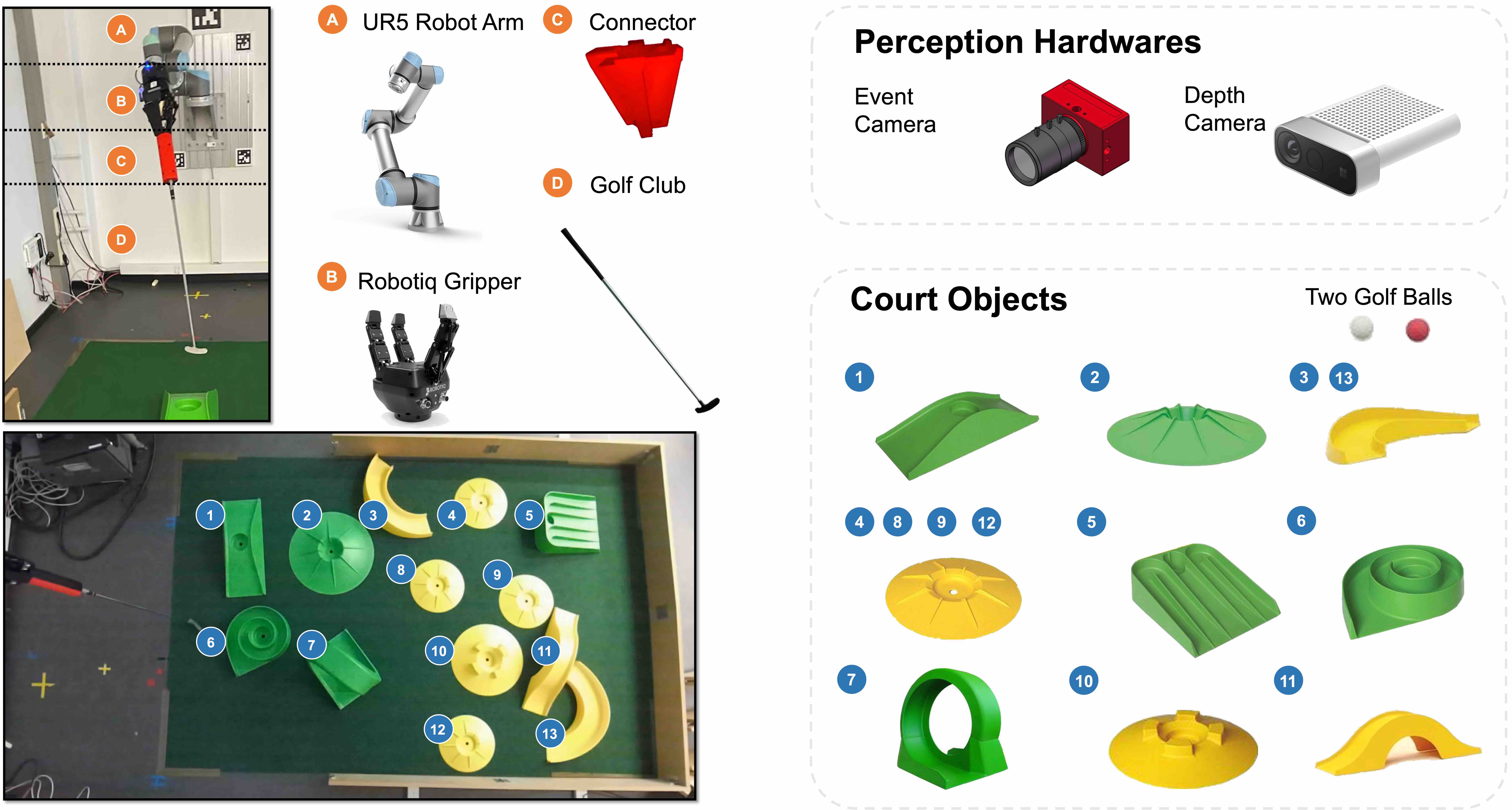}
    \caption{\small \textbf{Hardware and setup.}  The setup features a UR5 robotic arm with a 3D-printed connector, a Robotiq gripper, and a golf club. Perception hardware includes an event camera and a depth camera. The court includes 11 distinct object types, allowing varied minigolf configurations.}
    \label{fig:experimentalsetting}
    \vspace{-5mm}
\end{figure}

\section{Experiments}\label{sec:experiments}


In this study, we adopt a versatile hardware setup for robotic golf tasks, as depicted in Figure~\ref{fig:experimentalsetting}. The configuration enables the UR5 robotic arm, equipped with a 3D-printed connector and a Robotiq gripper, to wield a golf club, allowing it to execute strikes in all directions at practical forces. The setup includes 11 distinct types of court objects, enabling a highly variable arrangement of the playing field through numerous combinations of these elements. Additionally, our perception hardware comprises two devices: an event camera and a depth camera. For more details, please refer to Appendix C.

This study mainly targets the VLM inference framework. But even with comparably precise actuation through the robot arm, the episode dynamics upon hitting a ball remain a stochastic process. To circumvent this additional challenge for this study, we restrict our methods and qualitative evaluation to offline dataset analysis. We utilize the before-mentioned 500 independently recorded and annotated video sequences to approximate the parameter landscape of 25 court arrangements around a successful episode rollout. After each action prediction, we project the selected parameters onto the closest sample in the dataset and assume it as execution result.

\begin{figure}[t]
    \centering
    \begin{subfigure}[t]{1.0\textwidth}
        \centering
         \includegraphics[width=0.95\linewidth]{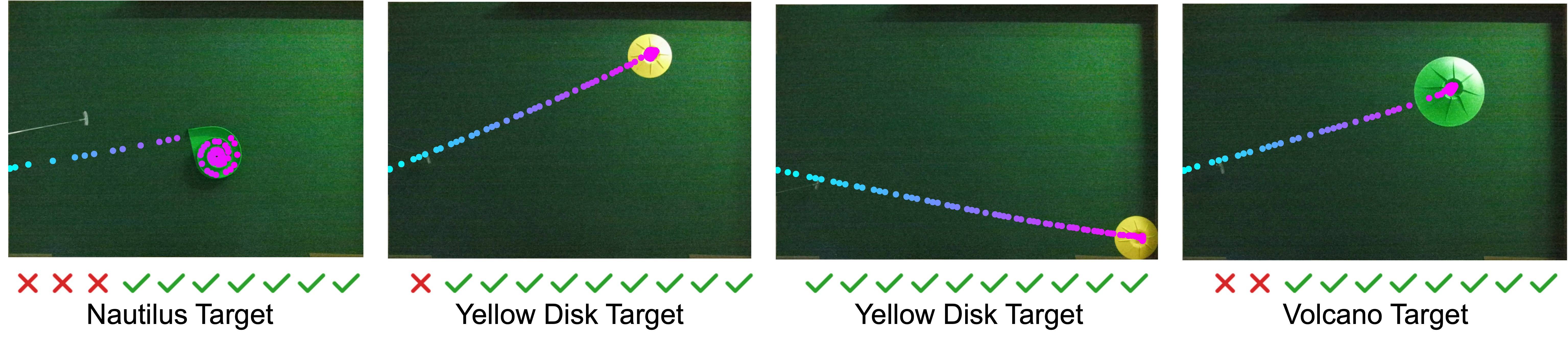}
           \vspace{-1mm}
         \caption{\small \textbf{Simple courts.}
         With one goal and no obstacles. Goal types are varied and moved across the court. 
         }
         \label{fig:easyscenarios}
         \vspace{1mm}
    \end{subfigure}
    \begin{subfigure}[t]{1.0\textwidth}
        \centering
        \includegraphics[width=0.95\linewidth]{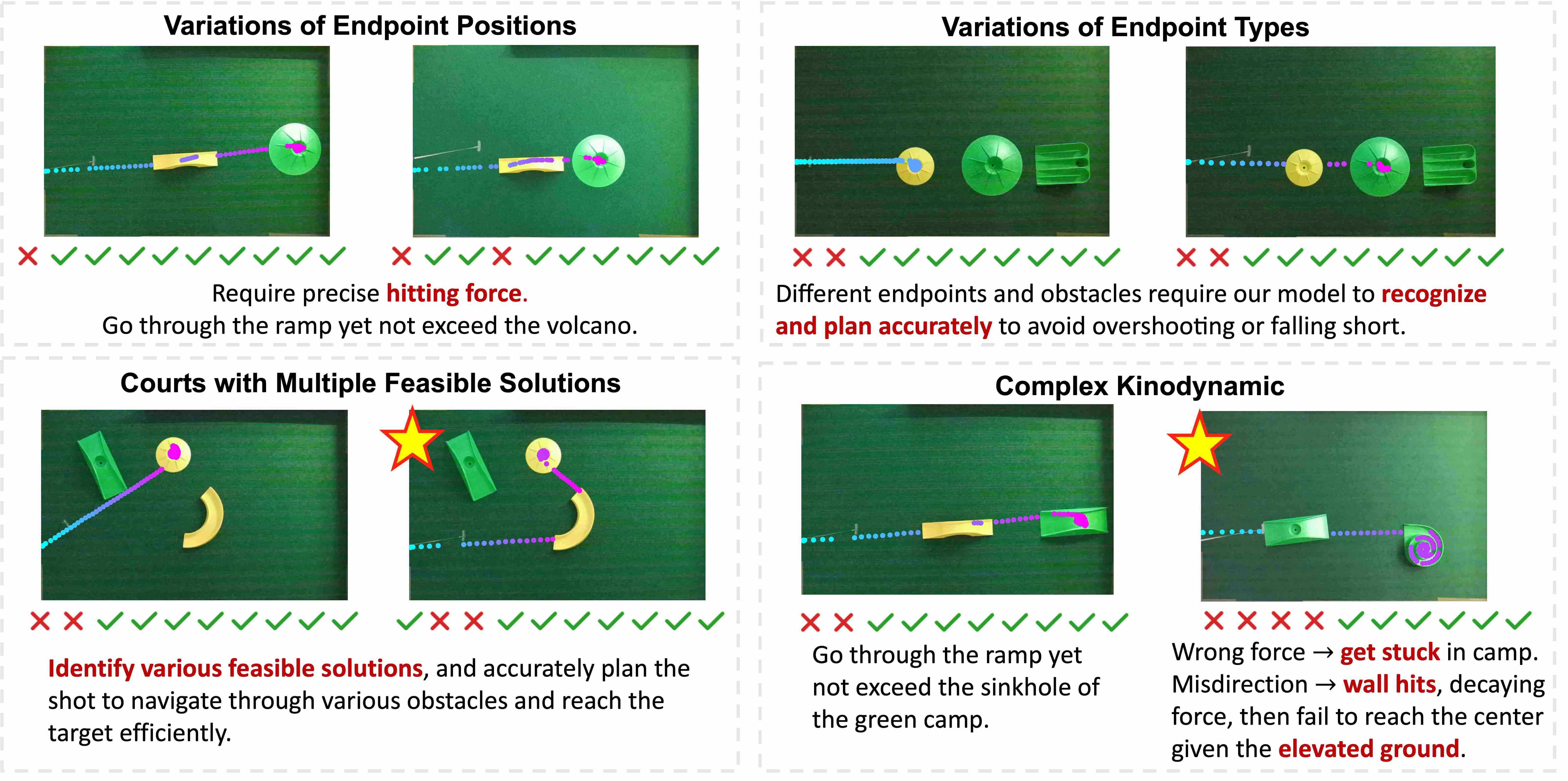}
         \vspace{-1mm}
        \caption{\small \textbf{Medium courts.} 
            Extra obstacles (e.g., ramp, bridge, or volcano) or confounding endpoints complicate topographic reasoning and endpoint identification.
        }
        \label{fig:medium-scenarios}
    \end{subfigure}
    \begin{subfigure}[t]{1.0\textwidth}
        \centering
        \includegraphics[width=0.98\linewidth]{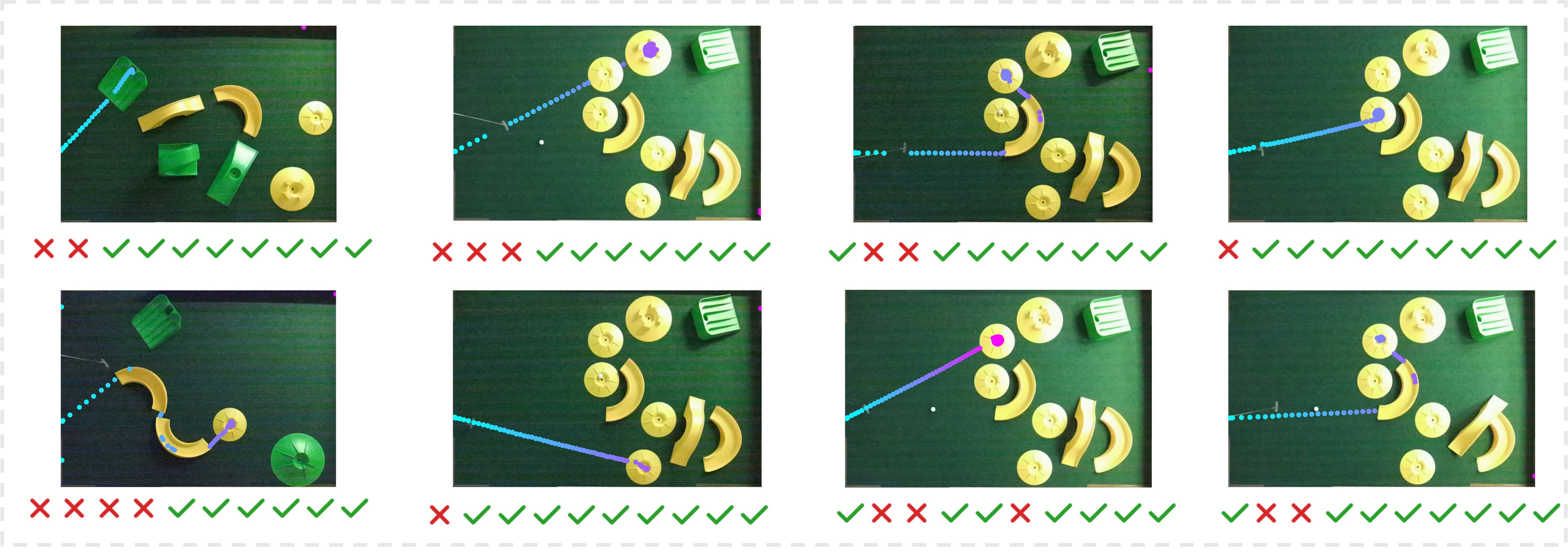}
        \caption{\small \textbf{Complex courts.}
            Multiple obstacles and endpoints create a complex court topology.
        }
        \label{fig:hard-scenario}
    \end{subfigure}
    \caption{\small \textbf{Dataset rollouts on courts of increasing complexity}. The dotted lines show the trajectory of successful hits, while the marks below indicate the success or failure to infer the correct action parameters in each inner loop iteration. 
    The system identifies the correct action parameters in each case within few trials.
    }
    \vspace{-4mm}
\end{figure}

\subsection{Action refinement on exemplary courts}\label{sec:exp-variouscourts}

\begin{figure}[t]
    \begin{minipage}[t]{1.0\textwidth}
        \centering
        \includegraphics[width=0.95\linewidth]{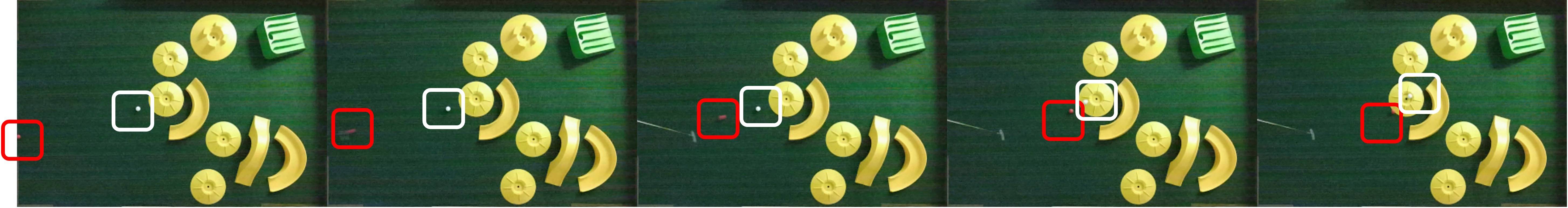}
        \vspace{-1mm}
        \caption{\small \textbf{Key frames of the successful demonstration in the billiard challenge.} The objective is to hit the red ball to bump the white ball into the yellow disk target. For visual clarity, the two balls are enclosed in boxes.}
        \vspace{-4mm}
        \label{fig:extreme-two-ball}
    \end{minipage}
\end{figure}

This task involves repeatedly hitting the ball from a fixed starting point, using a closed-loop framework to optimize parameters until the ball reaches the user-specified target. The objective is to achieve this with minimal attempts across various predefined courses. 
The main challenge lies in developing a spatial and kinodynamic understanding to plan a precise hitting route and generate the necessary hitting parameters. 
We craft multiple levels of challenges to test RoboGolf's abilities. 

\noindent\textbf{Simple courts.}\quad
For the simple courses, we use single endpoints and apply action refinement to improve the hitting parameters iteratively, directing the ball into the endpoint. Although generating a route to a single endpoint seems straightforward, initial attempts may fail due to real-world noise (e.g., accuracy, ground leveling, turf pattern, observation noise). Our RoboGolf system demonstrates robustness in handling these variations, as shown in Figure~\ref{fig:easyscenarios}.

\noindent\textbf{Medium courts.}\quad
The model recognizes different goals and obstacles and adjusts hitting parameters accordingly. It identifies feasible solutions and efficiently plans shots to navigate the obstacles. Additionally, it handles complex scenarios, such as circumventing wrong goals or misdirecting the ball into walls, by adjusting to elevated or uneven terrain.

\noindent\textbf{Complex courts.}\quad
The hard-level court features a sophisticated design with numerous redundant obstacles strategically placed to confound players and multiple feasible routes leading to various endpoints. Figure~\ref{fig:hard-scenario} highlights RoboGolf's reasoning capabilities in complex scenarios.

\noindent\textbf{Billiard challenge.}\quad 
We design a bilateral golf ball impact challenge, illustrated in Figure~\ref{fig:extreme-two-ball}. The goal is to hit the red ball from the starting point to bump the white ball, which is initially positioned at a different location, into the yellow round endpoint. This task significantly challenges the kinodynamic understanding of the VLM, and RoboGolf handles it after six attempts.

\subsection{Remediate the impossible via active modifications}\label{sec:exp-activemodify}
The key challenge in actively modifying the golf course to make impossible tasks feasible lies in first determining that a task is indeed impossible and then proposing reasonable modifications. The experiments demonstrate that RoboGolf could identify infeasible tasks and propose court modifications to achieve successful outcomes.

\begin{figure}[h]
    \begin{minipage}[t]{1.0\textwidth}
        \centering
            \includegraphics[width=0.95\textwidth]{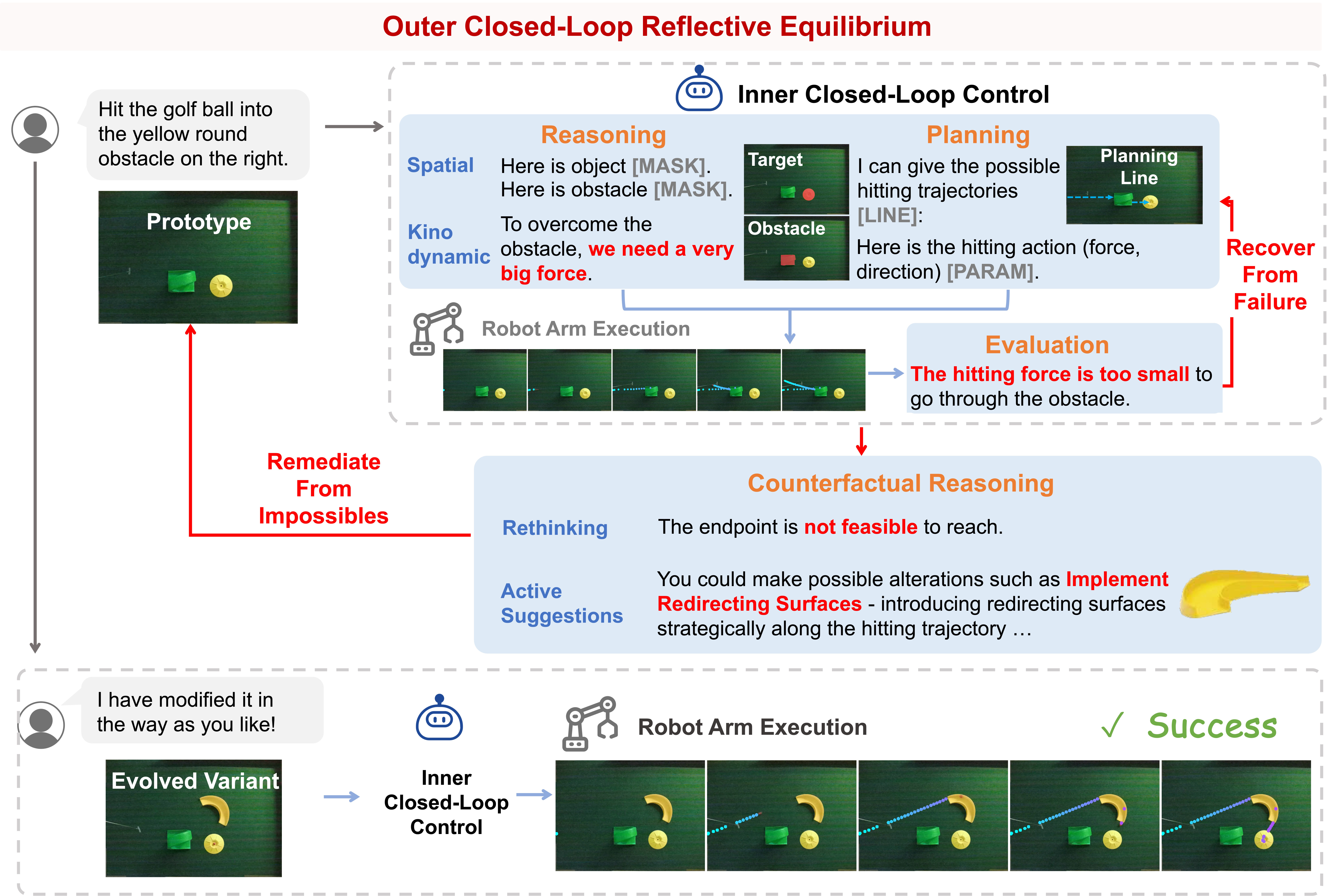}
            \caption{\small \textbf{reflective equilibrium loop.} 
            Observing that the ball cannot traverse the roller coaster on the only path to the endpoint, despite using the highest hitting speed, the reflective equilibrium system deems the task impossible and adjusts the court.
            }
        \label{fig:outerclosed-loop}
    \end{minipage}
    
    \vspace{6mm}
    \begin{minipage}[t]{1.0\textwidth}
        \centering
        \includegraphics[width=0.95\linewidth]{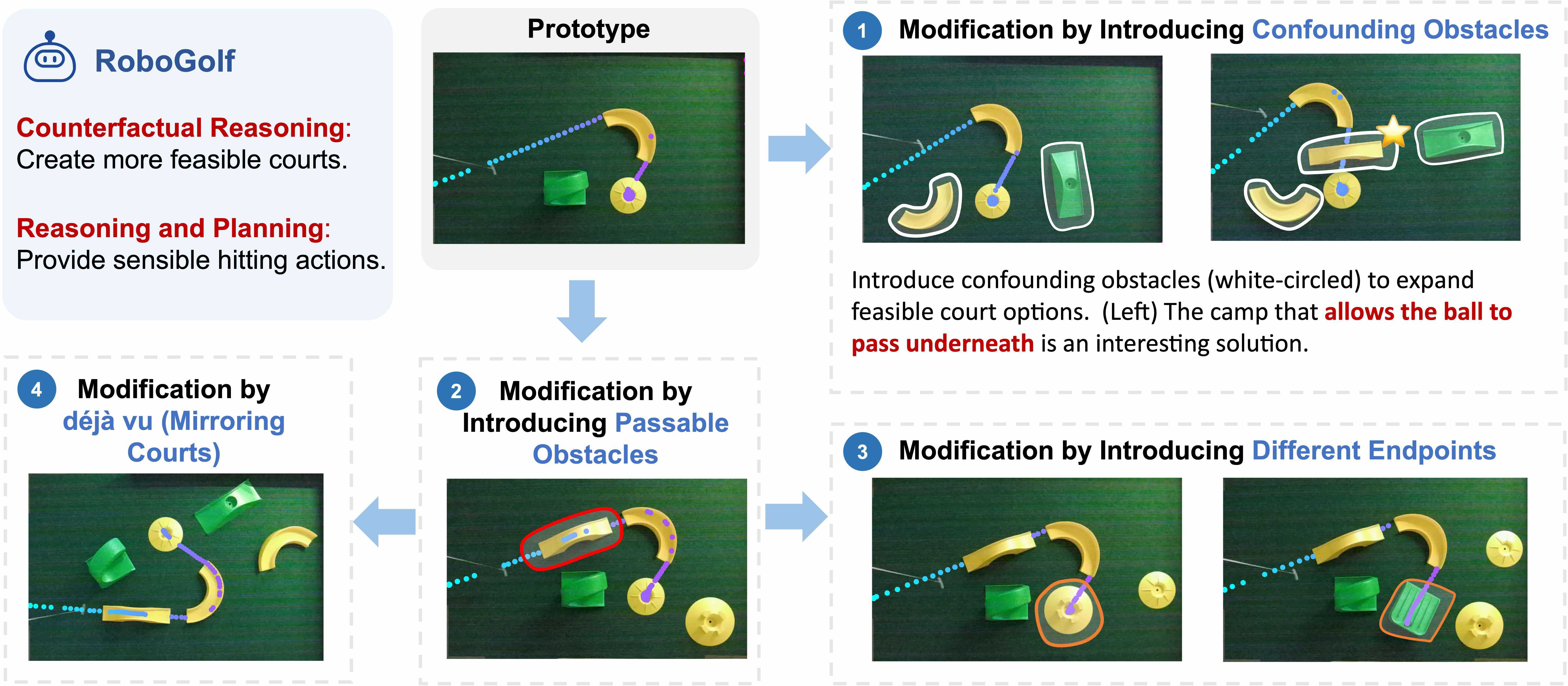}
        \caption{\small \textbf{Active golf court design.} Starting with the prototype, RoboGolf creates more feasible variants by adding confounding obstacles, incorporating passable obstacles, varying endpoints, mirroring courts, etc.}
        \label{fig:active-scenario-modification}
    \end{minipage}
\end{figure}

\noindent\textbf{Deem the impossibles.}\quad
Determining whether a minigolf challenge is impossible involves counterfactual reasoning. 
The VLM can directly reveal scenarios where no viable route exists from the starting point to the desired endpoint in the obstacle maps.
A more challenging form of impossible court emerges in dynamic limitations.
An example is shown in Figure~\ref{fig:outerclosed-loop}.
Initially, the system perceives the task as feasible, passing through the roller coaster obstacle.
However, despite attempting to strike with maximum velocity, the ball does not overcome the obstacle and rolls back.
Consequently, RoboGolf concludes the task to be impossible due to hardware constraints.

\noindent\textbf{Active modifications.}\quad
RoboGolf can actively propose modifications to the court to realize feasible variants of golf courts. To convert an impossible task into a feasible one, RoboGolf recommends incorporating a redirecting surface, as shown in Figure~\ref{fig:outerclosed-loop}. This alteration allows for an alternative ball trajectory, making the task achievable.

Furthermore, RoboGolf can enhance existing feasible setups to create evolved court variants, as depicted in Figures~\ref{fig:more-variants} and~\ref{fig:active-scenario-modification}. RoboGolf devises several evolved courts by adding confounding obstacles, varying endpoints, passable obstacles, and mirroring courts. These adjustments not only diversify the experimental conditions but also offer deeper insights into the system's adaptability and performance under different scenarios.

\section{Conclusion}\label{sec:conclusion}
Combining dual-camera perception, inner closed-loop refinement, and outer closed-loop reflective equilibrium, RoboGolf excels in the complex minigolf game. Notably, its advanced reflective equilibrium intelligence enables robots to understand intricate minigolf tasks and proactively propose feasible adjustments, enhancing their adaptability and performance in real-world scenarios. Extensive results show the potential of multi-modality VLMs in enhancing robot ability in the real world.

One direction that merits further investigation is to integrate an autonomous mobile robot to automate golf course modifications based on RoboGolf's suggestions. This would advance the system towards fully autonomous minigolf course management. Another intriguing direction may be a full online setting without fine-tune, taking considerations of online uncertainty in a more generalisable way.

\section{Acknowledgement}
This work was supported by the DFG Transregional Research Centre CML, TRR-169.
We would like to express our sincere gratitude to Tom Sanitz for the shared initial conceptualization of the robotic paradigm.
\addtocontents{toc}{\protect\setcounter{tocdepth}{3}}

\bibliography{Includes/IEEEabrv, CORL_2024-arxiv/ref}


\newpage
\appendix
{\Large \textbf{Appendices}}
{
\hypersetup{hidelinks}
\tableofcontents
}
\clearpage
\section{Illustrations of RoboGolf Execution}\label{sec:additionalablationstudies}
\subsection{Illustration of inner action refinement loop.}

\begin{figure}[H]
        \centering
        \includegraphics[width=1.0\textwidth]{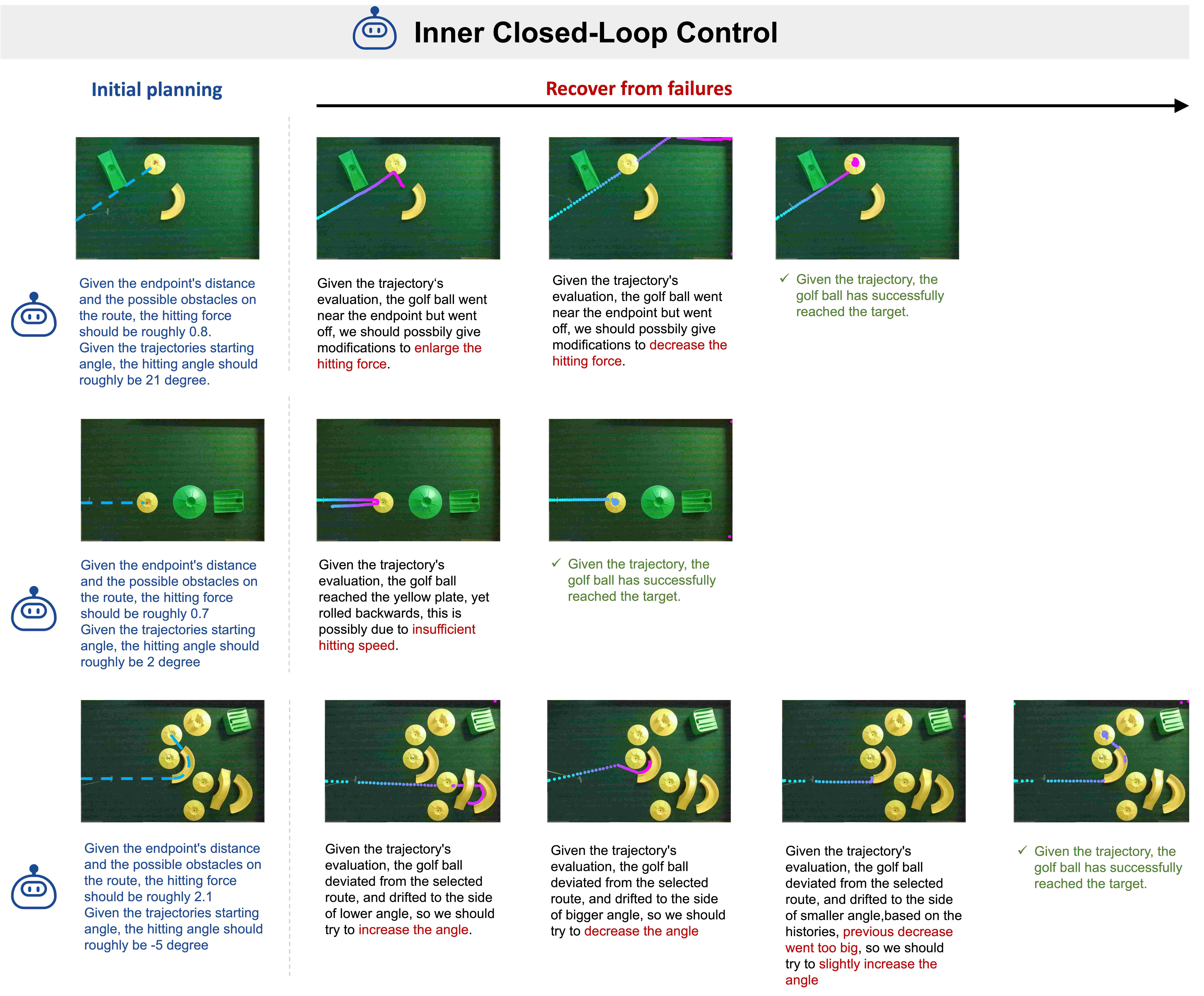}
        \caption{\textbf{Illustration of the inner action refinement loop of RoboGolf.}}
         \label{fig:inner-closed-loop-control}
\end{figure}

The inner action refinement loop involves reasoning and planning, executing hitting actions with a real-world robot arm, and evaluating episodes. As shown in Figure~\ref{fig:inner-closed-loop-control}, the process begins with initial planning, calculating the required hitting force and angle based on the endpoint's distance and obstacle positions. If the initial attempt fails, the system iteratively adjusts the hitting parameters—such as force and angle—based on trajectory evaluations until the golf ball successfully reaches the target. 

\subsection{Illustration of outer closed-loop reflective equilibrium.}
\begin{figure}[h]
    \centering
    \begin{subfigure}[t]{1.0\textwidth}
    \centering
        \includegraphics[width=1.0\textwidth]{Figures/mainsketchillustration-v3.jpg}
        \caption{\textbf{Illustration of outer closed-loop reflective equilibrium.}}
    \label{fig:appouterclosed-loop}
    \end{subfigure}
    \begin{subfigure}[t]{1.0\textwidth}
    \centering
        \includegraphics[width=1.0\textwidth]{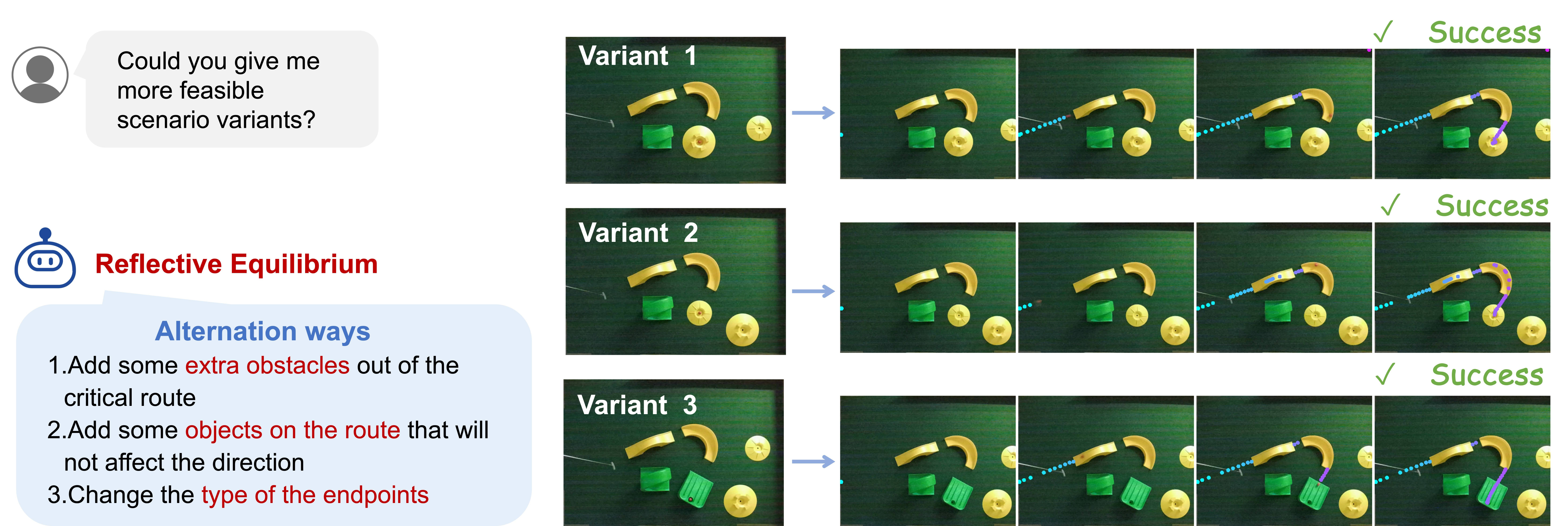}
        \caption{\textbf{Illustration of the active modification capability of \ourshort.}}
         \label{fig:more-variants}
    \end{subfigure}
    \caption{\textbf{Illustration of outer closed-loop reflective equilibrium and the counterfactual reasoning ability of RoboGolf.}}
    \label{fig:mainsketch}
\end{figure}

In Figure~\ref{fig:appouterclosed-loop}, RoboGolf attempts to hit the golf ball into a yellow round target, which requires traversing a roller coaster obstacle. Initially, the inner refinement loop requires a very strong force. Despite applying the maximum force in the correct direction, the execution results reveal that the ball cannot pass through the roller coaster and instead rolls backward. RoboGolf’s counterfactual reasoning identifies the endpoint as infeasible and actively suggests adding a redirecting surface, such as a yellow curved obstacle.

RoboGolf can enhance existing feasible setups to create evolved court variants. As depicted in Figure~\ref{fig:more-variants}, RoboGolf devises several evolved courts by adding confounding obstacles, varying endpoints, and passable obstacles. These adjustments not only diversify the experimental conditions but also offer deeper insights into the system’s adaptability and performance under different scenarios.

\clearpage
\section{Implementation Details}\label{sec:implementationdetails}

\subsection{Perception}\label{app:perception}

\subsubsection{Event camera}

\paragraph{Calibration of the Event Camera}

For intrinsic parameters, the distortion of the event camera is relatively significant, we use the calibration method provided by the DV GUI~\citep{inivation2024dvgui}. We use a metal-based checkerboard to calibrate the intrinsic parameters and save the intrinsic calibration parameters for later usage.

For the extrinsic parameters, we selected the event camera model with RGB+event information to align the event information with the RGB-D camera.

\paragraph{Recordings}

\begin{figure}[H]
\centering
    \includegraphics[width=0.9\linewidth]{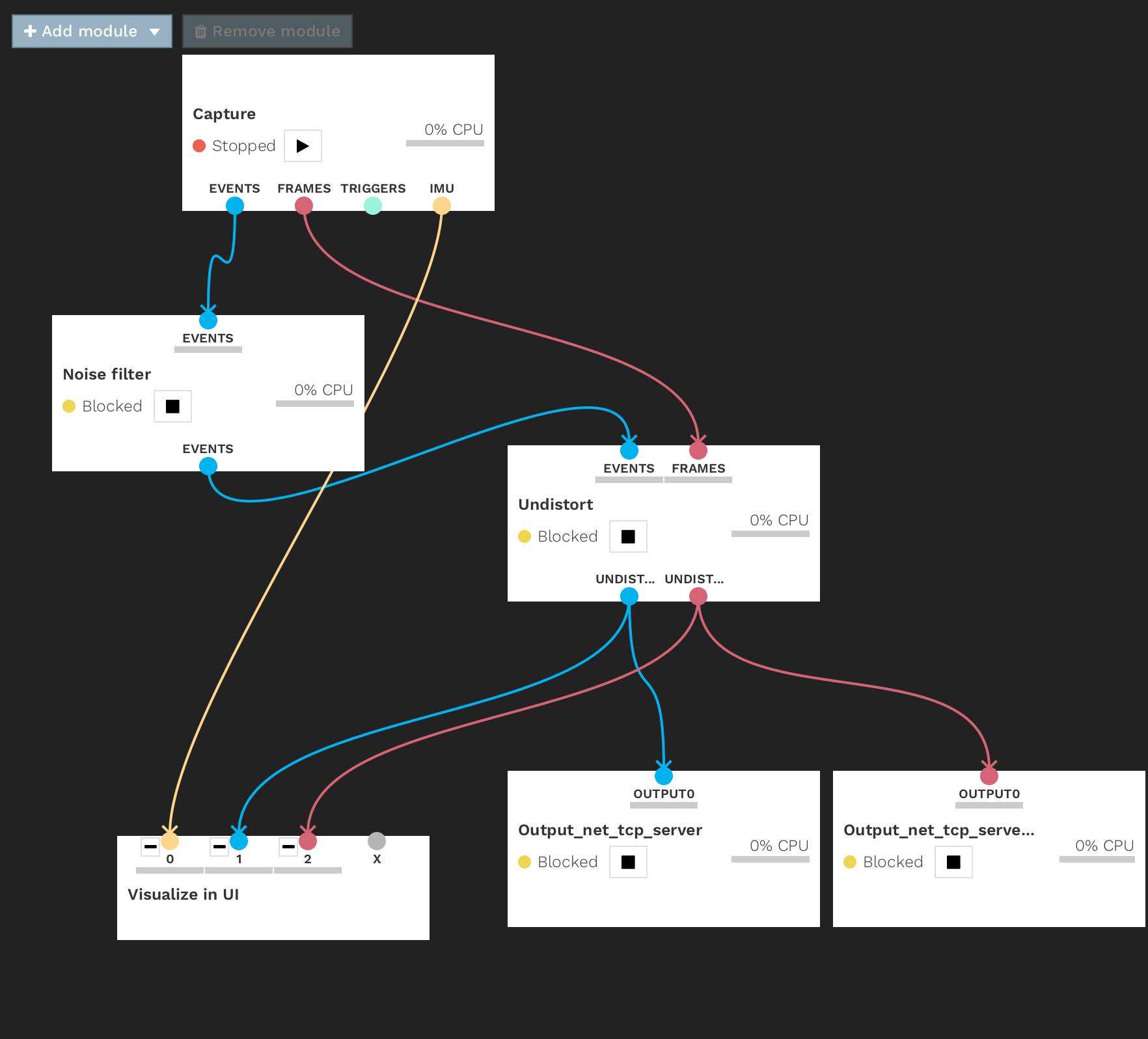}
    \caption{\textbf{Recorder flow for event and RGB information.}}
    \label{fig:recorder-flow}
\end{figure}

We use the DV GUI application to record information. As demonstrated in Figure~\ref{fig:recorder-flow} the event information first undergoes noise filtering and then proceeds to the undistortion module, utilizing the parameter file from the calibration. Next, we use the streaming module to stream the event information and RGB data. Subsequently, we employ the DV-processing~\citep{inivation2022dvprocessing} Python code to write a server that records both the event and RGB information.

\subsubsection{RGB-D camera}

For the selection of the RGB camera, we adopted the camera Moto from Microsoft Azure Kinnect, it is a structure light-based decks camera with RGB information.

Before the actual application of the RGB camera, we first calibrate the intrinsic parameters to avoid this distortion and then we also calibrate the relationship between the depth information and the RGB information in order to match them to a well-aligned cartesian space.

The recording process is conducted using the ROS recording system that is due to the nature of the data generated by the RGBD camera. It cannot process and store all the data in real-time causing certain frames to be lost therefore we have to adopt the ROS system to record the road, and after the recording, we can then process the data to store them as videos.
\subsection{Technical details on the inner and outer closed-loop}\label{app:vlm-fintune}
\subsubsection{VLM finetune}

\paragraph{Auto label data}

The preparation of data is crucial for the training of vision language models, serving two distinct purposes that necessitate a separate labeling mechanism, albeit with a shared processing mechanism.

The auto-labeling process comprises several stages:

\begin{enumerate}
\item \textbf{Stage 1:} We employ rule-based traditional methods to detect the rotation and direction of obstacles. Specifically, we adopt Structure-Aware Masks (SAM) to generate masks for objects on the ground and subsequently utilize rule-based methods to match with the original model, thereby obtaining a rotational angle.

\item \textbf{Stage 2:} We track the ball across the court and segment and extract critical frames where the ball interacts with obstacles on its path.

\item \textbf{Stage 3:} Finally, we generate data with questions and answers based on the previous information. This stage marks the separation of the dataset dedication for different purposes. 
\end{enumerate}

\paragraph{Kinodynamic understanding data.}

To improve the kinodynamic understanding aspect of the VLM, we evaluated the current comprehension abilities of leading models. Given that large multimodality models and vision-language models tend to generate unsatisfactory answers and potential hallucinations when asked vague questions, we  divide the labor required for this fine-tuning process.

We prepared several types of questions to label the data for training the visual language model. The first type of question addresses the number of obstacles on the court and their positional relationships.
The second type of question is the evaluation of whether the ball hits the goal successfully or not.
The third type of question, which is similar to the second, is concerned with the reasons behind the ball's deviation from its endpoint. This requires us to furnish the auto-labeling vision language models with extensive information about the golf ball's trajectory and speed circumstances.

\paragraph{Reflective equilibrium.}

To prepare the data for finetuning the visual language models to make it capable of providing reflective equilibrium by counterfactual reasoning, we have to label the data with questions focused more on the relationship between the obstacles. Collected scenarios are combined with an evaluation of the hitting histories constituting pairs of providing hitting histories and questions and a judgment of impossible.

\clearpage
\subsubsection{Instructions}

\paragraph{Endpoint intention detection.}

Firstly the framework provides a top-down RGB image of the court with the human's instruction describing which endpoint to select then firstly the prompts and the RGB image are provided to the vision language model LISA, to recognize the masks and the spatial relationship with the others. We also use the finetuned VLM to provide some basic comments on the situation.
\bigskip
\begin{mdframed}[linewidth=1pt]

    \bigskip
    \begin{center}
            \textbf{\textcolor{myblue}{\large Detailed Prompt Example for \\ Vision-Based Miniature Golf Endpoint Detection}}
    \end{center}
\medskip
\textcolor{myred}{\textbf{Purpose:}} This prompt is designed to enable a vision language model to accurately identify and segment the desired endpoint in a miniature golf setup using a provided top-down RGB image and user-defined language instructions. This process supports the robotic arm in executing a precise, single-shot attempt.

\textbf{\textcolor{myred}{Input Requests}}
\begin{itemize}
    \item \textbf{Image of the Miniature Golf Court:}
    \begin{itemize}
        \item \textit{Description:} A top-down RGB image of the 2-meter by 3-meter golf court featuring various plastic obstacles and designated endpoints.
    \end{itemize}

    \item \textbf{User Instruction:}
    \begin{itemize}
        \item \textit{Description:} Textual instructions specifying the target endpoint for the golf ball. Example: "Identify the endpoint near the large blue obstacle on the right side."
    \end{itemize}
\end{itemize}

\textbf{\textcolor{myred}{Output Response}}
\begin{itemize}
    \item \textbf{Format:} Image with segmentation mask overlay on the desired endpoint: 
    
    \textit{\textcolor{mygray}{
Example of expected output: 
A modified version of the input image highlights the segmented endpoint with a distinct color overlay, distinguishing it from other elements in the scene.}}
\end{itemize}

\textbf{\textcolor{myred}{Rule Definitions and Reasoning Steps}}
\begin{enumerate}
    \item \textbf{Image Preprocessing:}
    \begin{itemize}
        \item Perform necessary adjustments to enhance image quality for better feature recognition (e.g., brightness, contrast).
    \end{itemize}

    \item \textbf{Endpoint Detection:}
    \begin{itemize}
        \item Apply deep learning techniques to interpret and segment the area described in the user's instructions.
        \item Utilize natural language processing to correlate textual descriptions with visual features in the image.
    \end{itemize}

    \item \textbf{Segmentation and Annotation:}
    \begin{itemize}
        \item Use semantic segmentation algorithms to isolate and highlight the designated endpoint.
        \item Annotate the segmented area clearly to guide the robotic arm in targeting.
    \end{itemize}

    \item \textbf{Output Validation:}
    \begin{itemize}
        \item Ensure the accuracy of the segmentation by cross-referencing the visual output with the user's original instructions.
    \end{itemize}
\end{enumerate}

\end{mdframed}

\clearpage
\paragraph{Route planner design.}

\begin{figure}[t]
\centering
\begin{minipage}[t]{0.49\textwidth}
    \vspace{0pt}
        \centering
    \includegraphics[width=0.9\linewidth]{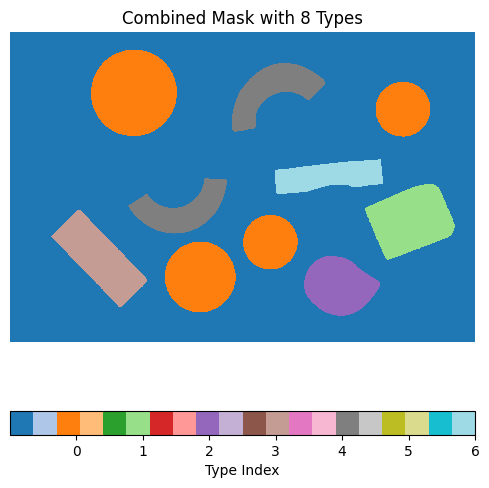}
    \caption{The demonstration of all masks for different types of obstacles and endpoints}
    \label{fig:combined-mask}
\end{minipage}
\hfill
\begin{minipage}[t]{0.49\textwidth}
\vspace{0pt}
    \centering
    \includegraphics[width=0.9\linewidth]{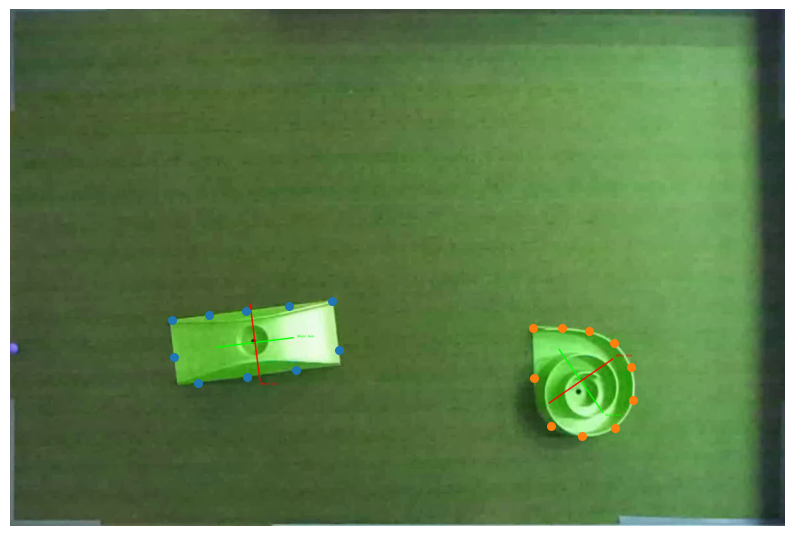}
    \caption{\textbf{Visual prompts for route planning}. 
    Visual prompts are obtained based on the contours of the segmentation masks.
    The visual prompts are provided combined with other visual information for GPT-4V to generate a reasonable route plan.}
    \label{fig:key-point}
\end{minipage}
\end{figure}

In this study, we address the challenge of developing a robotic golf ball playing assistant that can intelligently plan a path across a golf course by avoiding obstacles and optimizing trajectory. To this end, we propose a structured approach for path planning, which includes both high-level strategic planning and detailed motion planning. Our system takes as input a detailed description of the golf course and a top-down annotated image showing key obstacles and target locations. Visual prompts support the information for the obstacles and target locations, we first consider different masks for different obstacles in Figure~\ref{fig:combined-mask}, and then we apply segGPT~\citep{wang2023seggpt} for the zero-shot classification of the current court. Based on the contours of the obstacle in the court, we label the image with visual prompt points in Figure~\ref{fig:key-point}.

Based on this input, the system outputs a plan in the form of a JSON dictionary, which details waypoints, motion directions, and specific actions at each stage of the route. Key components of our solution include 1) the identification of critical key points where the golf ball should pass or change direction, 2) the determination of optimal waypoints that guide the ball along the safest and most efficient trajectory, and 3) the visualization of this route on the annotated map to verify and adjust the plan as needed. Each step is designed to ensure that the robot can effectively parse natural language instructions and translate them into a practical and executable motion plan, thus enhancing the robot's capabilities in outdoor sports environments.
\bigskip
\begin{mdframed}[linewidth=1pt]
\begin{minipage}{1.0\textwidth}
    \bigskip
    \begin{center}
            \textbf{\textcolor{myblue}{\large Detailed Prompt Example for \\ Robotic Golf Ball Playing Assistant}}
    \end{center}
    \medskip
    \textcolor{myred}{\textbf{Purpose:}} This prompt is designed to guide the robotic assistant in selecting an optimal route for a golf ball on a simulated golf course, given a top-down view image of the court with annotated assist lines and obstacle detections, to achieve a successful shot in a single attempt.
    
    \bigskip
    \textbf{\textcolor{myred}{Input Requests}}
    \begin{itemize}
        \item \textbf{Task Information:}
        \begin{itemize}
            \item \textit{Instruction:} \textbf{``Select the best route for the golf ball to reach the hole, avoiding all obstacles and minimizing the stroke count, given the provided top-down view image of the court.''}
            \item \textit{Desired Endpoint:} The hole location on the golf course.
        \end{itemize}
    
        \item \textbf{Image of the Court:}
        \begin{itemize}
            \item \textit{Description:} Annotated top-down view image of the golf course, including:
            \begin{itemize}
                \item \textit{Assist Lines:} Possible pathways for each obstacle detected.
                \item \textit{Obstacle Detections:} Locations of sand bunkers, water hazards, and other obstacles.
                \item \textit{Desired Points:} Annotated points of interest, such as the hole location.
            \end{itemize}
        \end{itemize}
    \end{itemize}
\end{minipage}
\begin{minipage}{\textwidth}
    \textbf{\textcolor{myred}{Output Response}}
    \begin{itemize}
    \item \textbf{Format:} JSON dictionary with strategic planning details:
    \end{itemize}
    \begin{minted}[bgcolor=white, frame=lines, framesep=2mm]{json}
    {
      "route": [
        {"keypoint": "Selected striking point 1", 
         "angle": "20 degrees to the north"},
        {"keypoint": "Selected striking point 2",  
         "angle": "15 degrees to the east"},
         ...
        {"keypoint": "Hole", "angle": "0 degrees"}
      ],
      "stroke_count": 1,
      "total_distance": 50.0
    }
    \end{minted}

\textbf{\textcolor{myred}{Rule Definitions and Reasoning Steps}}
\begin{enumerate}
    \item \textbf{Keypoint Selection:}
    \begin{itemize}
        \item Identify and mark feasible hitting points considering obstacle locations, estimated ball trajectory, and assist lines.
        \item Evaluate the feasibility of each key point based on the robotic arm's capabilities and the golf ball's dynamics.
    \end{itemize}
    \item \textbf{Route Planning:}
    \begin{itemize}
        \item Define a sequence of key points to form a route that minimizes the stroke count and total distance traveled.
        \item Optimize the route by adjusting the key points and angles to ensure a successful shot.
    \end{itemize}
    \item \textbf{Endpoint Verification:}
    \begin{itemize}
        \item Verify that the selected route reaches the desired endpoint (the hole).
        \item Adjust the route if necessary to ensure a successful shot.
    \end{itemize}
\end{enumerate}
\end{minipage}
\end{mdframed}

\clearpage
\paragraph{Parameter generation.} 
\bigskip
\begin{mdframed}[linewidth=1pt]

    \bigskip
    \begin{center}
            \textbf{\textcolor{myblue}{\large Detailed Prompt Example for Robotic Minigolf Playing Assistant}}
    \end{center}
    \medskip
\textcolor{myred}{\textbf{Purpose:}} This prompt is designed to guide the robotic assistant in estimating the hitting parameters for a minigolf shot, given a selected route and image of the court, to achieve a successful shot in a single attempt.

\textbf{\textcolor{myred}{Input Requests}}
\begin{itemize}
\item \textbf{Route Information:}
\begin{itemize}
\item \textit{Route JSON:} JSON dictionary with the selected route, including key points and angles:
\begin{minted}[bgcolor=white, frame=lines, framesep=2mm]{json}
{
    "route": [
        {"keypoint": "Selected striking point 1",
            "angle": "-20 degrees"},
        {"keypoint": "Selected striking point 2",
            "angle": "15 degrees"},
        ...
        {"keypoint": "Hole", "angle": "0 degrees"}
    ],
    "stroke_count": 1,
    "total_distance": 50.0
}
\end{minted}
\end{itemize}

\item \textbf{Image of the Court:}
\begin{itemize}
    \item \textit{Description:} Top-down view image of the minigolf court, including obstacles and endpoints.
\end{itemize}
\end{itemize}

\textbf{\textcolor{myred}{Output Response}}
\begin{itemize}
\item \textbf{Format:} JSON dictionary with hitting parameters:
\end{itemize}
\begin{minted}[bgcolor=white, frame=lines, framesep=2mm]{json}
{
    "hitting_angle": 25.0,
    "hitting_speed": 0.7,
    "confidence_score": 0.9
}
\end{minted}

\textbf{\textcolor{myred}{Rule Definitions and Reasoning Steps}}
\begin{enumerate}
\item \textbf{Hitting Parameter Estimation:}
\begin{itemize}
\item Utilize the provided route and image to estimate the hitting angle and speed required to reach the hole.
\item Consider the robotic arm's capabilities and the golf ball's dynamics in the estimation process.
\end{itemize}

\item \textbf{Confidence Score Calculation:}
\begin{itemize}
    \item Calculate a confidence score based on the estimated hitting parameters and the selected route.
    \item Adjust the hitting parameters if necessary to ensure a successful shot.
\end{itemize}
\end{enumerate}

\end{mdframed}

\clearpage
\paragraph{Hitting trajectory evaluation.}
\bigskip
\begin{mdframed}[linewidth=1pt]
    \bigskip
    \begin{center}
            \textbf{\textcolor{myblue}{\large Detailed Prompt Example for Robotic Minigolf Playing Assistant}}
    \end{center}
    \medskip

\textcolor{myred}{\textbf{Purpose:}} This prompt is designed to guide the robotic assistant in evaluating the hitting trajectory and parameters of a minigolf shot, given the recorded trajectory and original route estimation, to identify areas for improvement and provide suggestions for modification.

\textbf{\textcolor{myred}{Input Requests}}
\begin{itemize}
\item \textbf{Trajectory Information:}
\begin{itemize}
\item \textit{Recorded Trajectory:} The actual trajectory of the golf ball during the previous execution.
\item \textit{Original Route Estimation:} The originally expected route is estimated by connecting the key points and trajectory histories.
\item \textit{Hitting Parameters:} The hitting speed and angle used in the previous execution.
\end{itemize}
\end{itemize}

\textbf{\textcolor{myred}{Output Response}}
\begin{itemize}
\item \textbf{Format:} JSON dictionary with evaluation and suggestion details:
\end{itemize}
\begin{minted}[bgcolor=white, frame=lines, framesep=2mm]{json}
{
    "endpoint_reached": true/false,
    "deviation_reason": "insufficient speed" / "incorrect angle" /...,
    "modification_suggestion": {
        "parameter": "hitting speed" / "hitting angle",
        "direction": "increase" / "decrease",
        "amount": 0.1
    }
}
\end{minted}

\textbf{\textcolor{myred}{Rule Definitions and Reasoning Steps}}
\begin{enumerate}
\item \textbf{Endpoint Verification:}
\begin{itemize}
\item Check if the golf ball stopped in the desired endpoint (under the red mask).
\item If yes, no further evaluation is needed.
\end{itemize}

\item \textbf{Deviation Analysis:}
\begin{itemize}
    \item Identify the reason for deviation from the endpoint (e.g., insufficient speed, incorrect angle).
    \item Evaluate the deviation from several perspectives (e.g., speed, angle, trajectory shape).
\end{itemize}

\item \textbf{Modification Suggestion:}
\begin{itemize}
    \item Based on the deviation reason, suggest modifying the hitting speed or angle in a specific direction and amount.
    \item Provide a detailed explanation for the suggestion.
\end{itemize}
\end{enumerate}

\end{mdframed}
\clearpage
\paragraph{Feasibility evaluations.}
\bigskip
\begin{mdframed}[linewidth=1pt]
    \bigskip
    \begin{center}
            \textbf{\textcolor{myblue}{\large Detailed Prompt Example for Robotic Minigolf Playing Assistant}}
    \end{center}
    \medskip

\textcolor{myred}{\textbf{Purpose:}} This prompt is designed to guide the robotic assistant in evaluating the feasibility of a minigolf court scenario, given the image with assistive lines and obstacle masks, as well as the evaluation of previous hitting attempts, to determine whether the court is feasible and provide reasons for impossibility.

\textbf{\textcolor{myred}{Input Requests}}
\begin{itemize}
\item \textbf{Court Information:}
\begin{itemize}
\item \textit{Image with Assistive Lines:} Top-down view image of the minigolf court with annotated assistive lines.
\item \textit{Obstacle Masks:} Masks of the plastic obstacles on the court.
\end{itemize}

\item \textbf{Previous Attempt Evaluation:}
\begin{itemize}
    \item \textit{Hitting Attempt Results:} Evaluation of previous hitting attempts, including success/failure and reasons for failure.
\end{itemize}
\end{itemize}

\textbf{\textcolor{myred}{Output Response}}
\begin{itemize}
\item \textbf{Format:} JSON dictionary with feasibility decision and reason:
\end{itemize}
\begin{minted}[bgcolor=white, frame=lines, framesep=2mm]{json}
{
    "feasibility": true/false,
    "reason": "insufficient space" / "obstacle blocking" /...
}
\end{minted}

\textbf{\textcolor{myred}{Rule Definitions and Reasoning Steps}}
\begin{enumerate}
\item \textbf{Court Analysis:}
\begin{itemize}
\item Analyze the court image with assistive lines and obstacle masks to identify potential obstacles and challenges.
\item Evaluate the feasibility of the court based on the robotic arm's capabilities and the golf ball's dynamics.
\end{itemize}

\item \textbf{Previous Attempt Consideration:}
\begin{itemize}
    \item Consider the evaluation of previous hitting attempts to identify patterns and reasons for failure.
    \item Use this information to inform the feasibility decision.
\end{itemize}

\item \textbf{Feasibility Decision:}
\begin{itemize}
    \item Based on the court analysis and previous attempt consideration, make a feasibility decision (true/false).
    \item Provide a reason for the decision if the court is deemed infeasible.
\end{itemize}
\end{enumerate}

\end{mdframed}

\clearpage
\paragraph{Modification Suggestions}
\begin{mdframed}[linewidth=1pt]
\begin{minipage}{\textwidth}
    \bigskip
    \begin{center}
            \textbf{\textcolor{myblue}{\large Detailed Prompt Example for Robotic Minigolf Playing Assistant}}
    \end{center}
    \medskip

\textcolor{myred}{\textbf{Purpose:}} This prompt is designed to guide the robotic assistant in suggesting modifications to the minigolf course or golf ball selection to achieve reflective equilibrium, making the task possible, given the evaluation results and possible modification types.

\textbf{\textcolor{myred}{Input Requests}}
\begin{itemize}
\item \textbf{Evaluation Results:}
\begin{itemize}
\item \textit{Visibility Value:} The current visibility value of the golf ball on the court.
\item \textit{Modification Types:} The types of modifications that can be applied to the current status, including:
\begin{itemize}
\item Removing obstacles
\item Setting obstacles
\item Watering angles and positions of existing obstacles
\item Changing the endpoint
\item Adding new obstacles
\end{itemize}
\end{itemize}

\item \textbf{Obstacle Information:}
\begin{itemize}
    \item \textit{Descriptions and Names:} Descriptions and names of the obstacles that can be added, including their properties and effects on the game.
\end{itemize}
\end{itemize}

\textbf{\textcolor{myred}{Output Response}}
\begin{itemize}
\item \textbf{Format:} JSON dictionary with exact execution instructions:
\end{itemize}
\begin{minted}[bgcolor=white, frame=lines, framesep=2mm]{json}
{
    "modification_instructions": [
        {"type": "remove", "obstacle": "obstacle 1"},
        {"type": "add", "obstacle": "obstacle 2", "position": "top-left",
            "angle": "30 degrees"},
        {"type": "change", "endpoint": "new endpoint location"}
    ]
}
\end{minted}
\end{minipage}
\textbf{\textcolor{myred}{Rule Definitions and Reasoning Steps}}
\begin{enumerate}
\item \textbf{Modification Analysis:}
\begin{itemize}
\item Analyze the evaluation results and possible modification types to identify the most effective modifications.
\item Evaluate the feasibility of each modification based on the robotic arm's capabilities and the golf ball's dynamics.
\end{itemize}

\item \textbf{Obstacle Selection:}
\begin{itemize}
    \item Select the most suitable obstacles to add or remove based on their properties and effects on the game.
    \item Determine the optimal positions and angles for the added obstacles.
\end{itemize}

\item \textbf{Endpoint Adjustment:}
\begin{itemize}
    \item Determine if the endpoint needs to be changed to achieve reflective equilibrium.
    \item Calculate the new endpoint location if necessary.
\end{itemize}

\item \textbf{Instruction Generation:}
\begin{itemize}
    \item Generate exact execution instructions based on the modification analysis, obstacle selection, and endpoint adjustment.
    \item Provide a detailed explanation for the suggested modifications.
\end{itemize}
\end{enumerate}

\end{mdframed}

\subsubsection{Robot Trajectory Generation}\label{app:robotexecution}

We compute the robot trajectory to hit the ball with the head of the club based on two input parameters for hitting angle and hitting velocity and assume a fixed Cartesian position of the ball.
Initially, the system computes a joint configuration through inverse kinematics to reach the golf ball position with the head of the minigolf club at the correct angle.
To achieve a target Cartesian velocity of the club at the ball location, we consider a second inverse kinematics solution assumed to be reached 100ms before the ball location, thus fixing the Cartesian position to the ball location given the target velocity.
The resulting two waypoints define the plateau of a trapezoidal velocity profile in joint space and the corresponding ramp up/ramp down phases are computed with pre-determined accelerations. Due to the kinematic design of our setup and the dynamics limits for the robot joints, the ball position always constitutes the lowest point of the computed trajectory.
All motions are kinematically validated through the MoveIt framework before execution.

\section{Experimental Setups}\label{app:experimentalsetups}

\subsection{Robot Hardware}

For the robot, we adopted the UR5 robot arm, mounted on the wall. It is a 6 DoF robot arm with reasonable accuracy. The gripper we employed the 3-Finger Adaptive Robot Gripper from Robotiq, is renowned for its high versatility and adaptability across varied robotic applications. This advanced gripper is designed to efficiently handle objects of almost any shape, thanks to its unique ability to adjust grip based on the object's form. Equipped with multiple gripping modes and precise control over each finger, the gripper integrates seamlessly with systems like Universal Robots, making it a prime choice for intricate tasks in modern automated environments.

The golf club's holder is 3D-printed, constituted by two parts. The first part is the holder that fits the 3-finger gripper, the second part is the gripper for the golf club. The two parts are printed using PLA material and glued together.

\subsection{Minigolf settings}

The minigolf setting, sourced as a complete set from the manufacturer, is designed to provide everything needed for playing minigolf. This includes the convenience of having all necessary equipment like clubs, balls, scorecards, and a variety of obstacles. These components are engineered for durability, being break-resistant, UV-protected, and weatherproof, making them suitable for both indoor and outdoor use. The modular design allows for the creation of customized courses, maximizing space and enhancing gameplay flexibility, ideal for family entertainment and varied play environments.

The minigolf set comes with different types of balls for selection, in different colours and textures. By testing, we figured that the red-colored ball with a relatively soft texture provides the most stable performance with different obstacles and the most notable optical character in the recordings.

For the ground, we select an artificial grass carpet to better simulate the real golf court. we set the course to be the size of 2 meters by 3 meters.

\subsection{Hardware for recording}

We adopt the davis 346~\citep{inivation2019davis346} for the event info recordings 
The DAVIS 346 is a high-performance event camera featuring a 346 x 260-pixel dual-mode sensor, combining Dynamic Vision Sensor~(DVS) technology and Active Pixel Sensor~(APS) capabilities. This model boasts remarkable specs such as a 120 dB DVS dynamic range and minimal latency of around 20 microseconds, suited for precise imaging tasks. Its robust design includes an anodized aluminum case and a CS-mount lens system, making it adaptable for various optical needs. Additionally, the camera is equipped for multi-camera synchronization and supports power and data transfer via a USB 3.0 connection, ensuring versatility in deployment for advanced visual applications.

We adopt the Azure Kinect~\citep{microsoft2024azurekinect} as the RGB-D camera. The Azure Kinect DK represents a significant advancement in sensor technology, integrating a sophisticated depth camera system and a high-resolution color camera into a single device. This tool is designed for developers and commercial applications, featuring a robust set of capabilities including multiple operating modes for depth perception and high-resolution video capture. Notably, the device is housed in a durable case, supports extensive environmental operating ranges, and offers advanced synchronization features. This makes the Azure Kinect DK a versatile and powerful tool for a variety of technological applications.

\section{Typical Courts / Tasks and Underlying Challenges}\label{ap:typical_scenairos}
\subsection{Underlying planning and kinodynamic reasoning challenges in typical courts}
As shown in Table~\ref{apptab:typicalcourts}, we highlight several typical complex courts to demonstrate the inherent challenges of planning reasonable pathways and generating refined hitting actions, which necessitate a profound kinodynamic understanding.

\begin{table}[H]
    \centering
    \setlength{\tabcolsep}{2pt} 
    \renewcommand{\arraystretch}{3} 
    \begin{tabular}{|>{\centering\arraybackslash}m{0.24\textwidth}|>{\centering\arraybackslash}m{0.7\textwidth}|}
        \hline
        \textbf{Court Visualization} & \textbf{Underlying Challenges} \\
        \hline
        \raisebox{-0.25\height}{\includegraphics[width=0.6\linewidth]{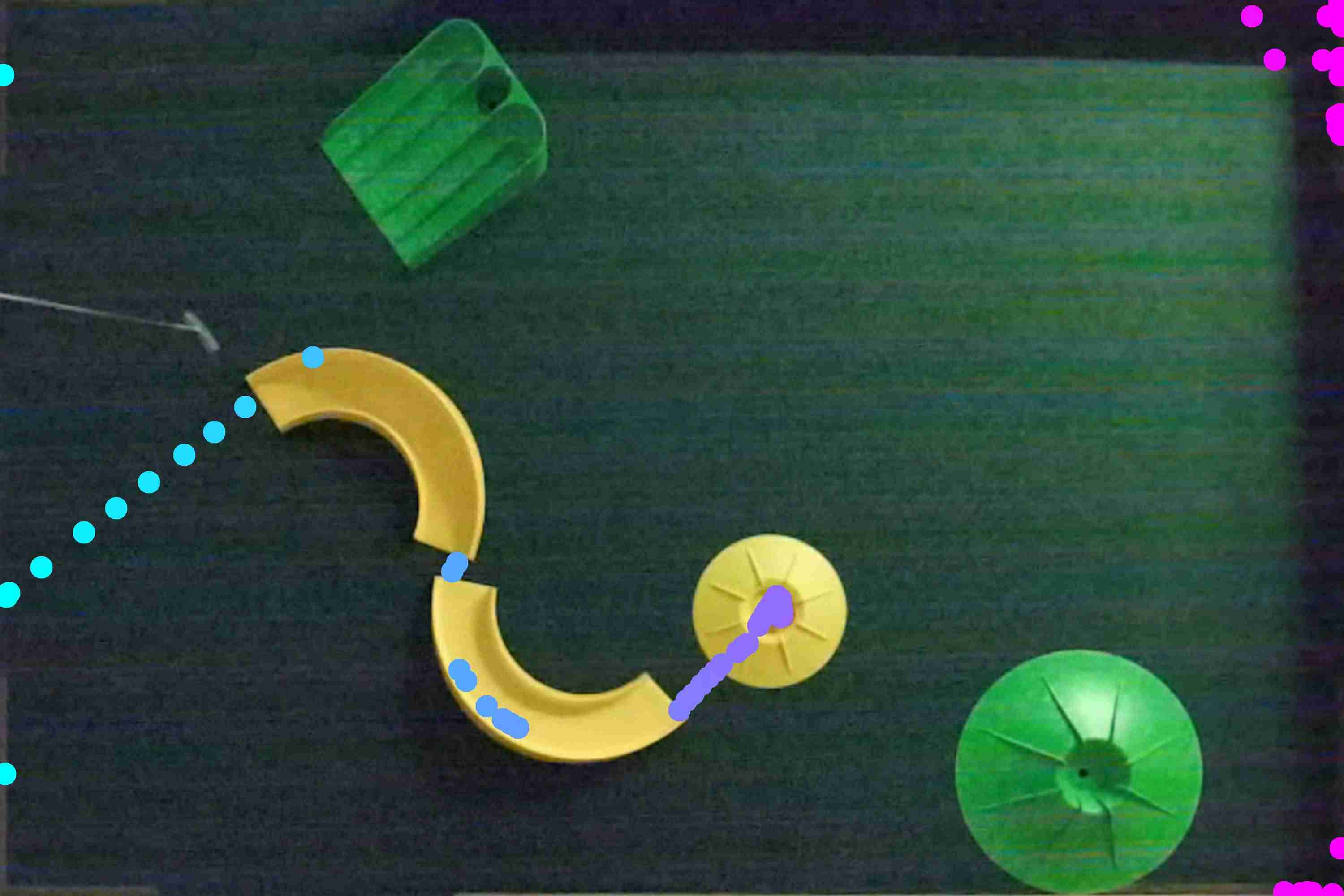}} & 
        \begin{itemize}[leftmargin=20pt,rightmargin=10pt]
            \item An imprecise angle or force could induce displacement of the two yellow curves via collision
            \item Deviated hitting angle will cause a great loss in speed, unable to go through the second curve
        \end{itemize}
        \\
        \hline
        \raisebox{-0.25\height}{\includegraphics[width=0.6\linewidth]{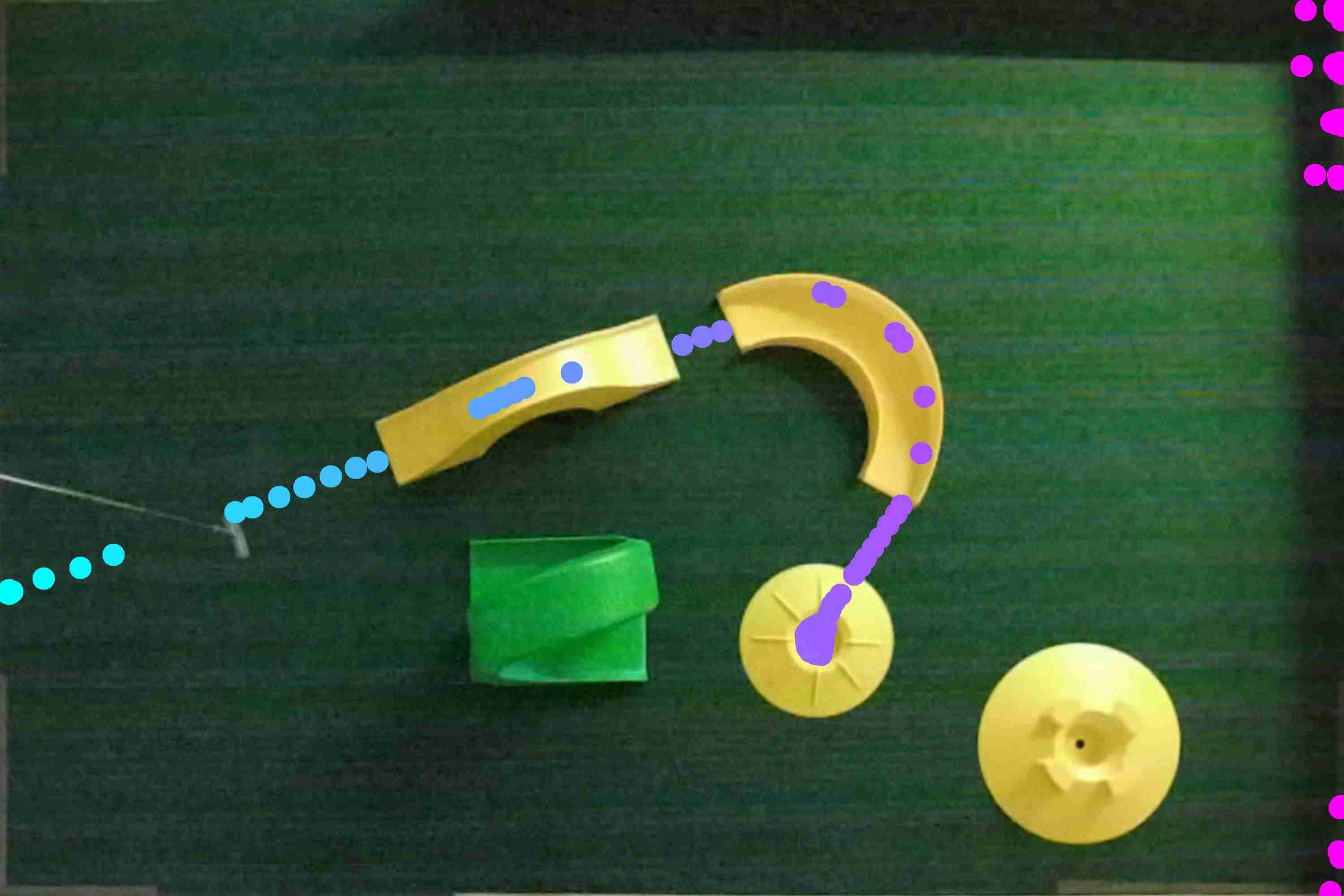}} & 
        \begin{itemize}[leftmargin=20pt, rightmargin=10pt]
            \item Identify the path involving the rollercoaster as impossible
            \item A small force would not traverse the camp, whereas an excessively large force might collide with the yellow curve
        \end{itemize}
        
        \\
        \hline
        \raisebox{-0.25\height}{\includegraphics[width=0.6\linewidth]{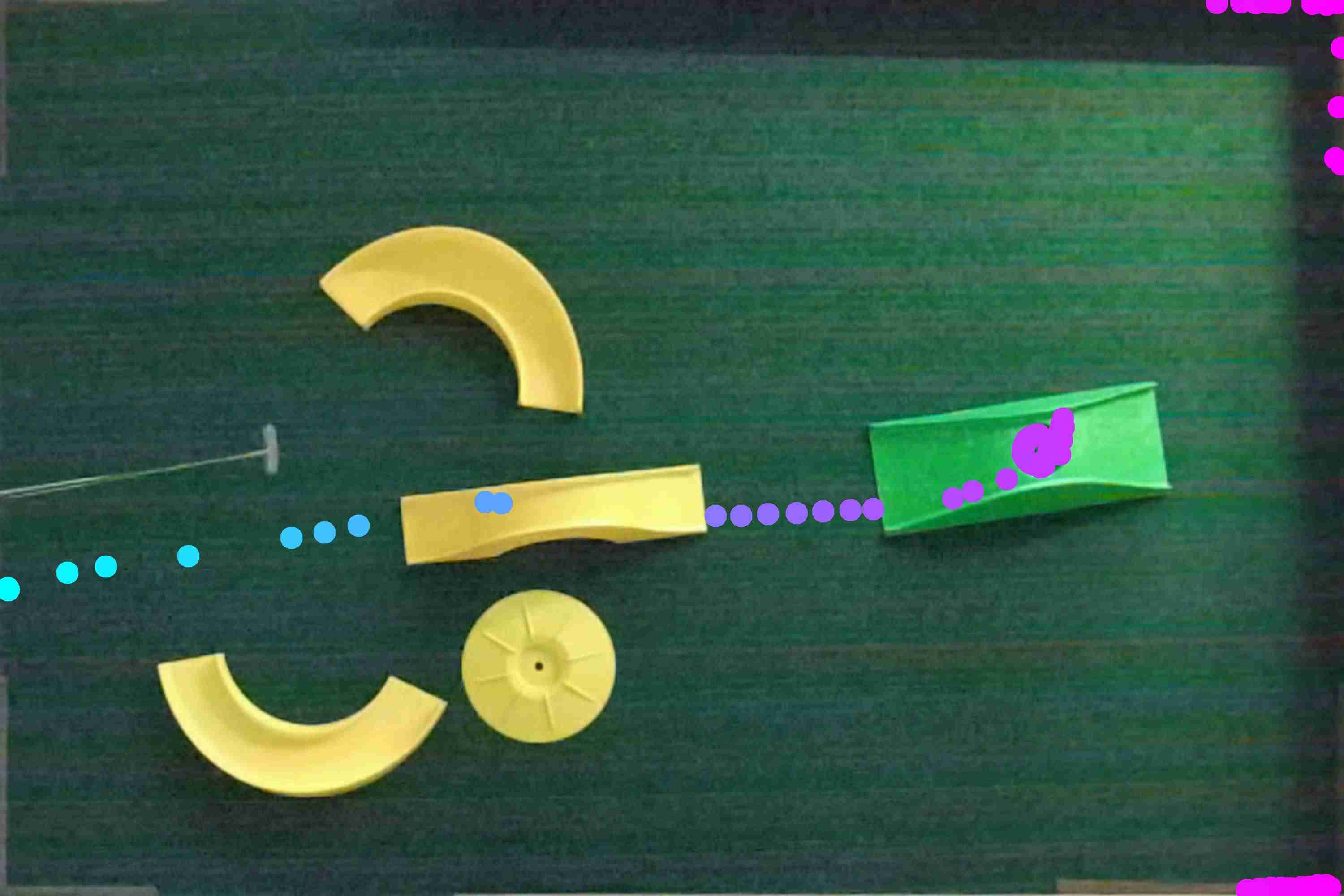}} & 
        \begin{itemize}[leftmargin=20pt, rightmargin=10pt]
            \item Hitting angle need to stay in the middle, colliding with the edge will cause deviation
            \item An insufficient force will not pass the ramp, yet excessive will go beyond the endpoint
        \end{itemize} \\
        \hline
        \raisebox{-0.25\height}{\includegraphics[width=0.6\linewidth]{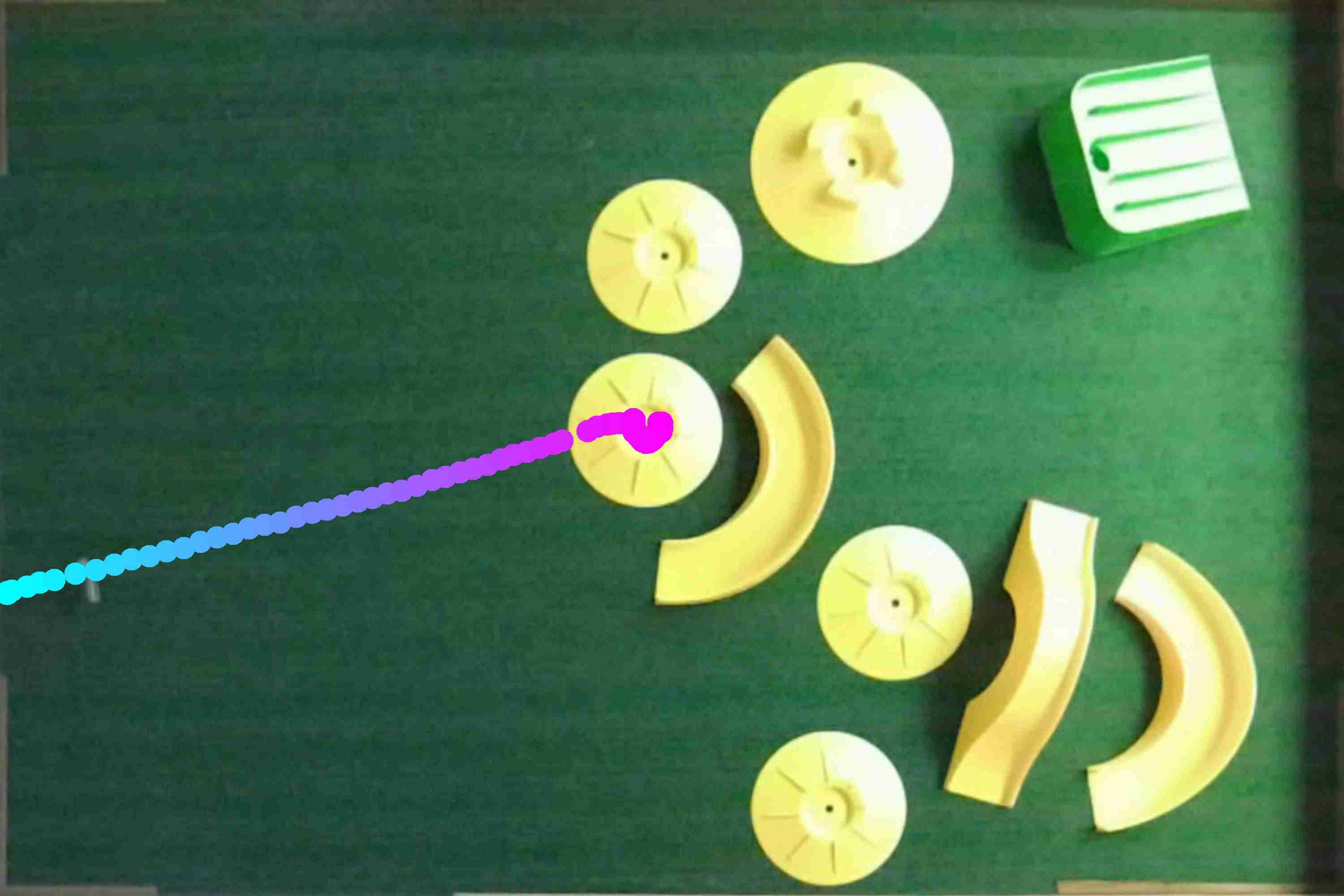}} & 
        \begin{itemize}[leftmargin=20pt, rightmargin=10pt]
            \item Distinguish the exact yellow disk endpoint from other compounding yellow disks 
        \end{itemize} \\
        \hline
        \raisebox{-0.25\height}{\includegraphics[width=0.6\linewidth]{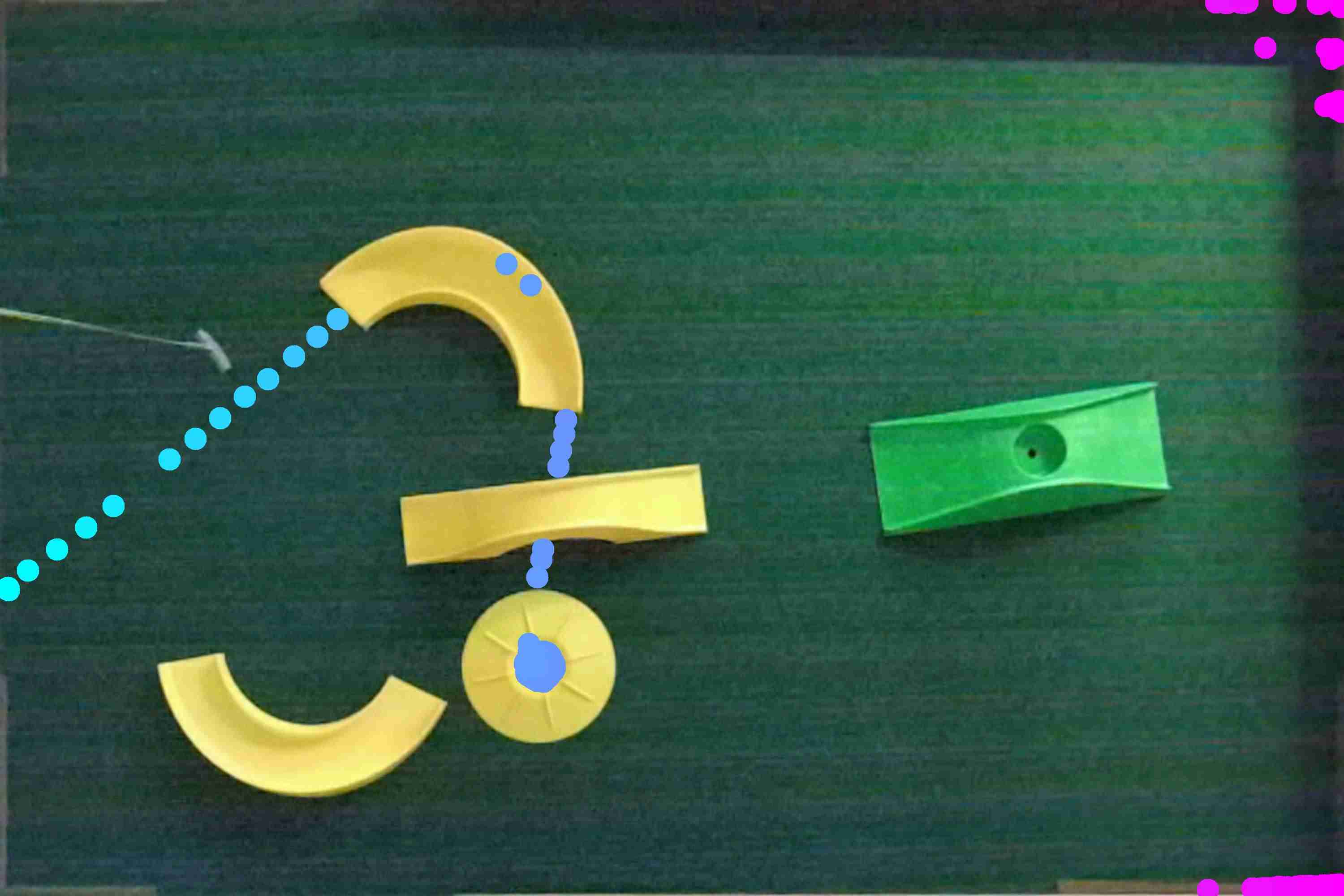}} & 
        \begin{itemize}[leftmargin=20pt, rightmargin=10pt]
            \item Identify the feasible pathway involving tunneling the hole under the ramp
        \end{itemize} \\
        \hline
        \raisebox{-0.25\height}{\includegraphics[width=0.6\linewidth]{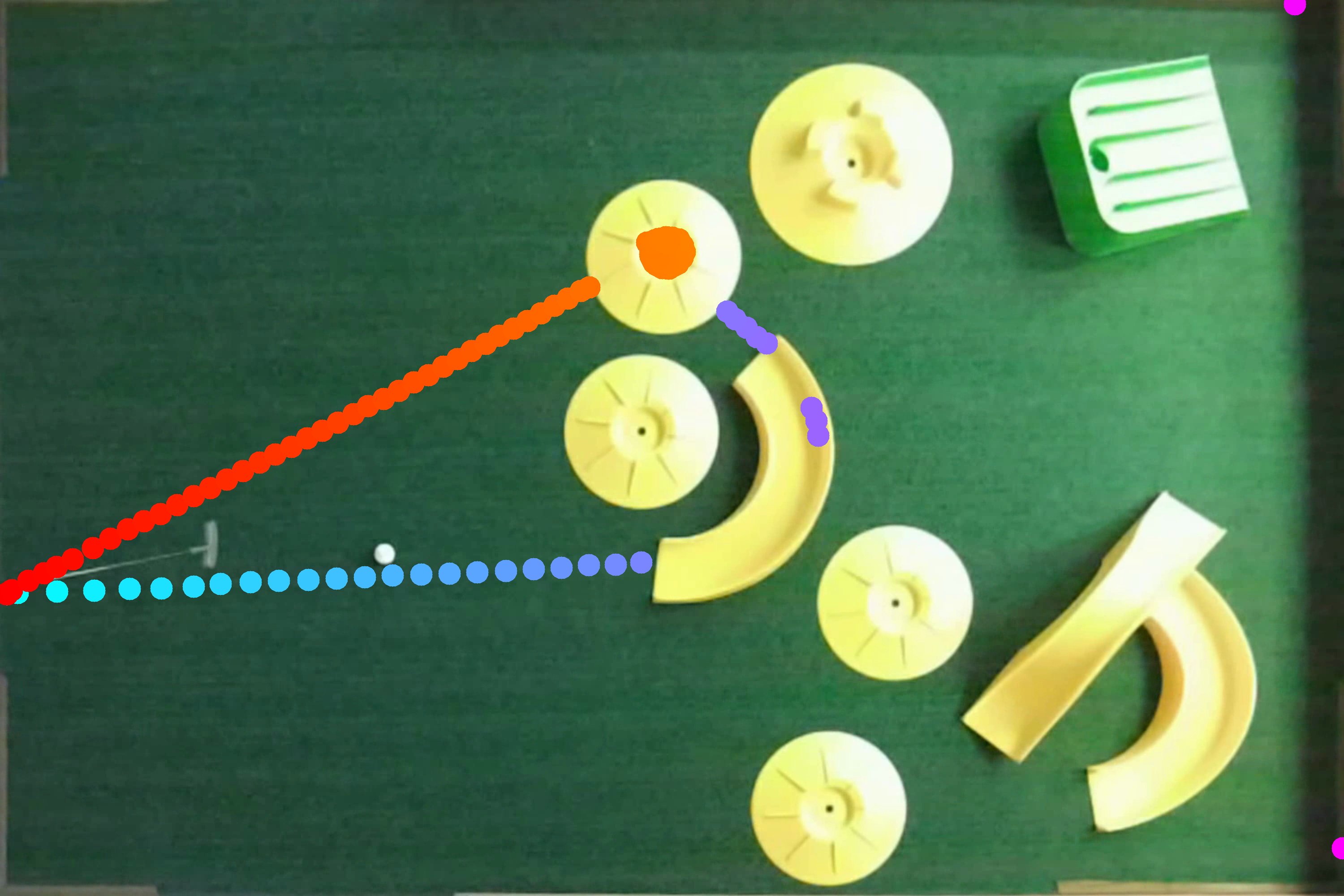}} & 
        \begin{itemize}[leftmargin=20pt, rightmargin=10pt]
            \item Identify the two feasible pathways and generate corresponding hitting actions
        \end{itemize} \\
        \hline
    \end{tabular}
    \vspace{2mm}
    \caption{\textbf{Underlying trajectory planning and kinodynamic reasoning challenges.}}
    \label{apptab:typicalcourts}
\end{table}

\subsection{Underlying counterfactual reasoning challenges in modifying courts}

As shown in Table~\ref{apptable:modifyingcourt}, we present prototype and evolved variant pairs, highlighting challenges in counterfactual reasoning. These include adjusting endpoints for excessive forces, adding ramps without altering directions, ensuring sufficient speed, replacing endpoints, modifying positions to avoid impossible hits, and preventing path obstructions. 

\begin{table}[H]
    \centering
    \setlength{\tabcolsep}{2pt} 
    \renewcommand{\arraystretch}{3} 
    \begin{tabular}{|>{\centering\arraybackslash}m{0.22\textwidth}|>{\centering\arraybackslash}m{0.22\textwidth}|>{\centering\arraybackslash}m{0.5\textwidth}|}
        \hline
        \textbf{Prototype} &\textbf{Evolved Variant} & \textbf{Modification Challenges} \\
        \hline
        \raisebox{-0.25\height}{\includegraphics[width=0.8\linewidth]{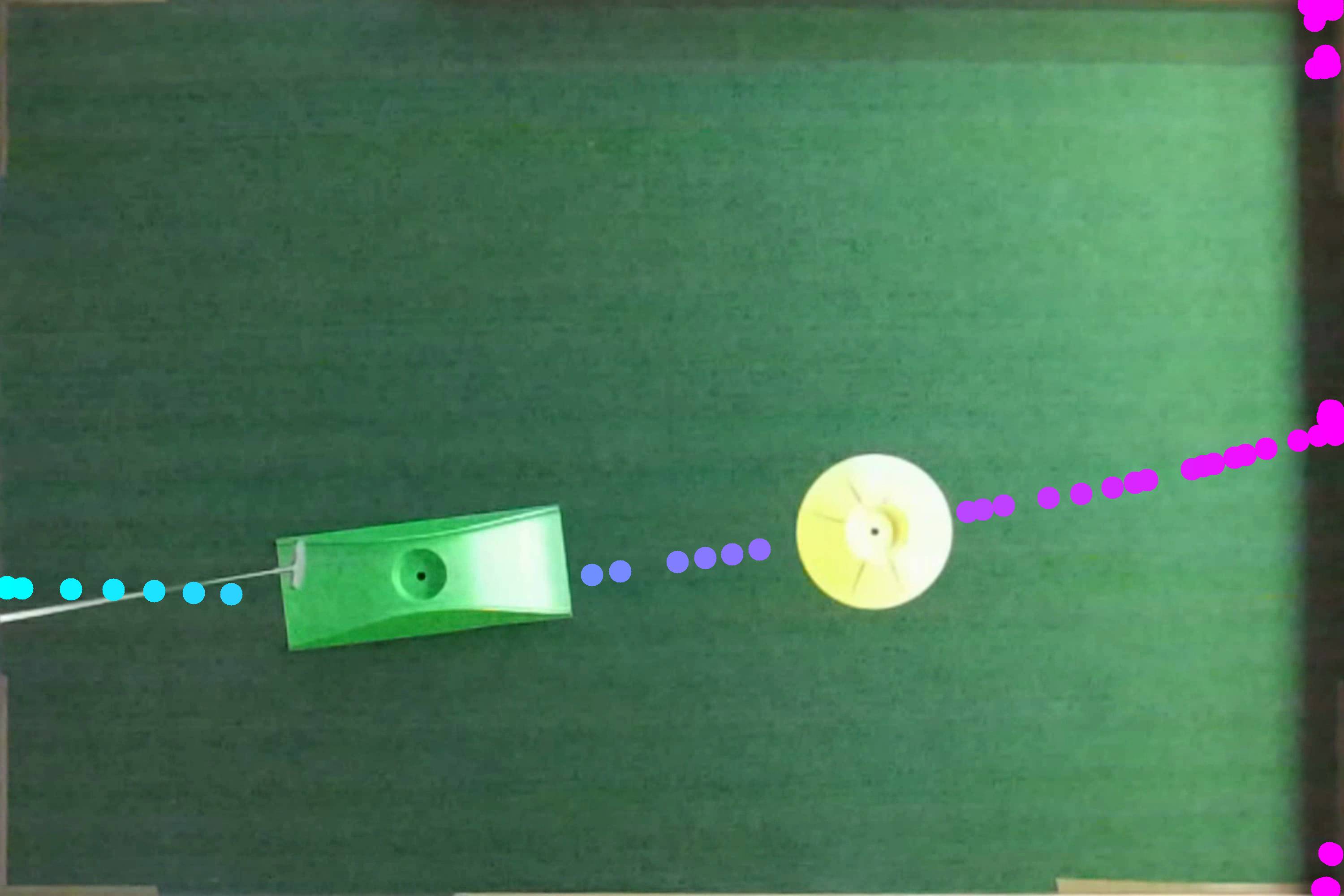}} & 
        \raisebox{-0.25\height}{\includegraphics[width=0.8\linewidth]{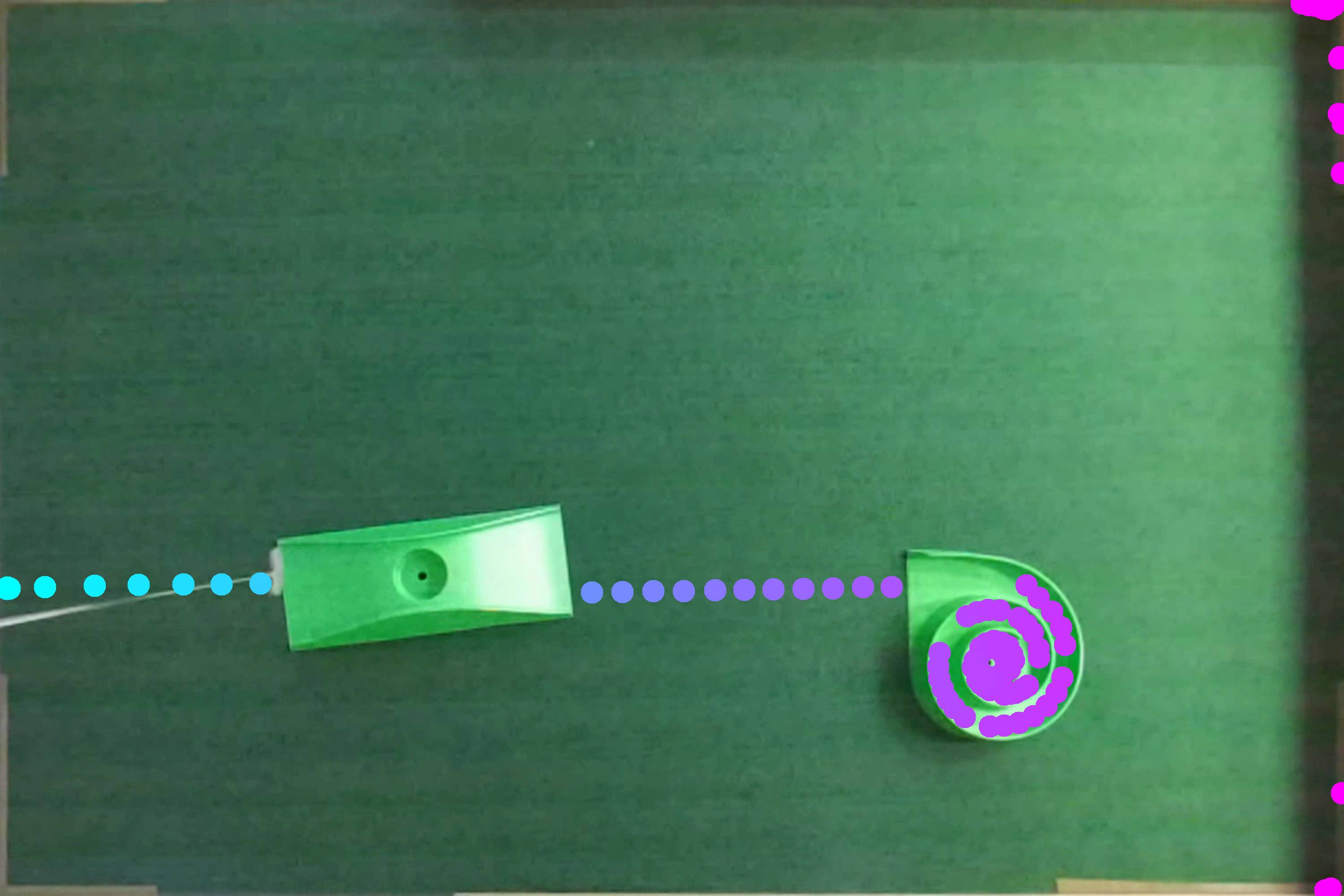}} & 
        \begin{itemize}[leftmargin=20pt,rightmargin=10pt]
            \item Realizing the force to pass the first ramp is too big to stop in the shallow endpoint
            \item Make suggestions on changing the endpoint to be more steep
        \end{itemize}
        \\
        \hline
        \raisebox{-0.25\height}{\includegraphics[width=0.8\linewidth]{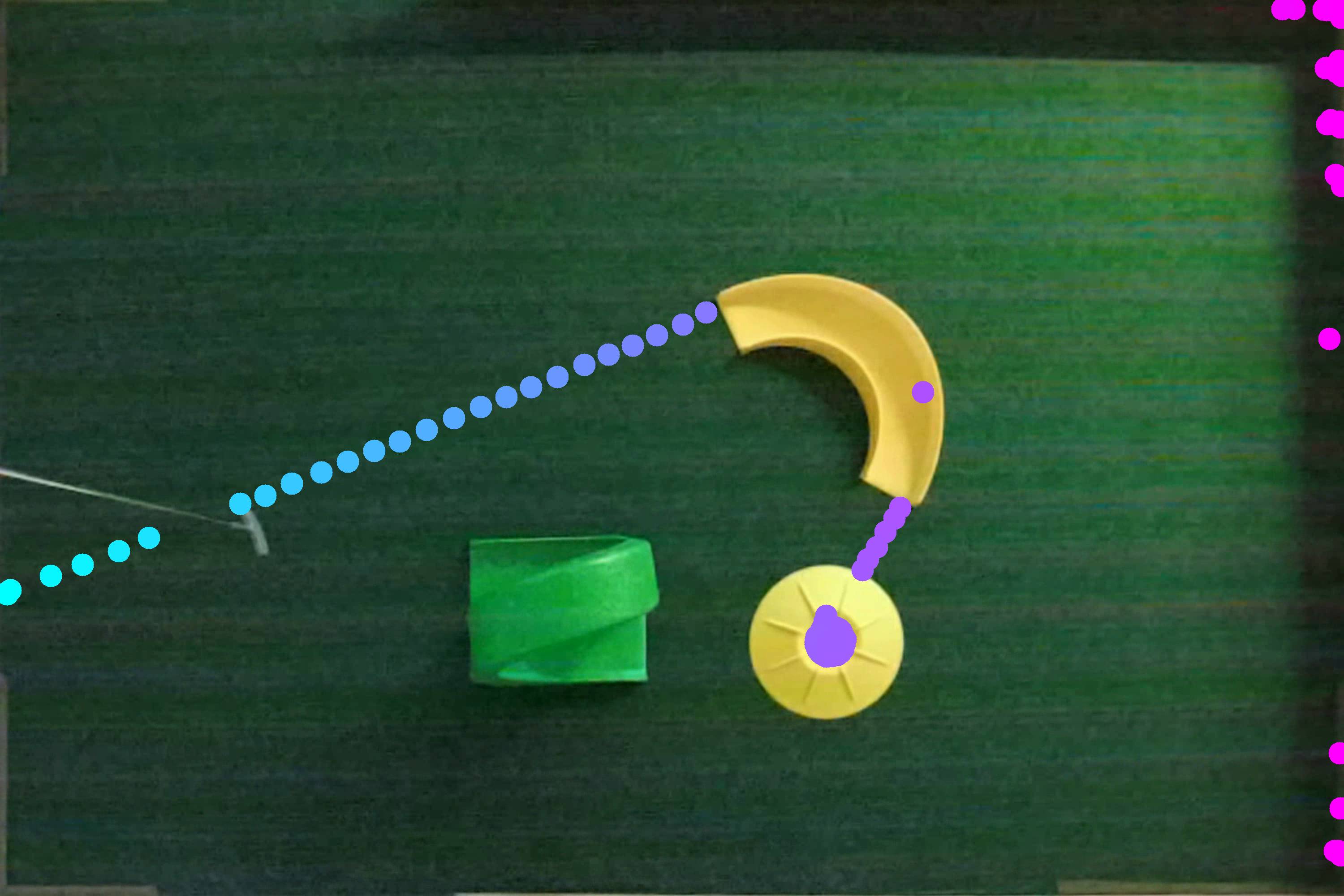}} & 
        \raisebox{-0.25\height}{\includegraphics[width=0.8\linewidth]{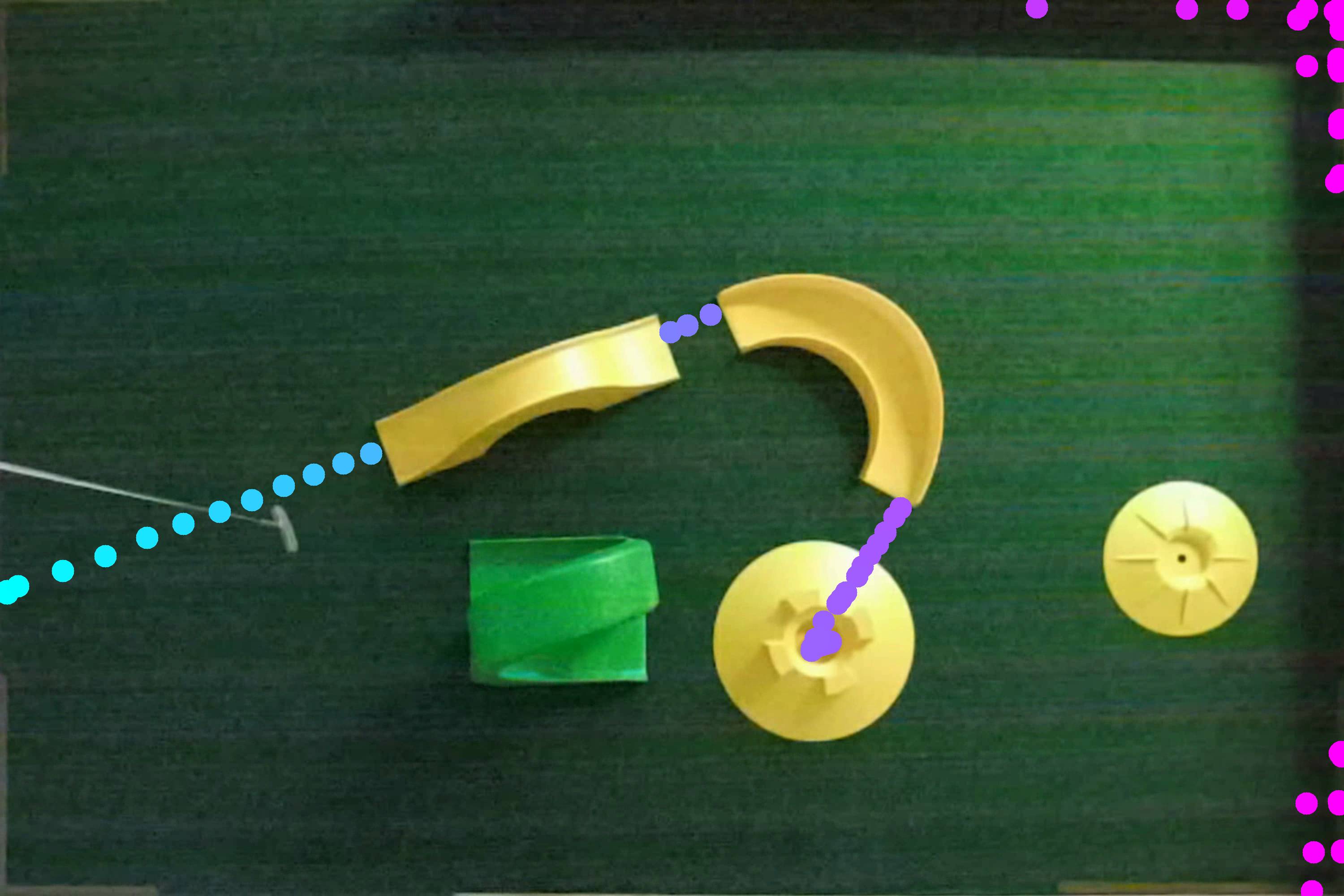}} & 
        \begin{itemize}[leftmargin=20pt,rightmargin=10pt]
            \item Realizing adding ramp along the path will have no alteration to the direction
            \item Change endpoint due to the high speed required by the ramp
        \end{itemize}
        \\
        \hline

        \raisebox{-0.25\height}{\includegraphics[width=0.8\linewidth]{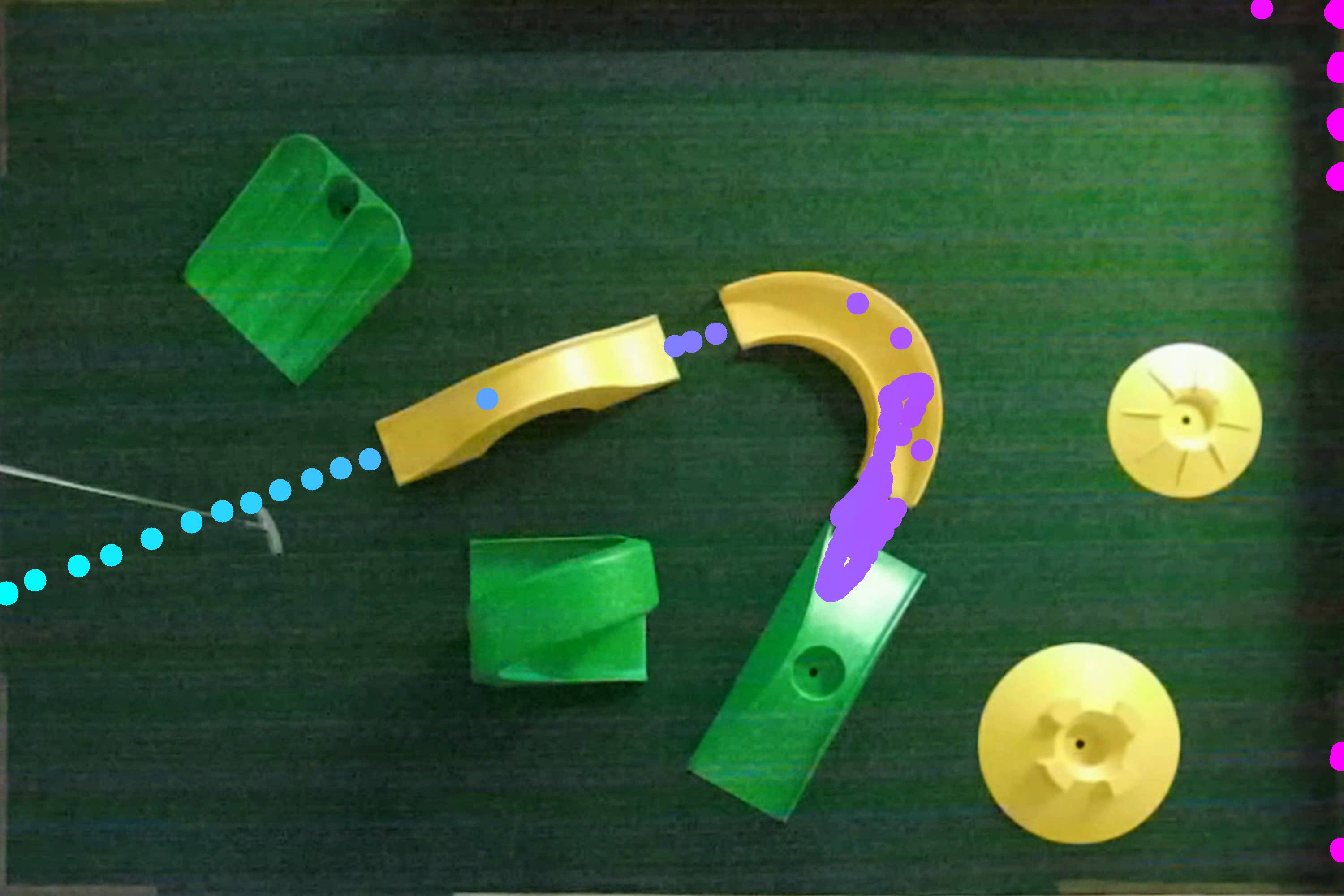}} & 
        \raisebox{-0.25\height}{\includegraphics[width=0.8\linewidth]{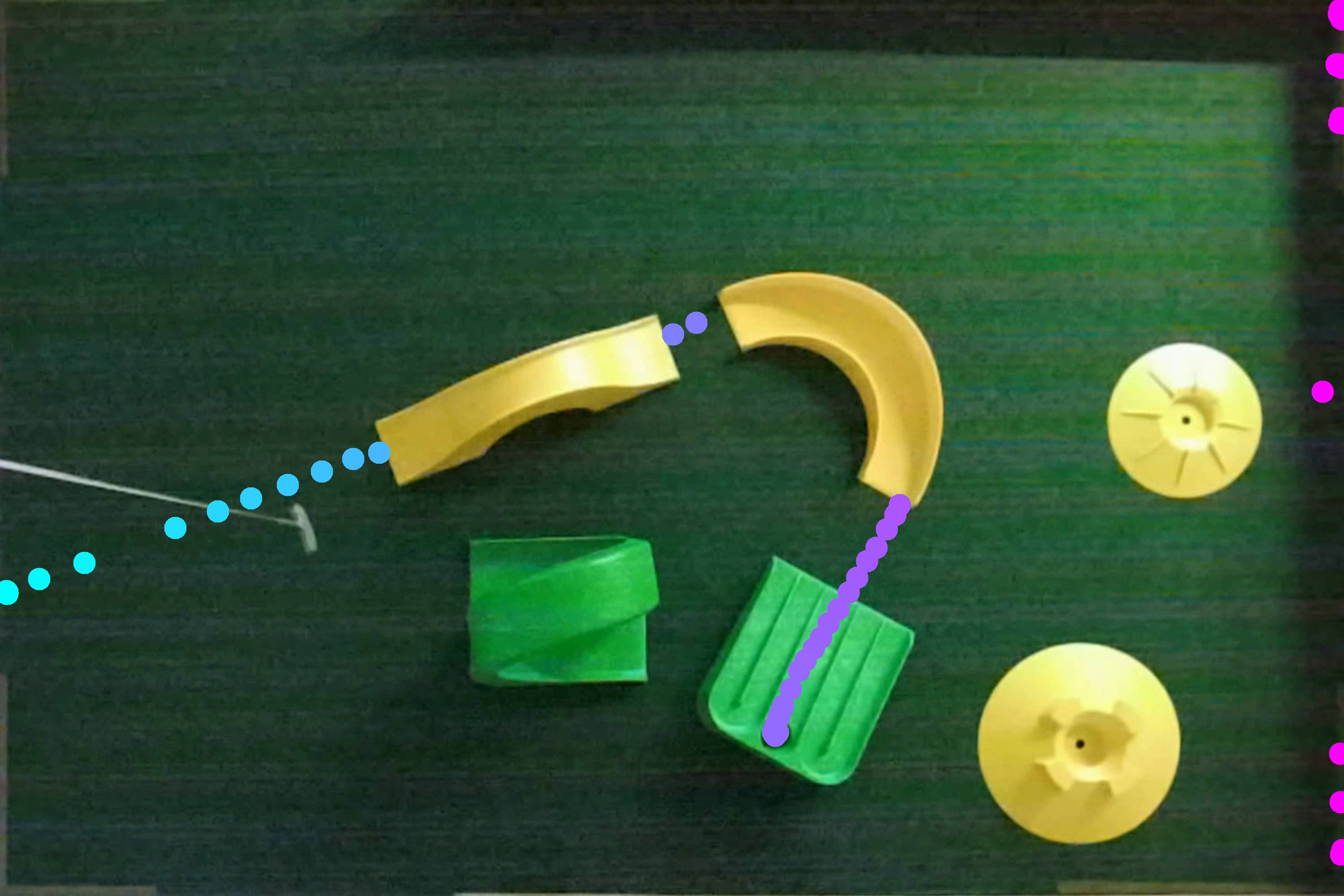}} & 
        \begin{itemize}[leftmargin=20pt,rightmargin=10pt]
            \item Identify the speed is insufficient to reach the endpoint
            \item Select endpoint that requires smaller speed
        \end{itemize}
        \\
        \hline

        \raisebox{-0.25\height}{\includegraphics[width=0.8\linewidth]{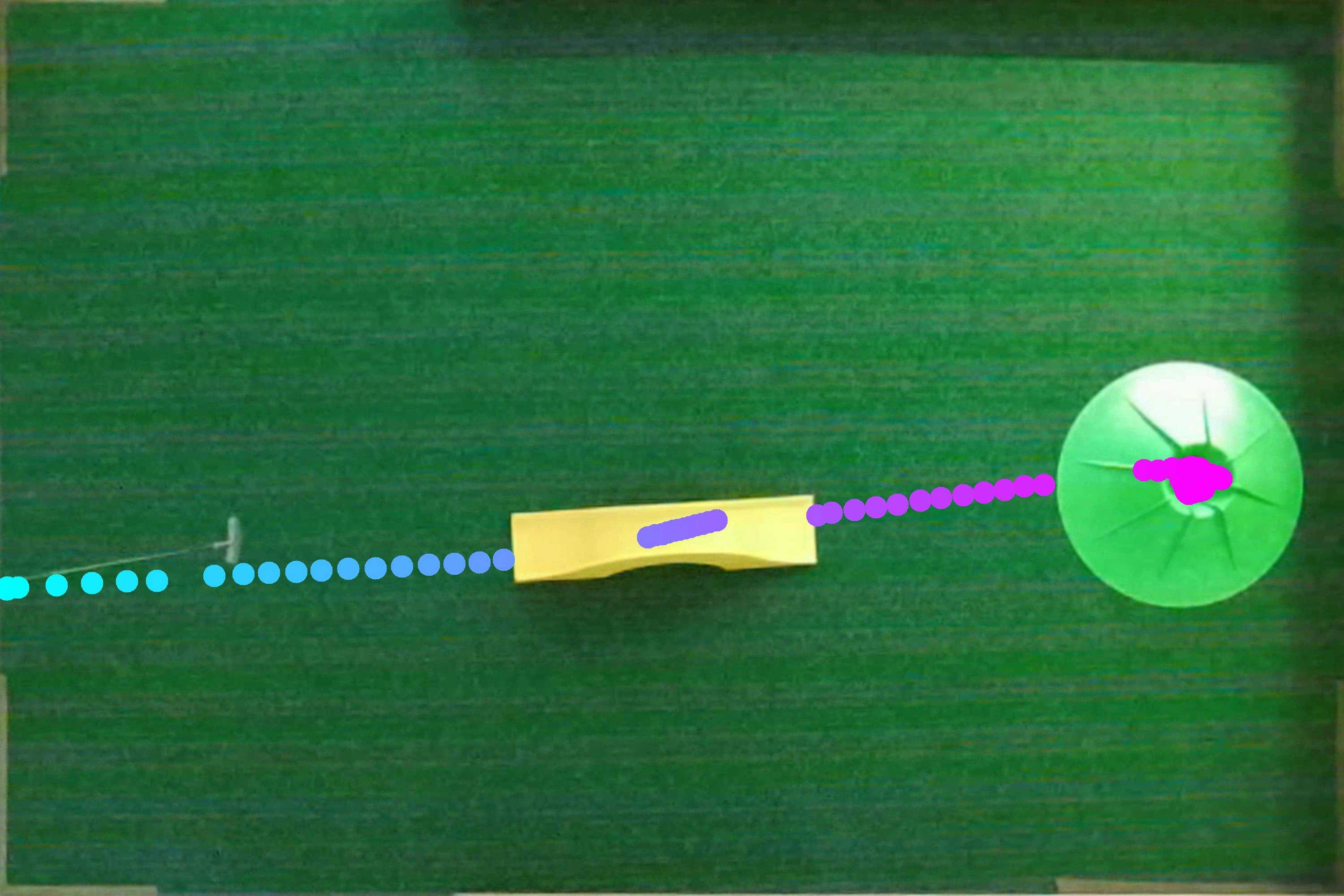}} & 
        \raisebox{-0.25\height}{\includegraphics[width=0.8\linewidth]{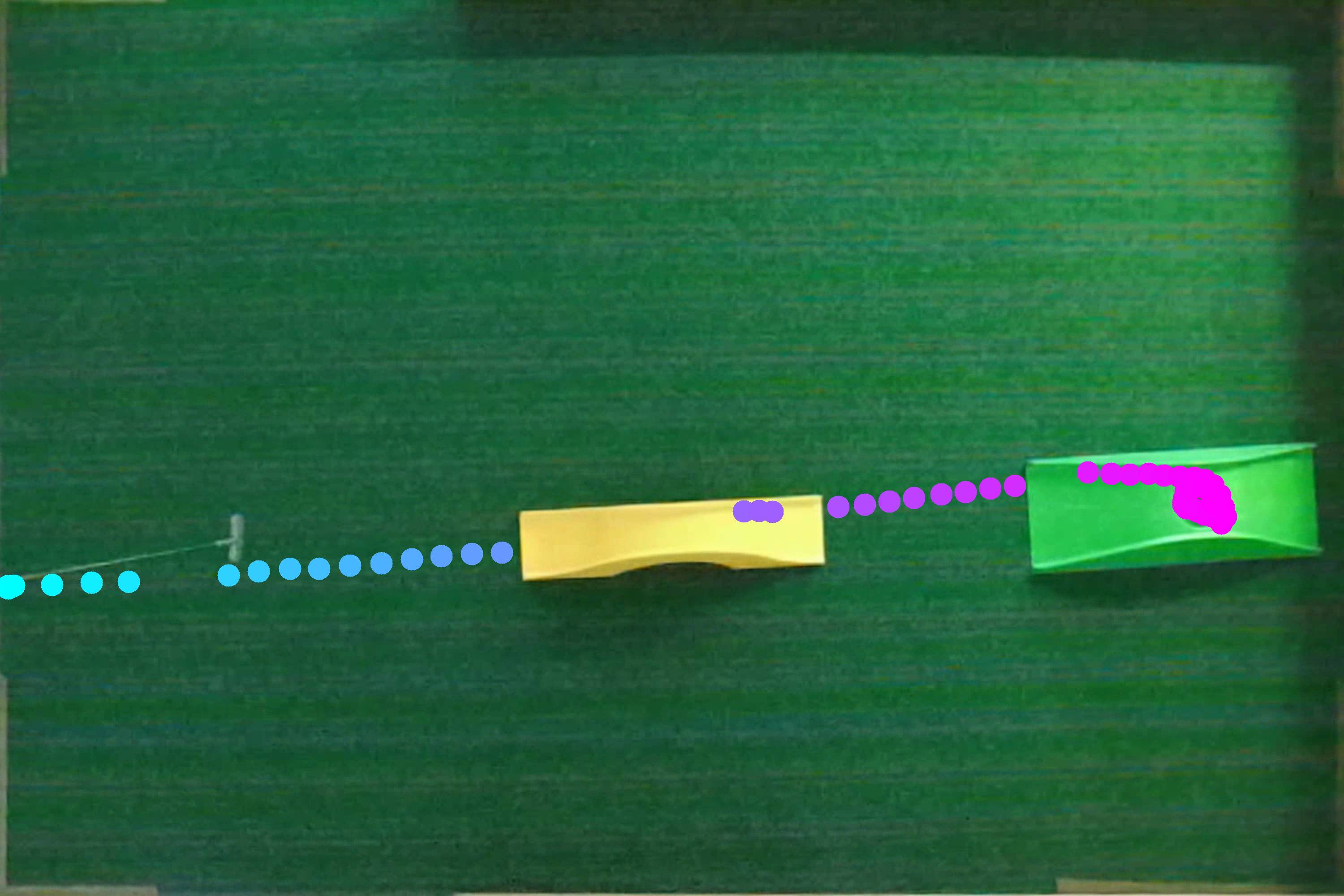}} & 
        \begin{itemize}[leftmargin=20pt,rightmargin=10pt]
            \item Decide the replacement of the endpoint
            
        \end{itemize}
        \\
        \hline

        \raisebox{-0.25\height}{\includegraphics[width=0.8\linewidth]{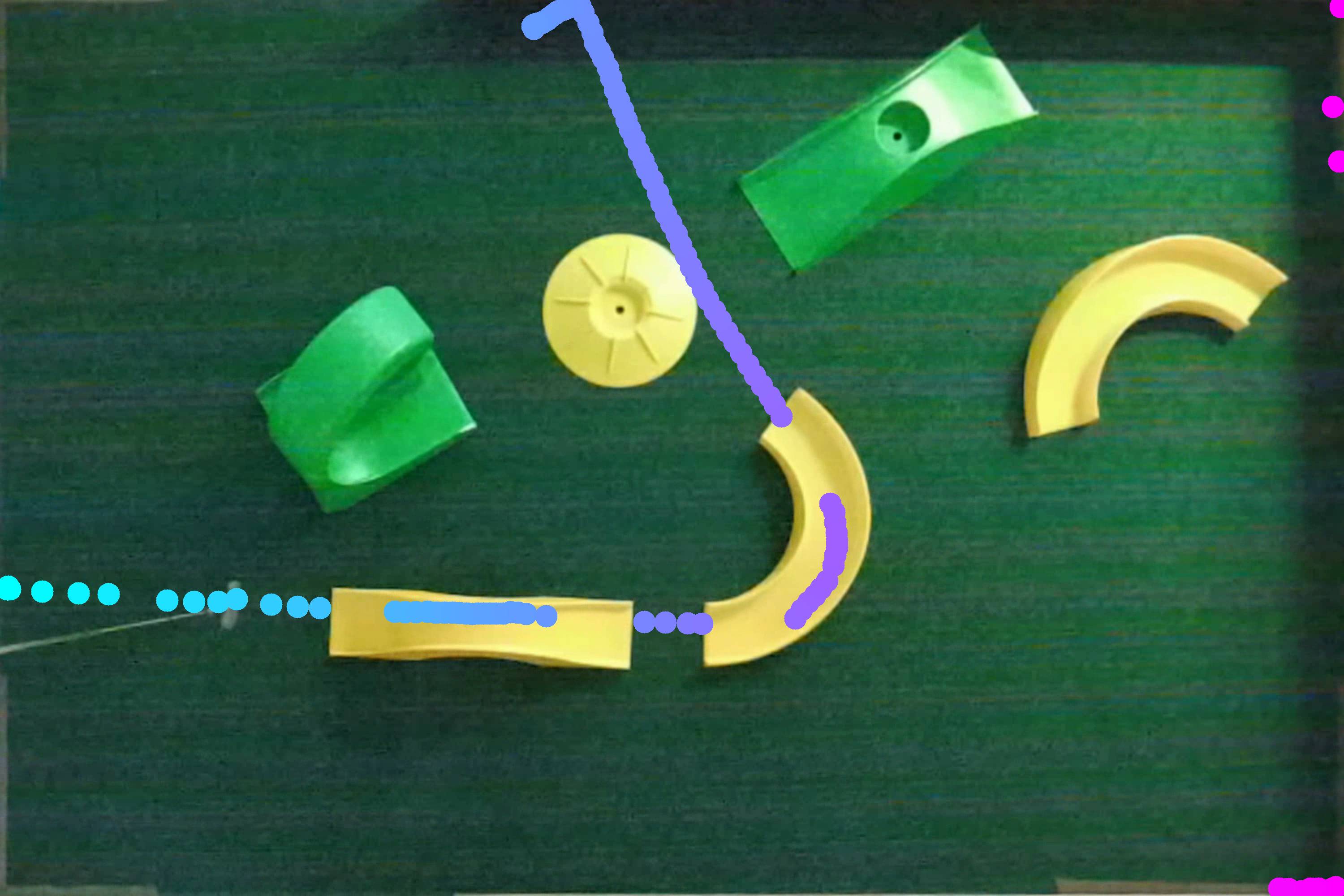}} & 
        \raisebox{-0.25\height}{\includegraphics[width=0.8\linewidth]{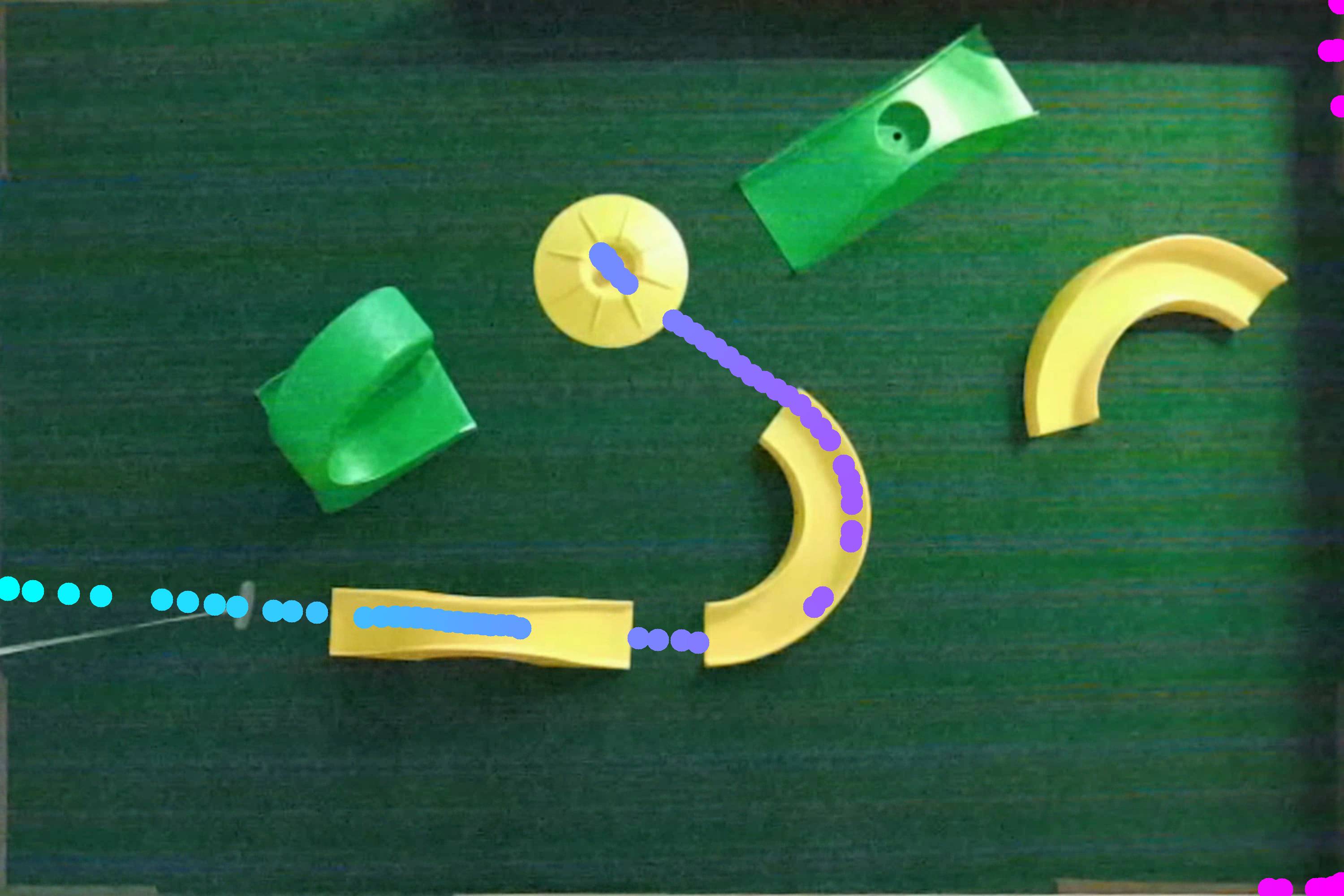}} & 
        \begin{itemize}[leftmargin=20pt,rightmargin=10pt]
            \item Identify the intricate position deviation causes the impossible hit
            \item Decide the intricate modification of the position
        \end{itemize}
        \\
        \hline

        \raisebox{-0.25\height}{\includegraphics[width=0.8\linewidth]{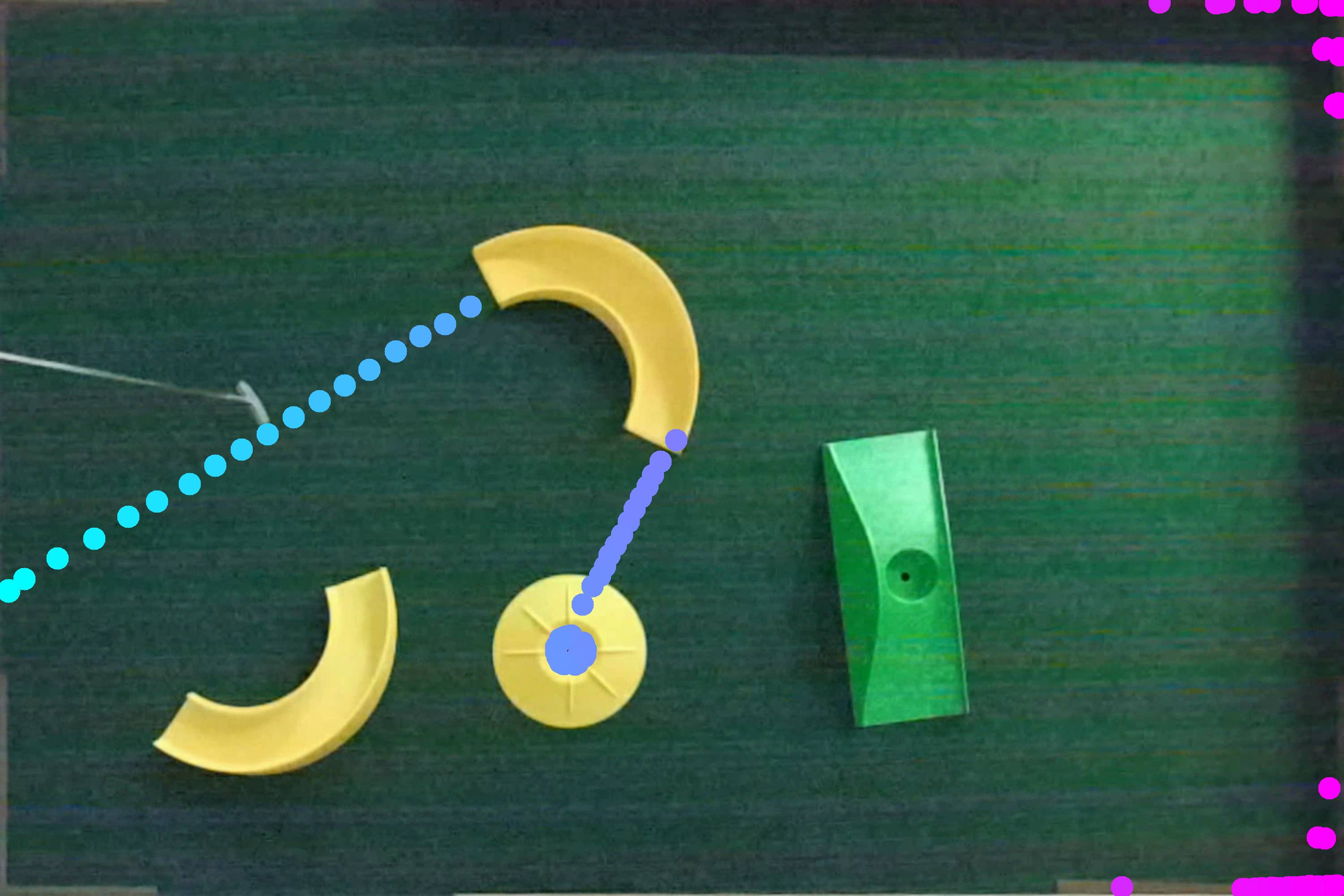}} & 
        \raisebox{-0.25\height}{\includegraphics[width=0.8\linewidth]{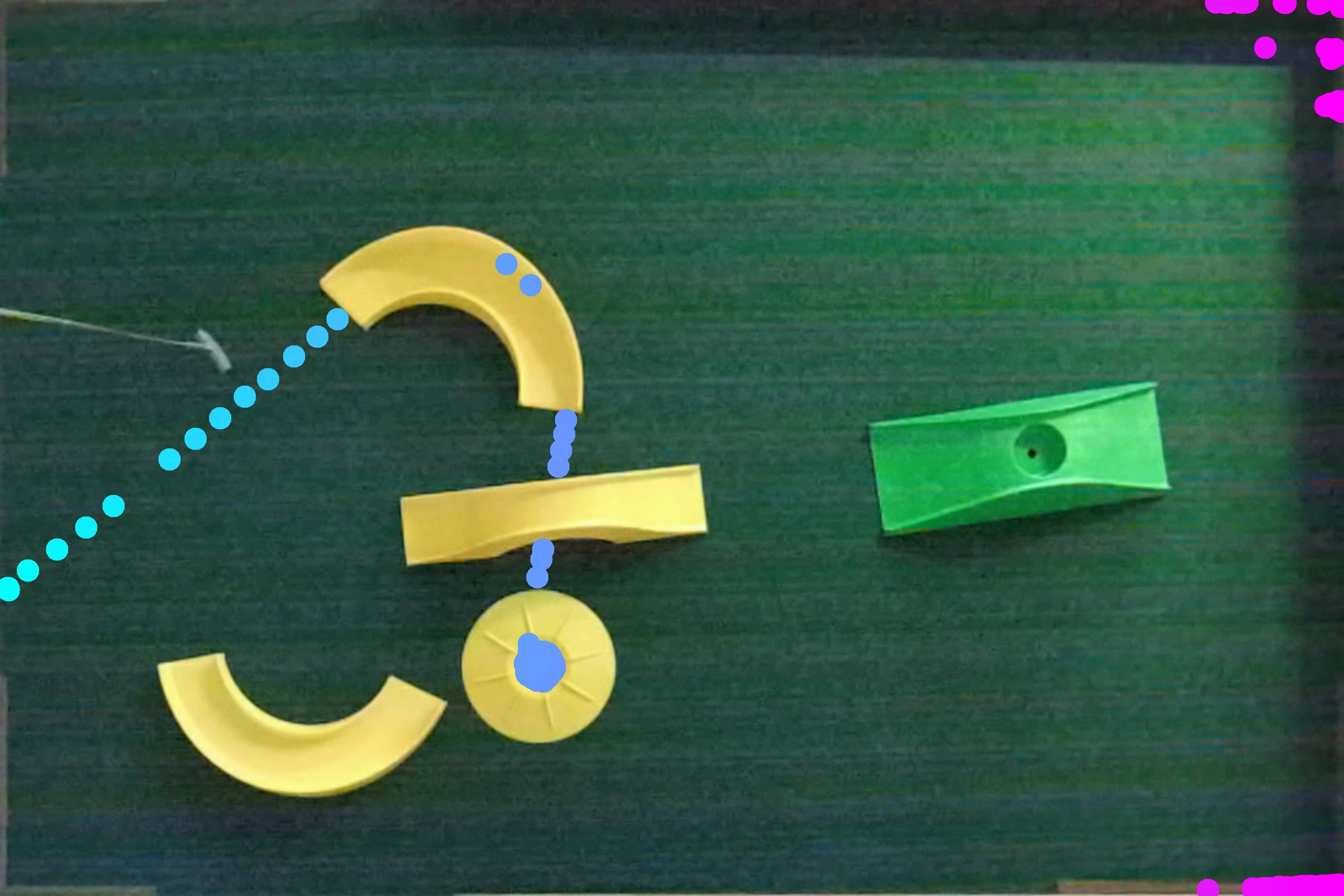}} & 
        \begin{itemize}[leftmargin=20pt,rightmargin=10pt]
            \item identify the adding the ramp in this direction won't cause obstruse in the similar path
        \end{itemize}
        \\
        \hline

        \raisebox{-0.25\height}{\includegraphics[width=0.8\linewidth]{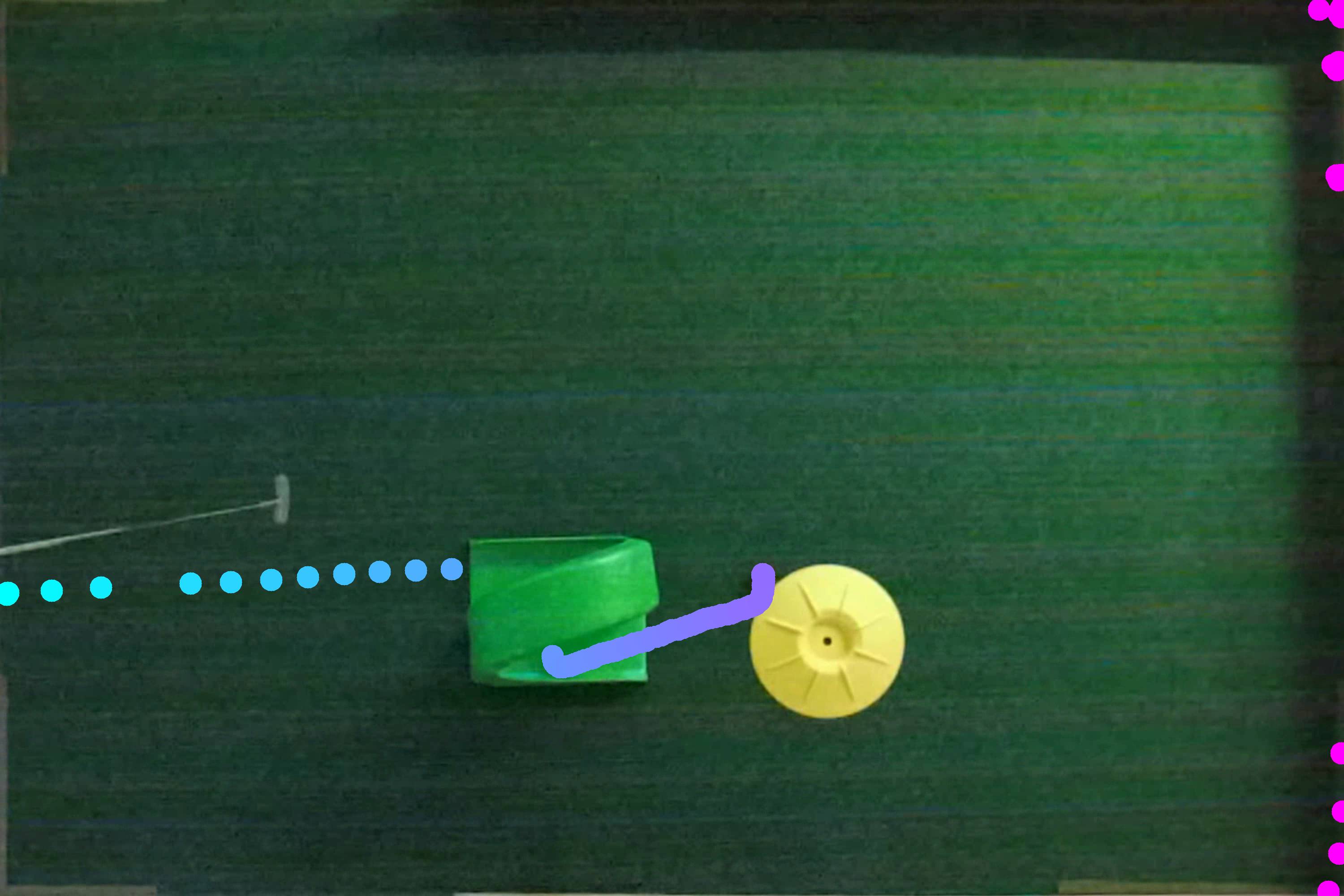}} & 
        \raisebox{-0.25\height}{\includegraphics[width=0.8\linewidth]{Figures/comparison/61_d9739d4b-4918-446f-ab56-7f2e45b63da5_transformed_rgb_trajectory_.jpg}} & 
        \begin{itemize}[leftmargin=20pt,rightmargin=10pt]
            \item Identify that the robot's speed is insufficient to hit an object after the loop
            \item Adding a curve could go around the impossible obstacle
        \end{itemize}
        \\
        \hline
    \end{tabular}
    \vspace{2mm}
    \caption{\textbf{Underlying reflective equilibrium reasoning challenges in modifying courts.}}
    \label{apptable:modifyingcourt}
\end{table}

\clearpage
\section{More Experimental Results}\label{ap:moreresults}
\begin{figure}[H]
    \centering
    \setlength{\tabcolsep}{2pt}
    \begin{tabular}{cccc}
        \includegraphics[width=0.24\textwidth]{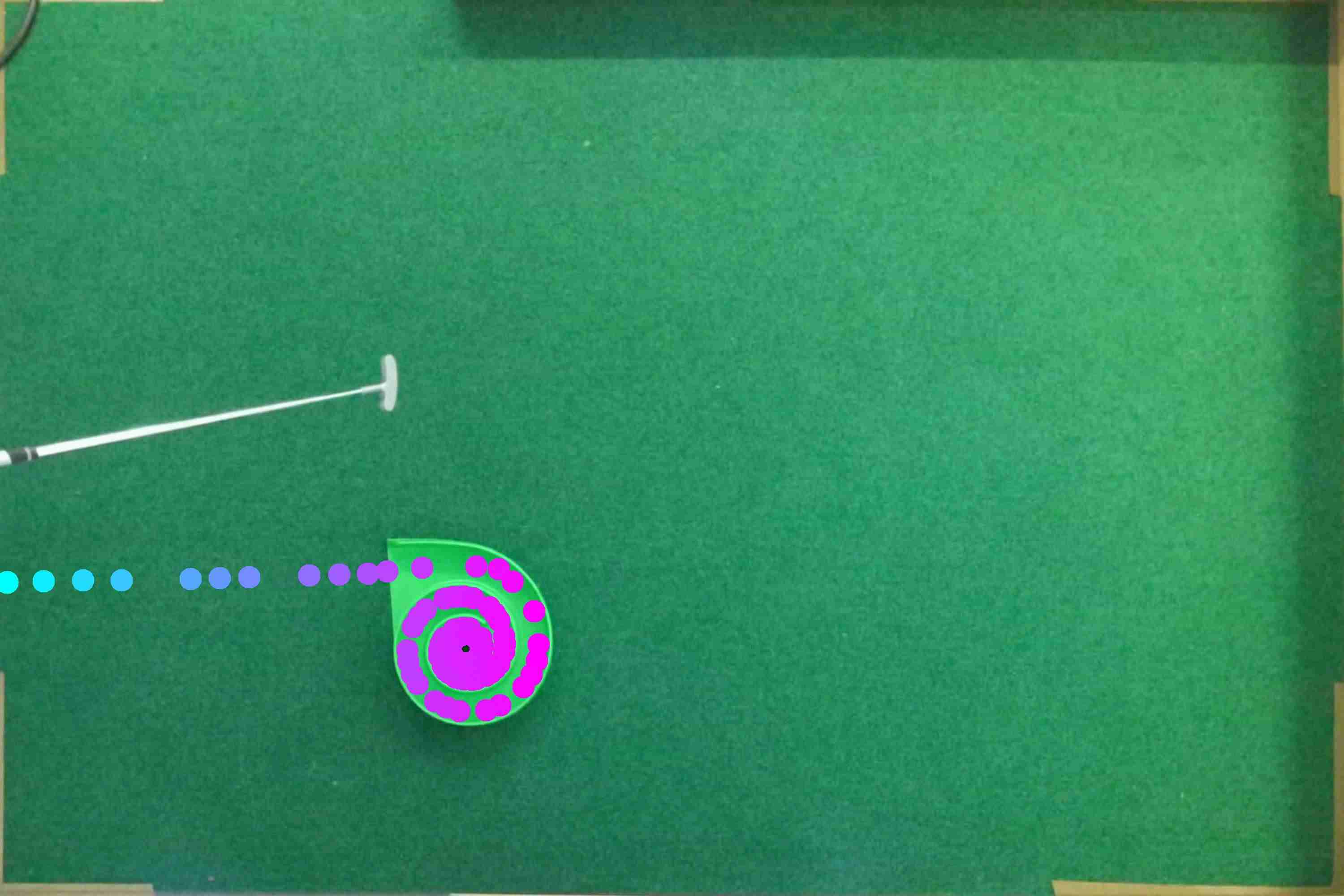} &
        \includegraphics[width=0.24\textwidth]{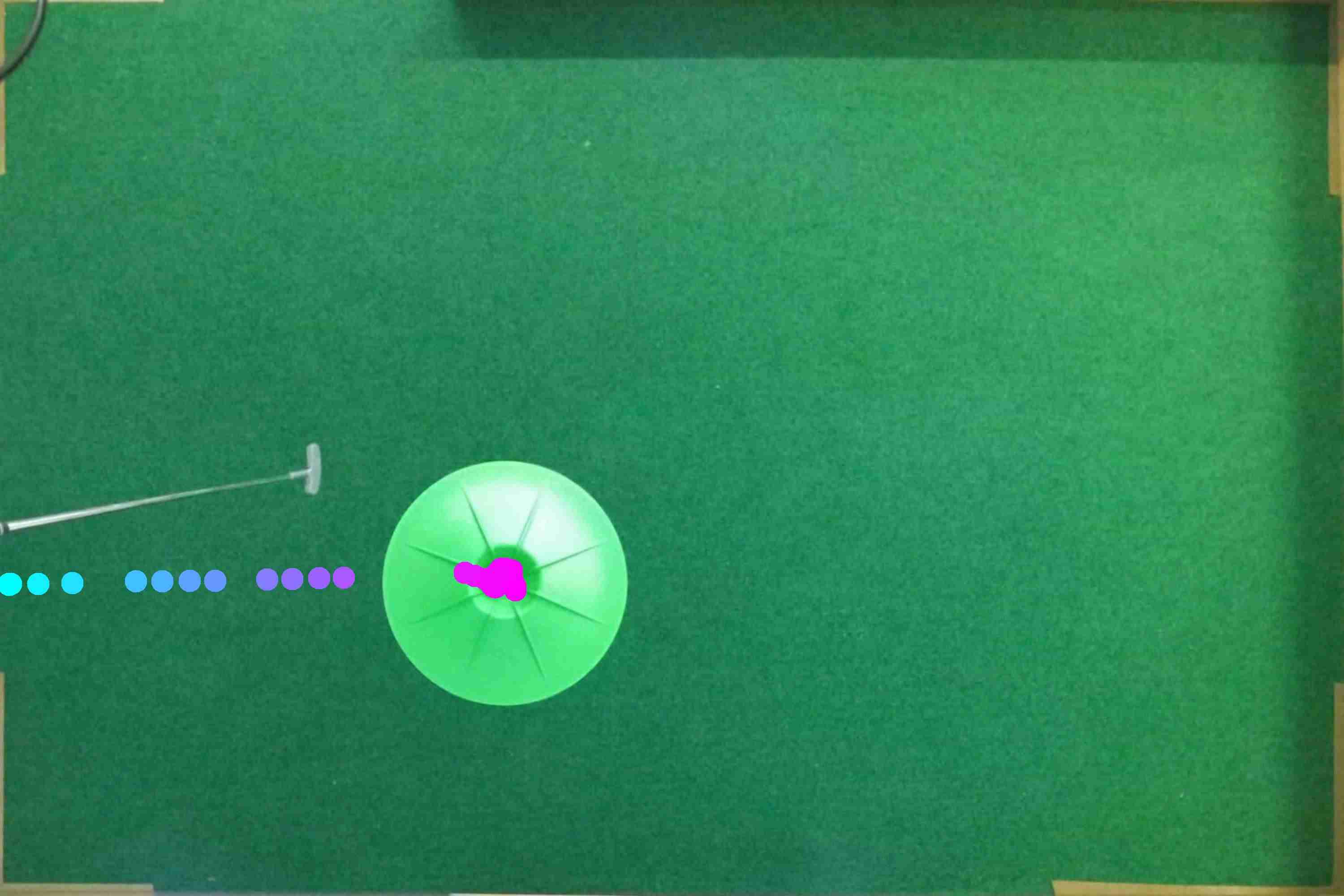} &
        \includegraphics[width=0.24\textwidth]{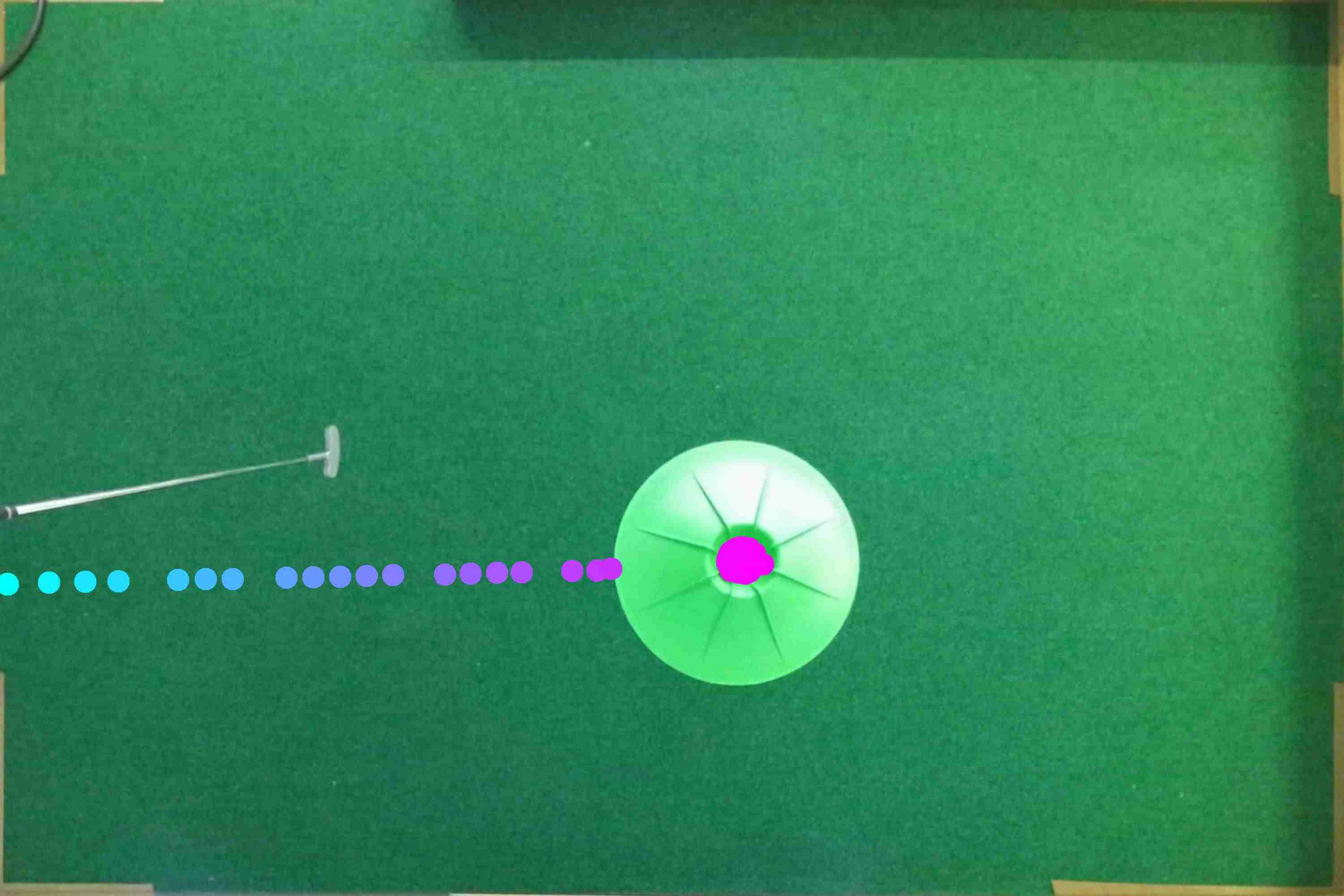} &
        \includegraphics[width=0.24\textwidth]{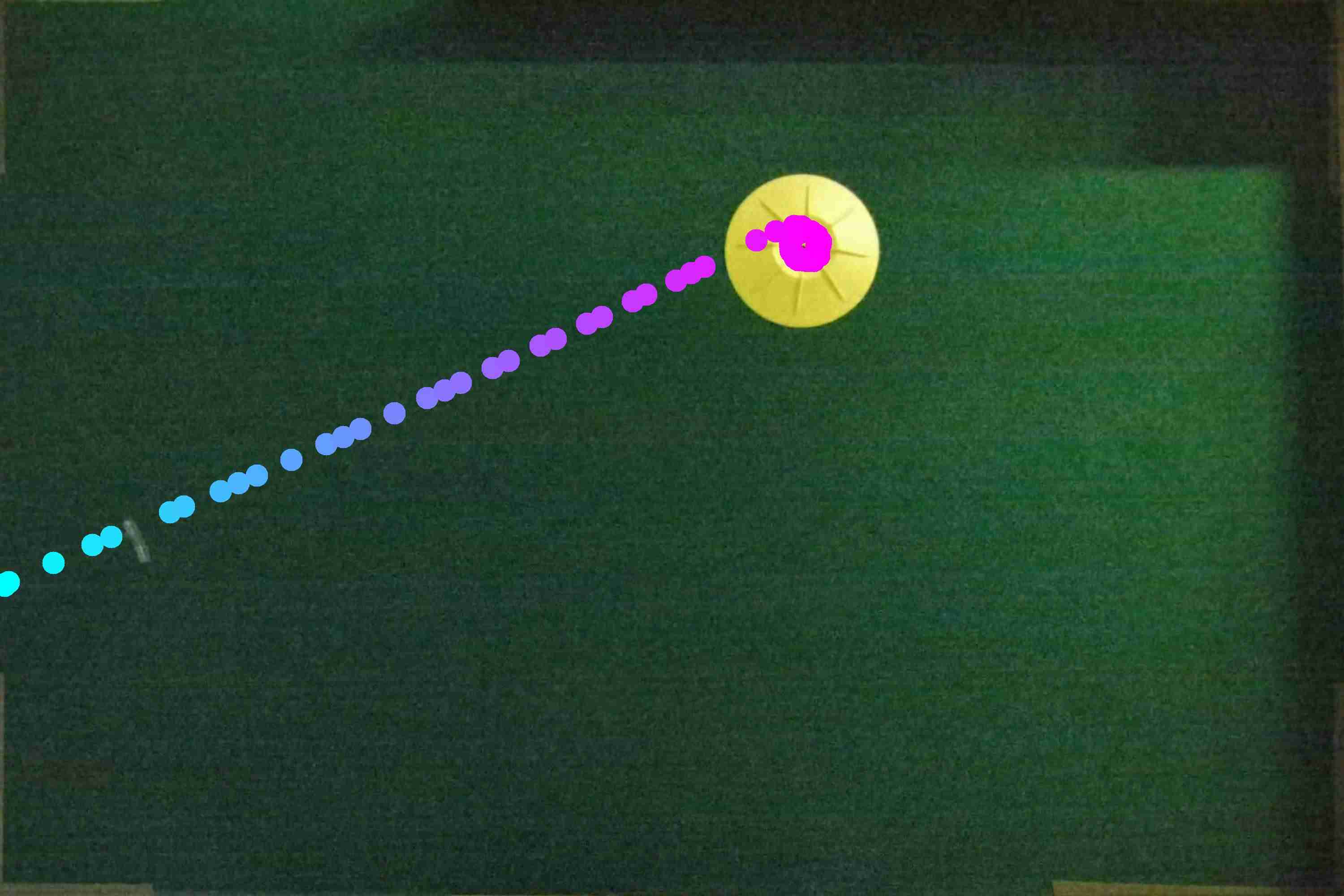} \\
        \includegraphics[width=0.24\textwidth]{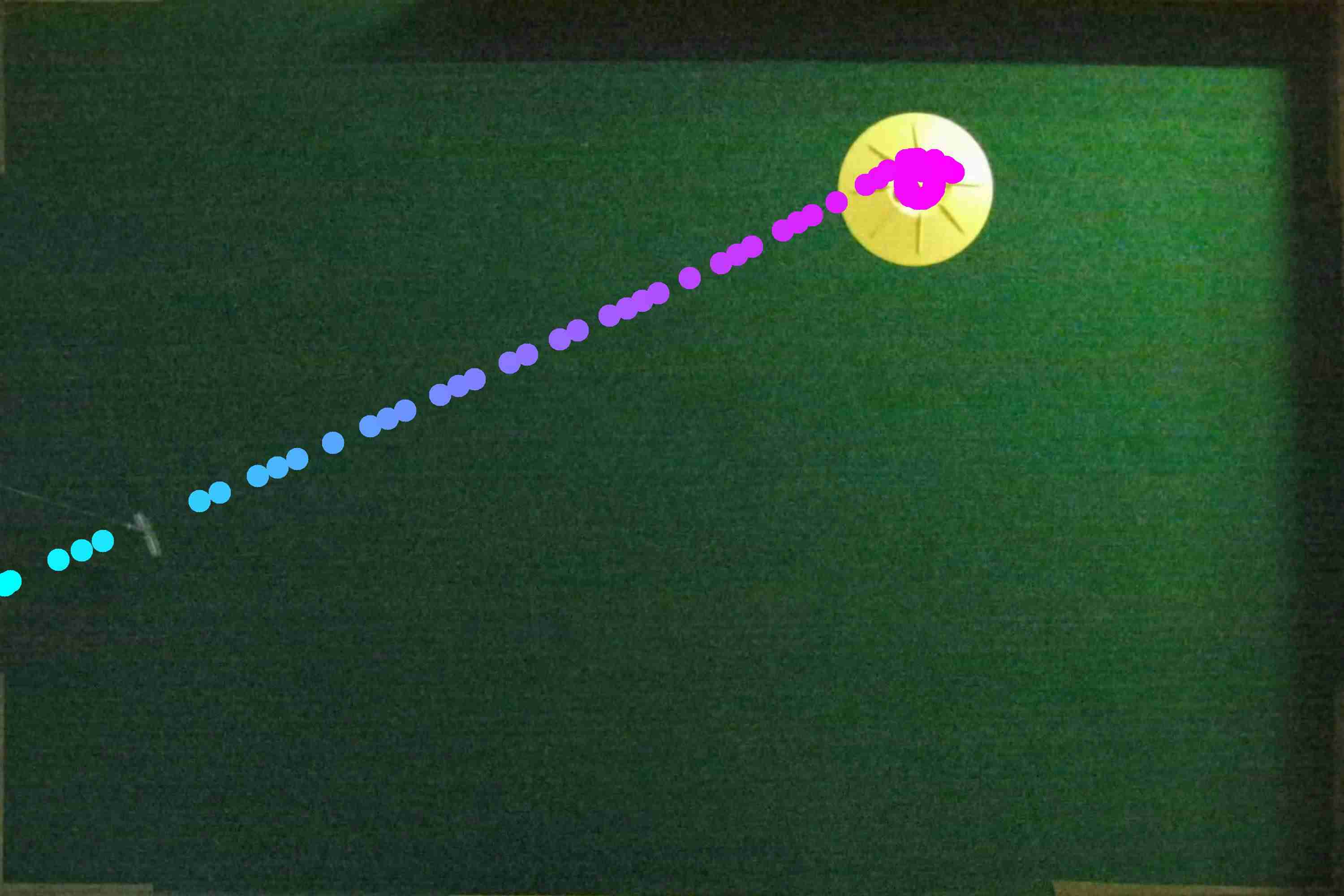} &
        \includegraphics[width=0.24\textwidth]{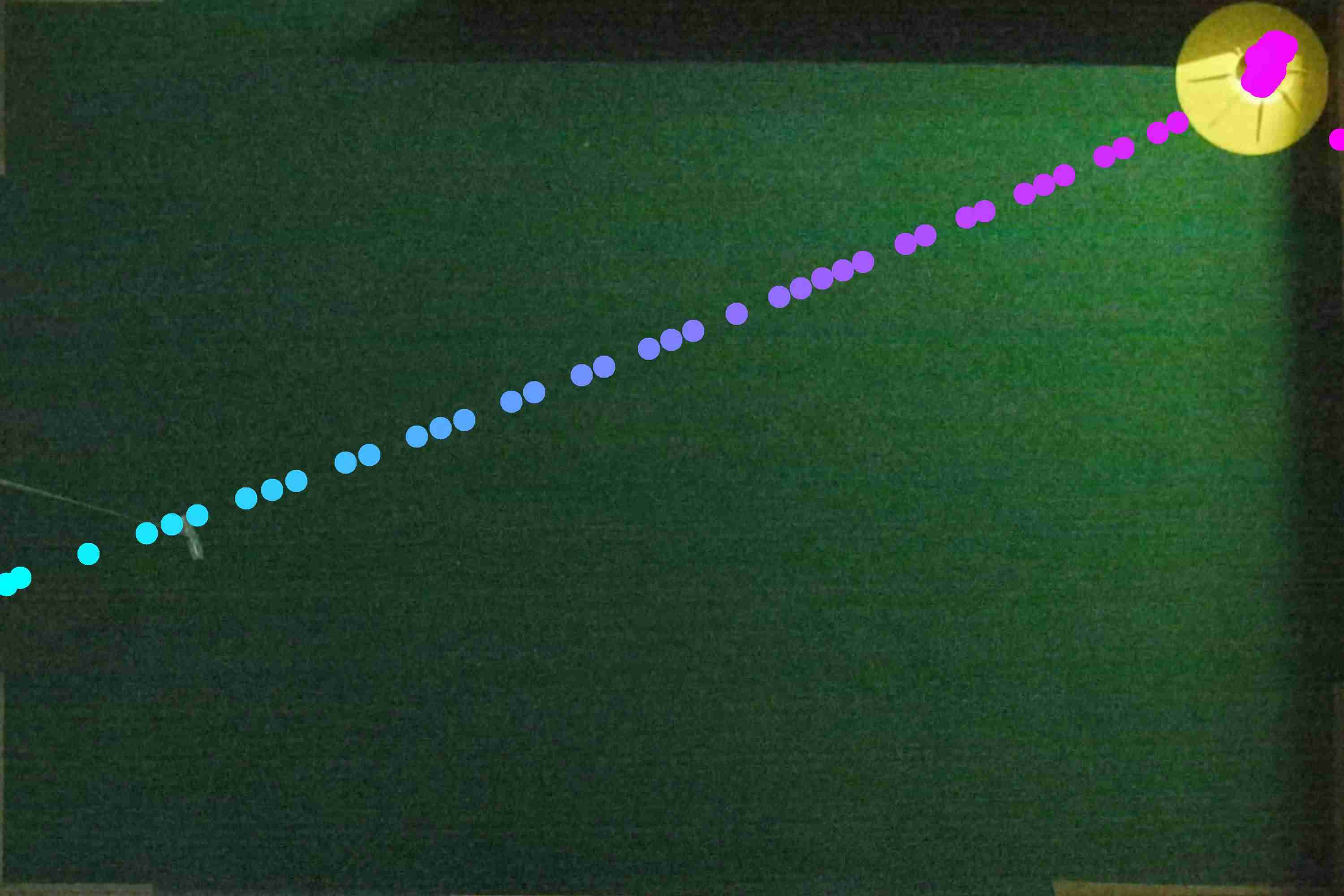} &
        \includegraphics[width=0.24\textwidth]{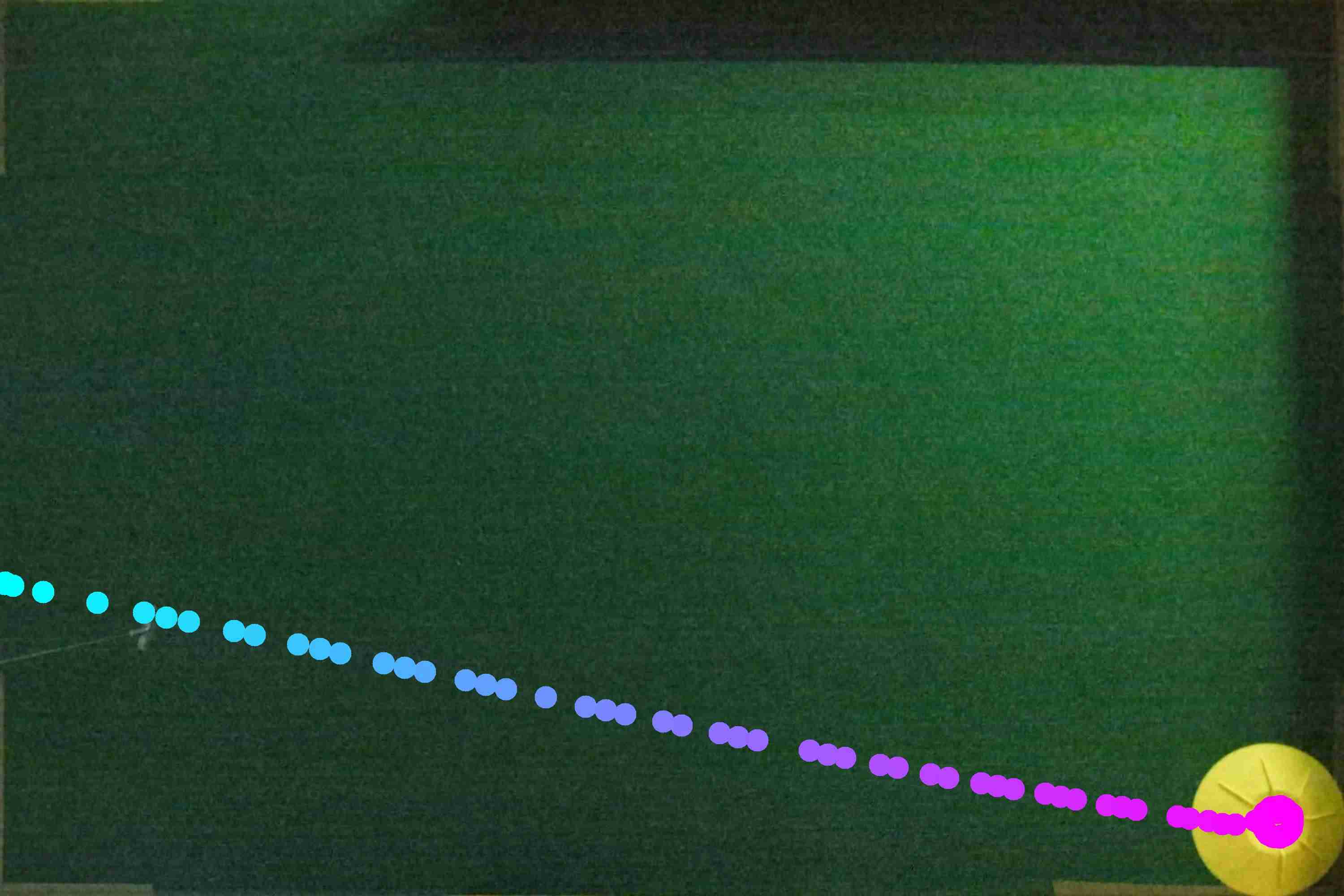} &
        \includegraphics[width=0.24\textwidth]{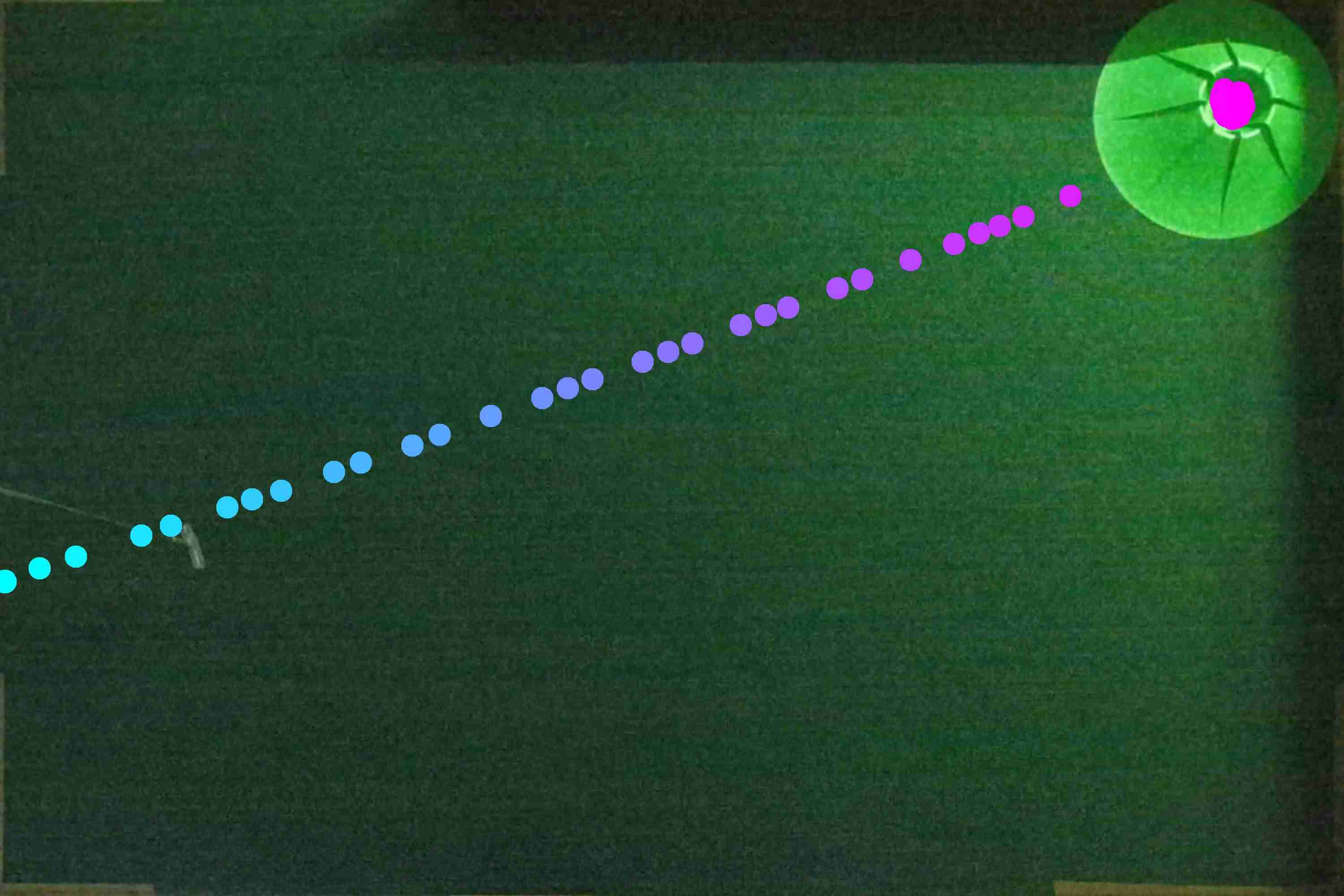} \\
        \includegraphics[width=0.24\textwidth]{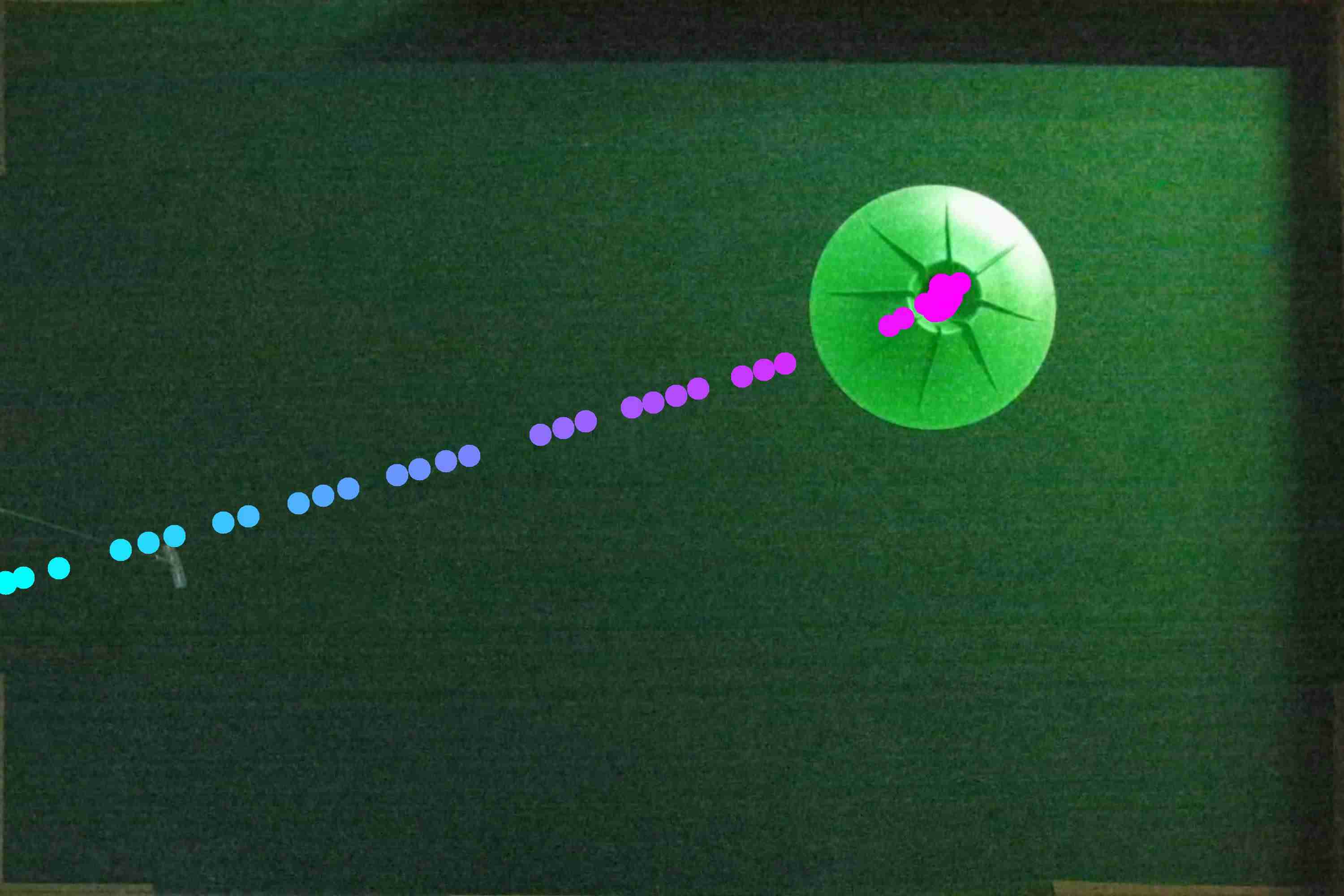} &
        \includegraphics[width=0.24\textwidth]{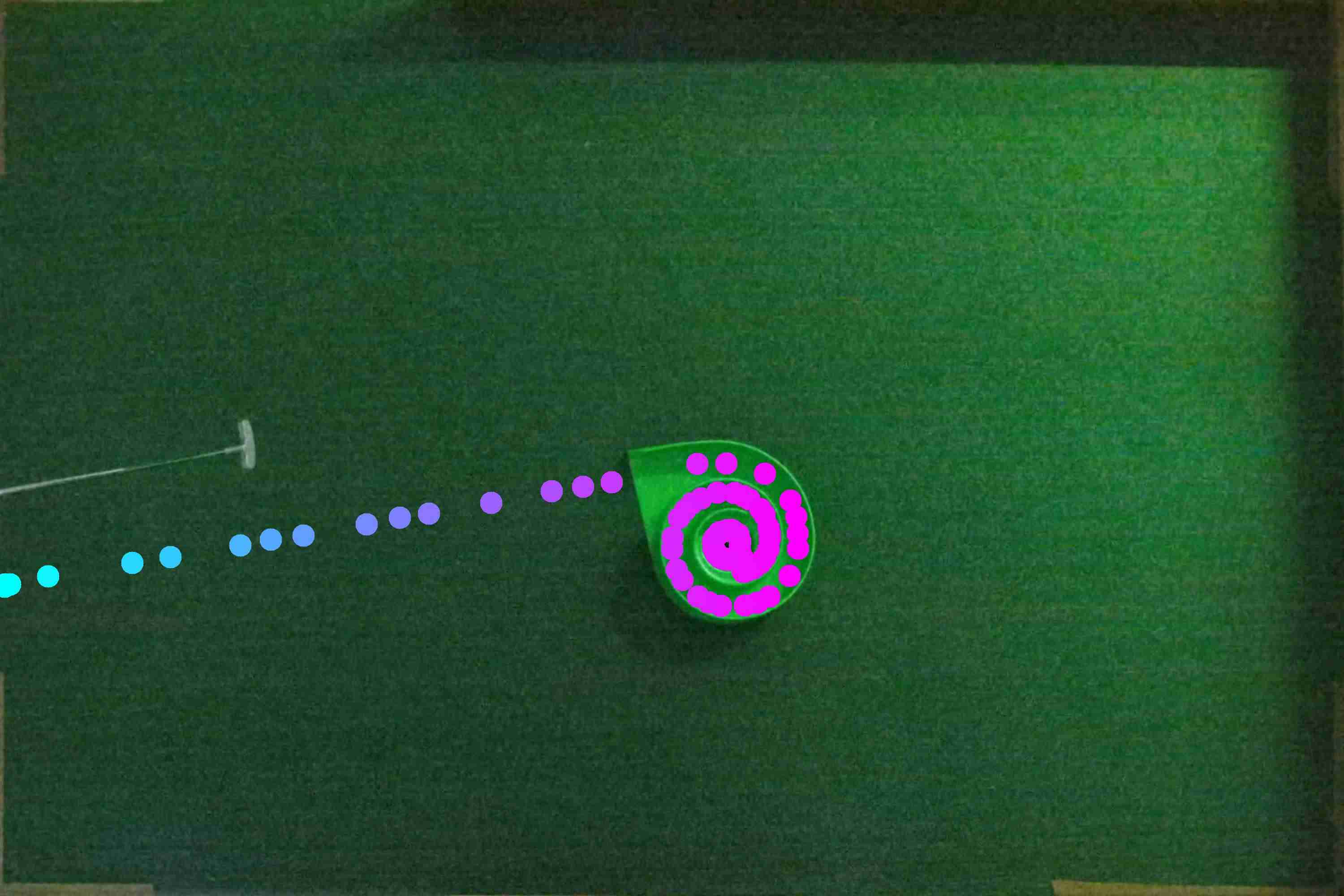} & 
    \end{tabular}
    \caption{\textbf{Hitting trajectories in simple courts.}}
    \label{appfig:simplecourts}
\end{figure}

\begin{figure}[H]
    \centering
    \setlength{\tabcolsep}{2pt}  
    \begin{tabular}{cccc}
        \includegraphics[width=0.24\textwidth]{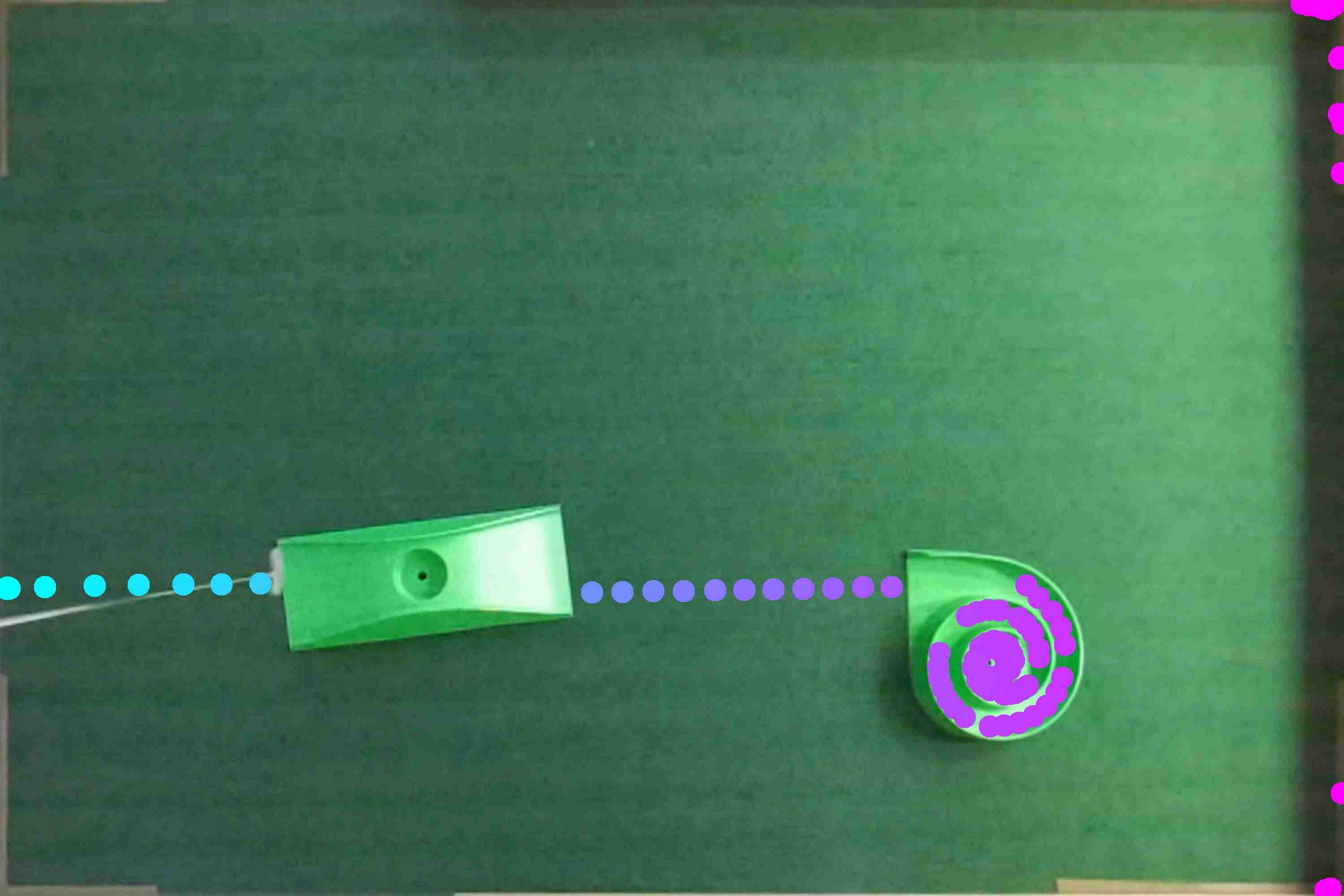} &
        \includegraphics[width=0.24\textwidth]{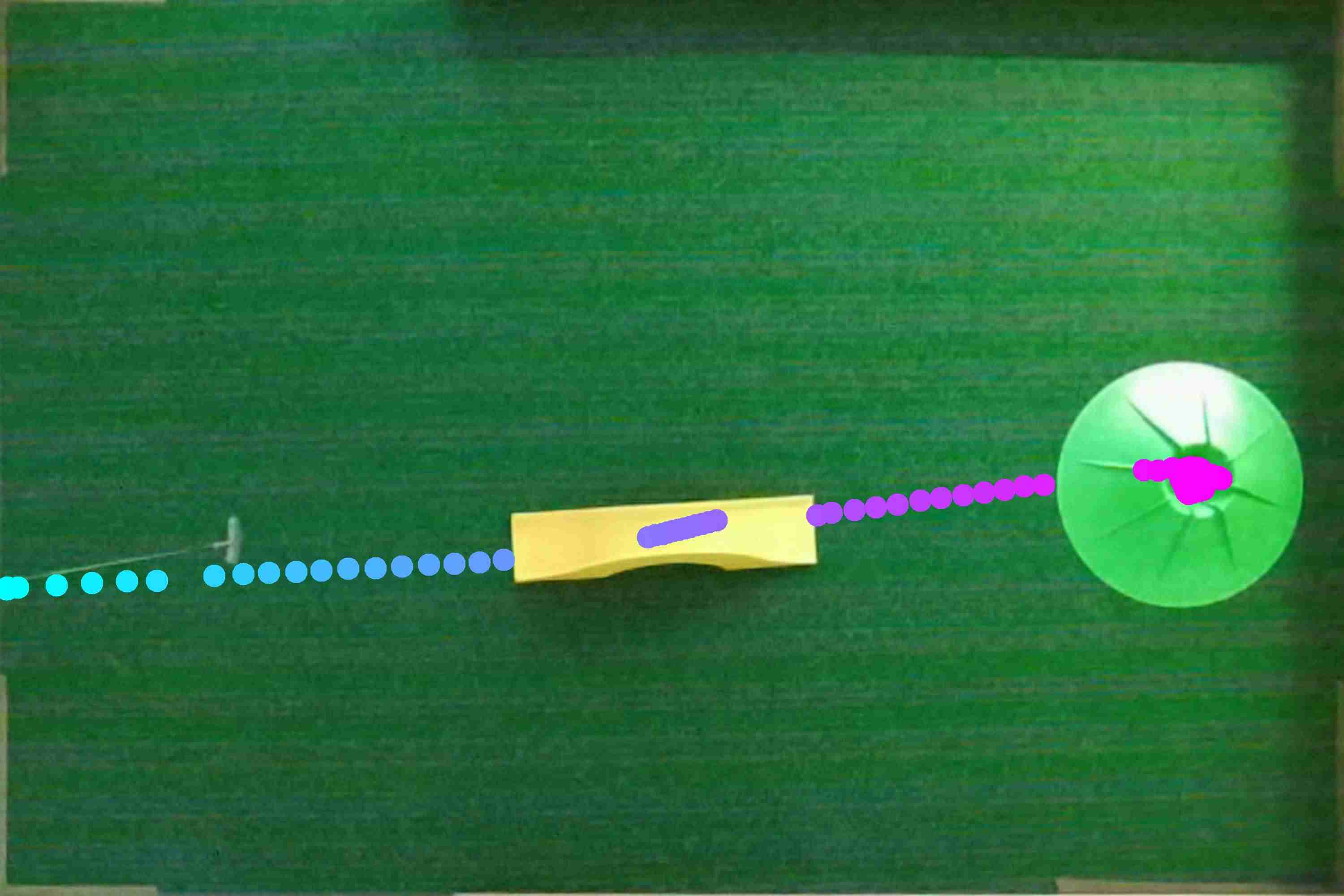} &
        \includegraphics[width=0.24\textwidth]{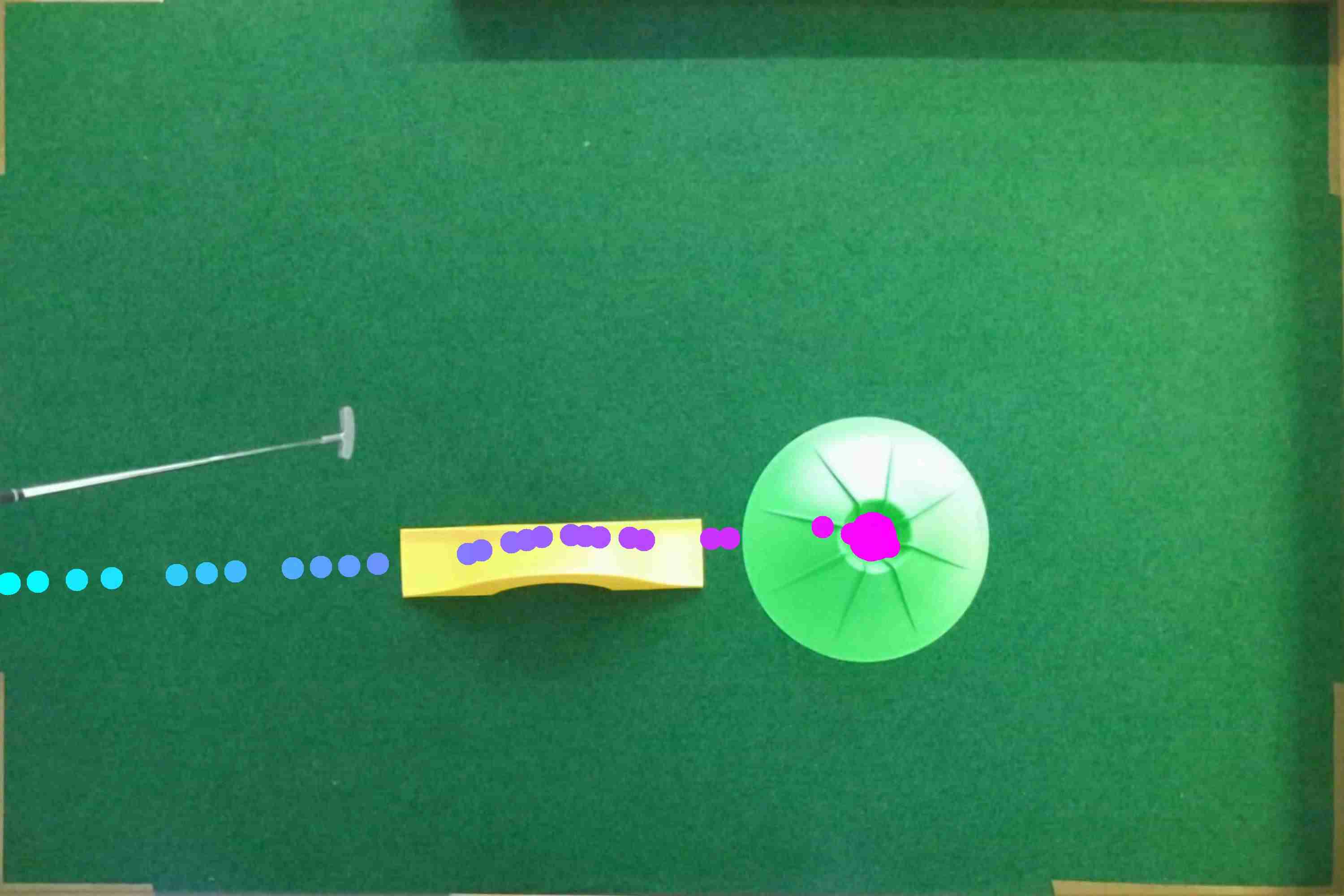} &
        \includegraphics[width=0.24\textwidth]{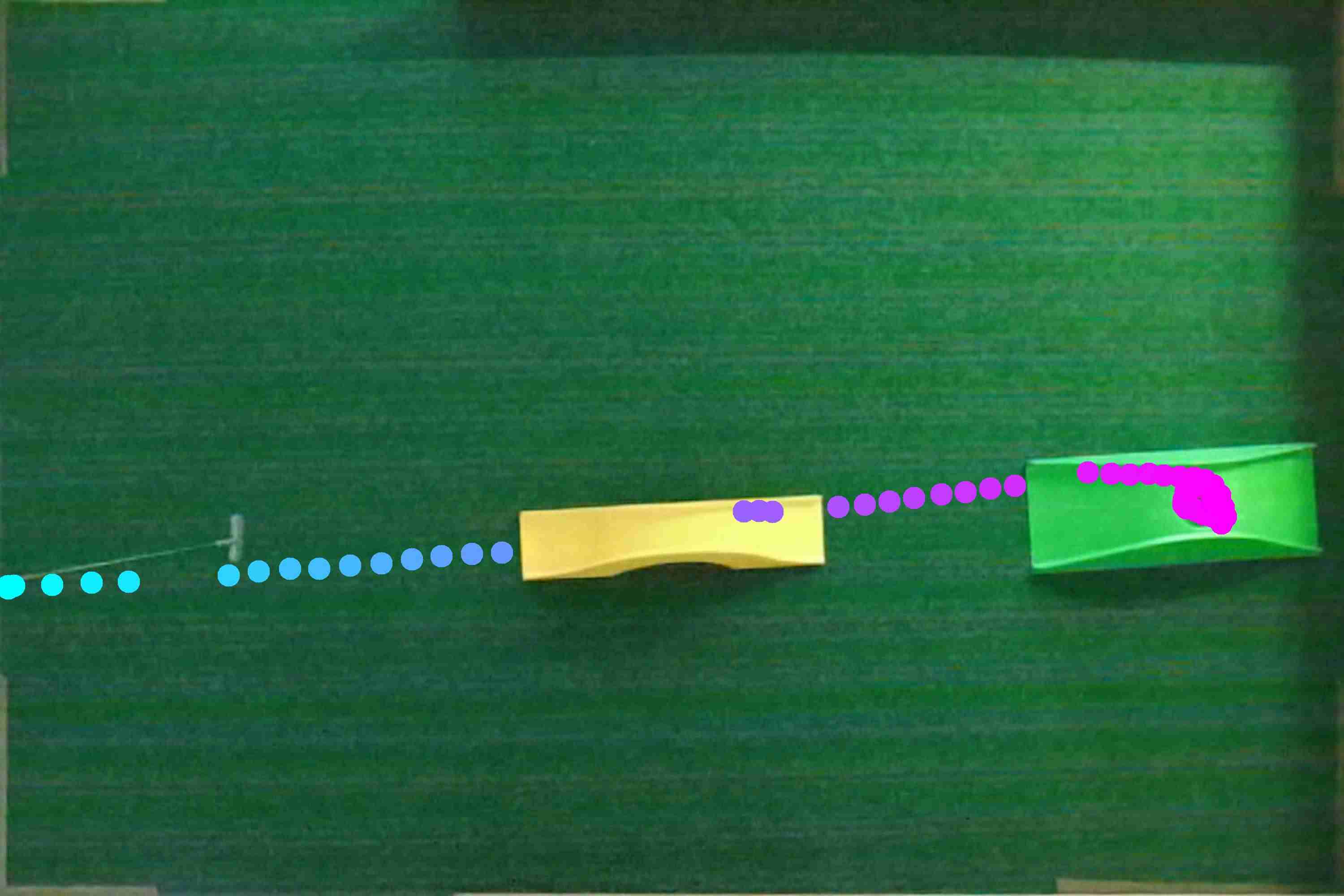} \\
        \includegraphics[width=0.24\textwidth]{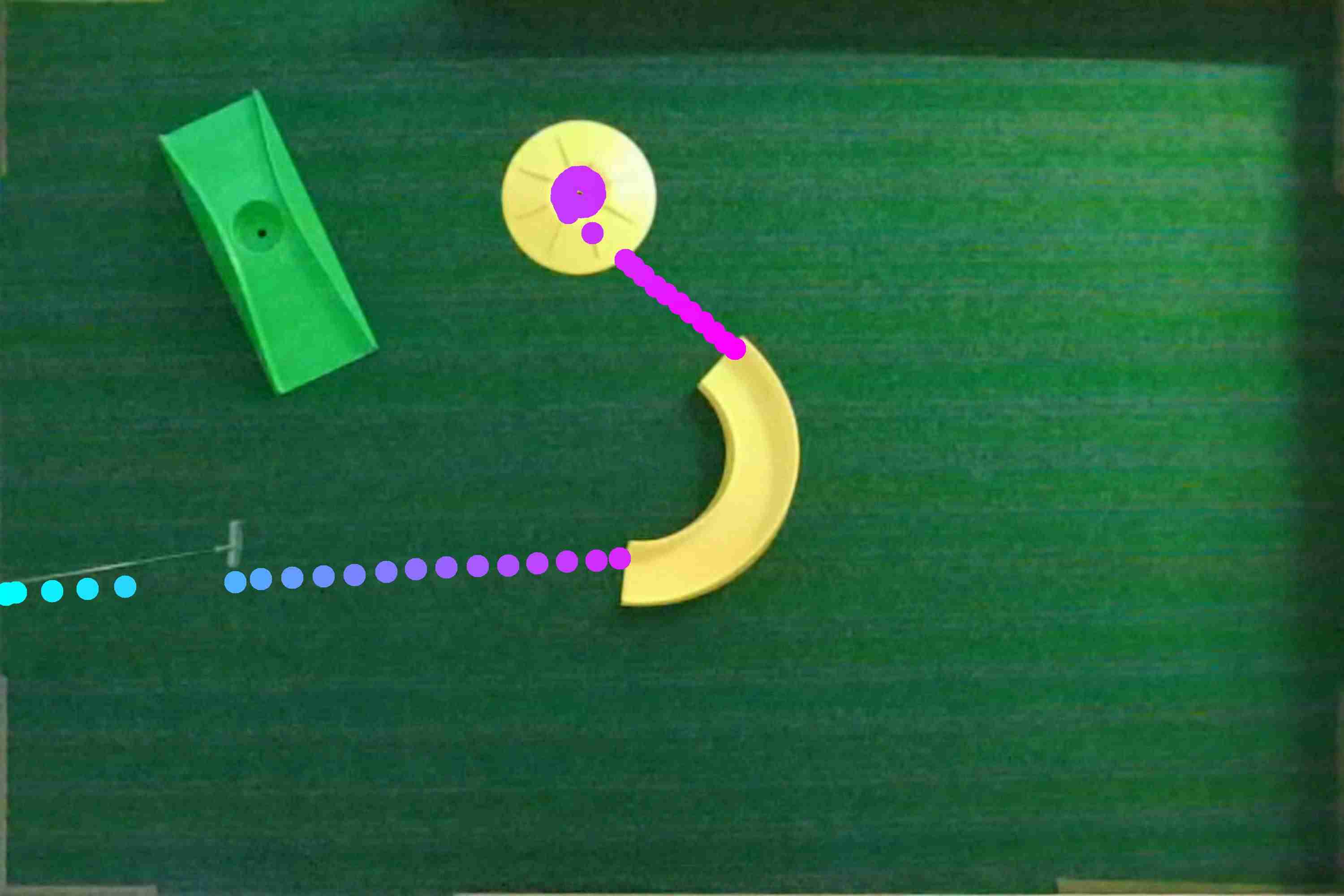} &
        \includegraphics[width=0.24\textwidth]{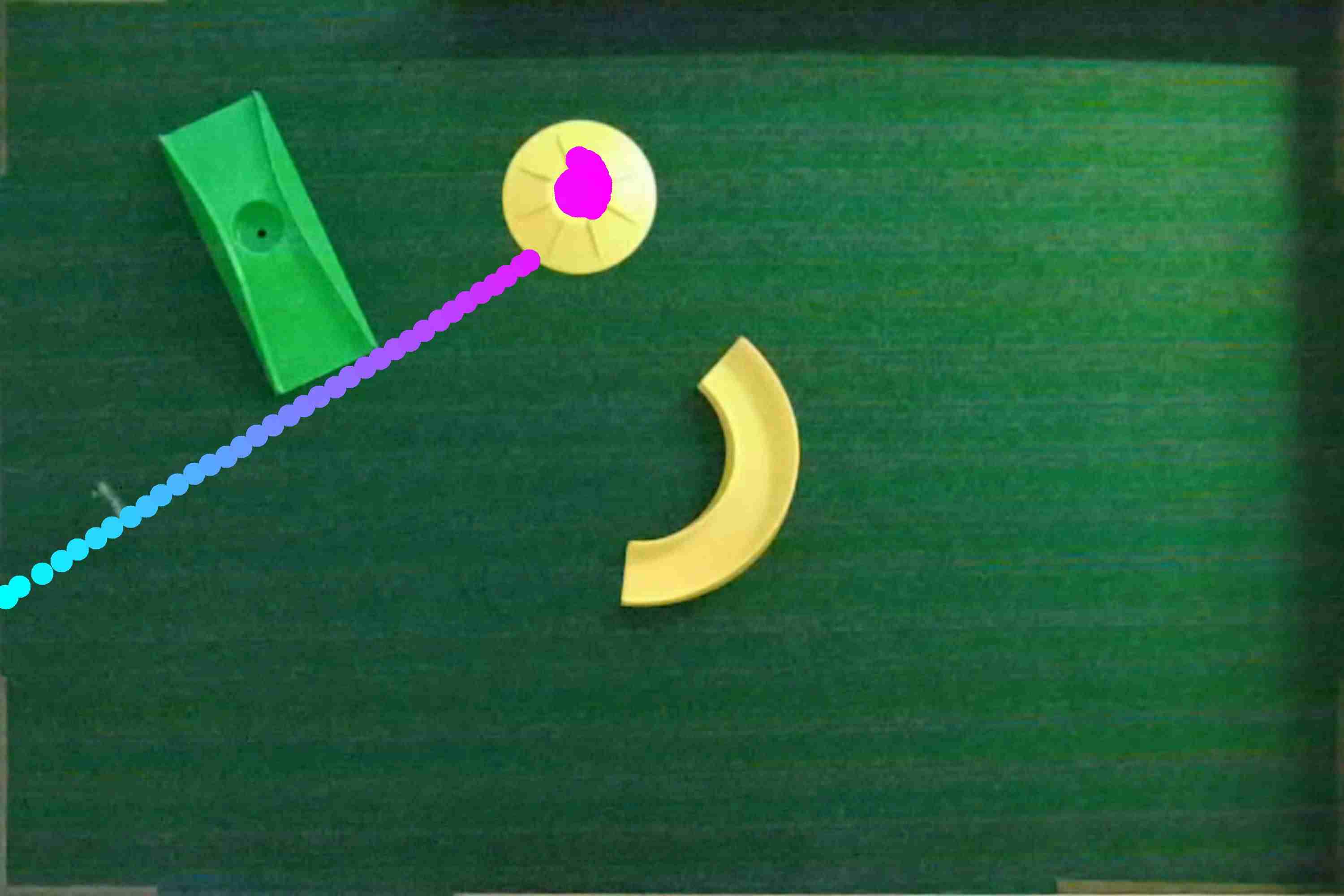} &
        \includegraphics[width=0.24\textwidth]{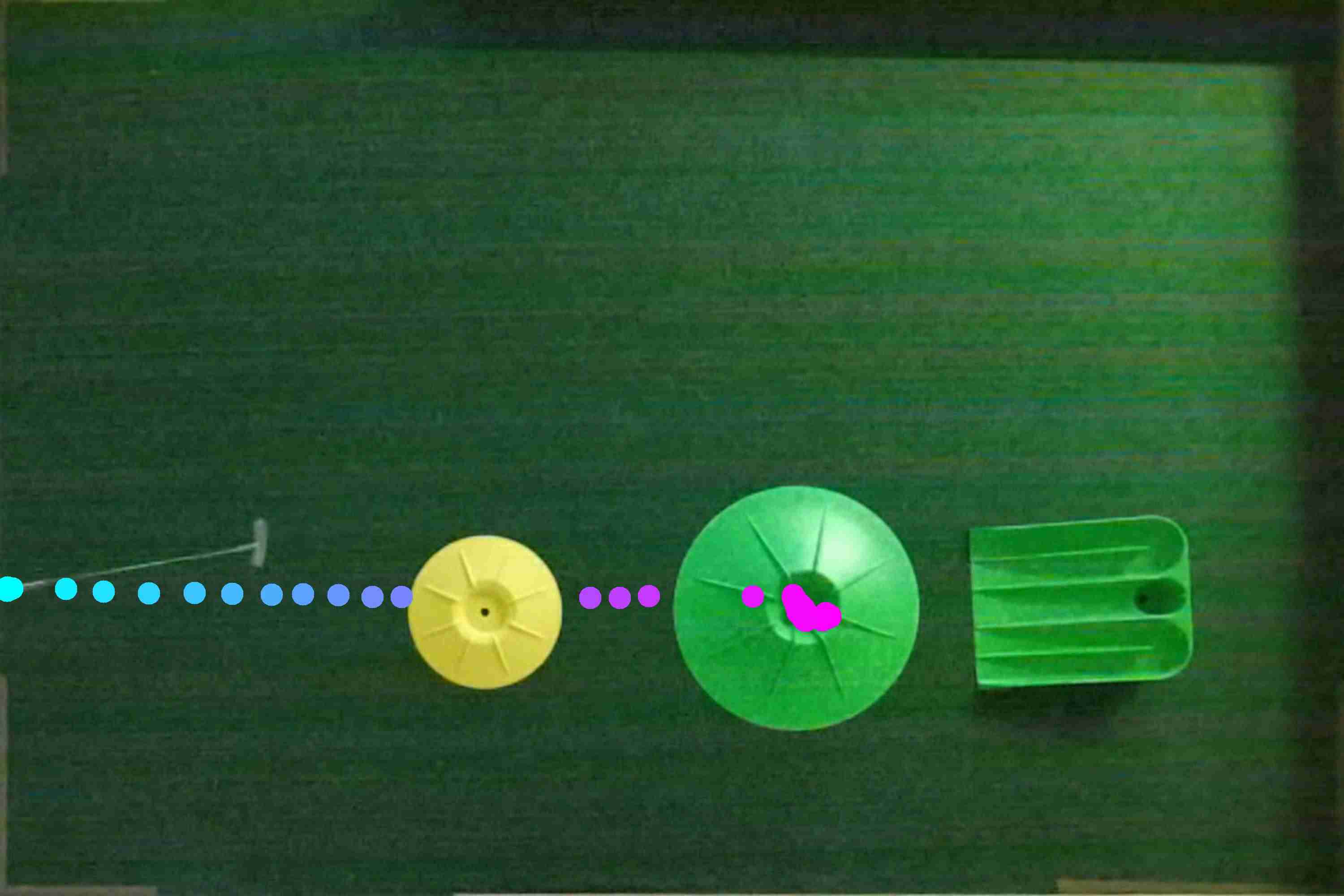} &
        \includegraphics[width=0.24\textwidth]{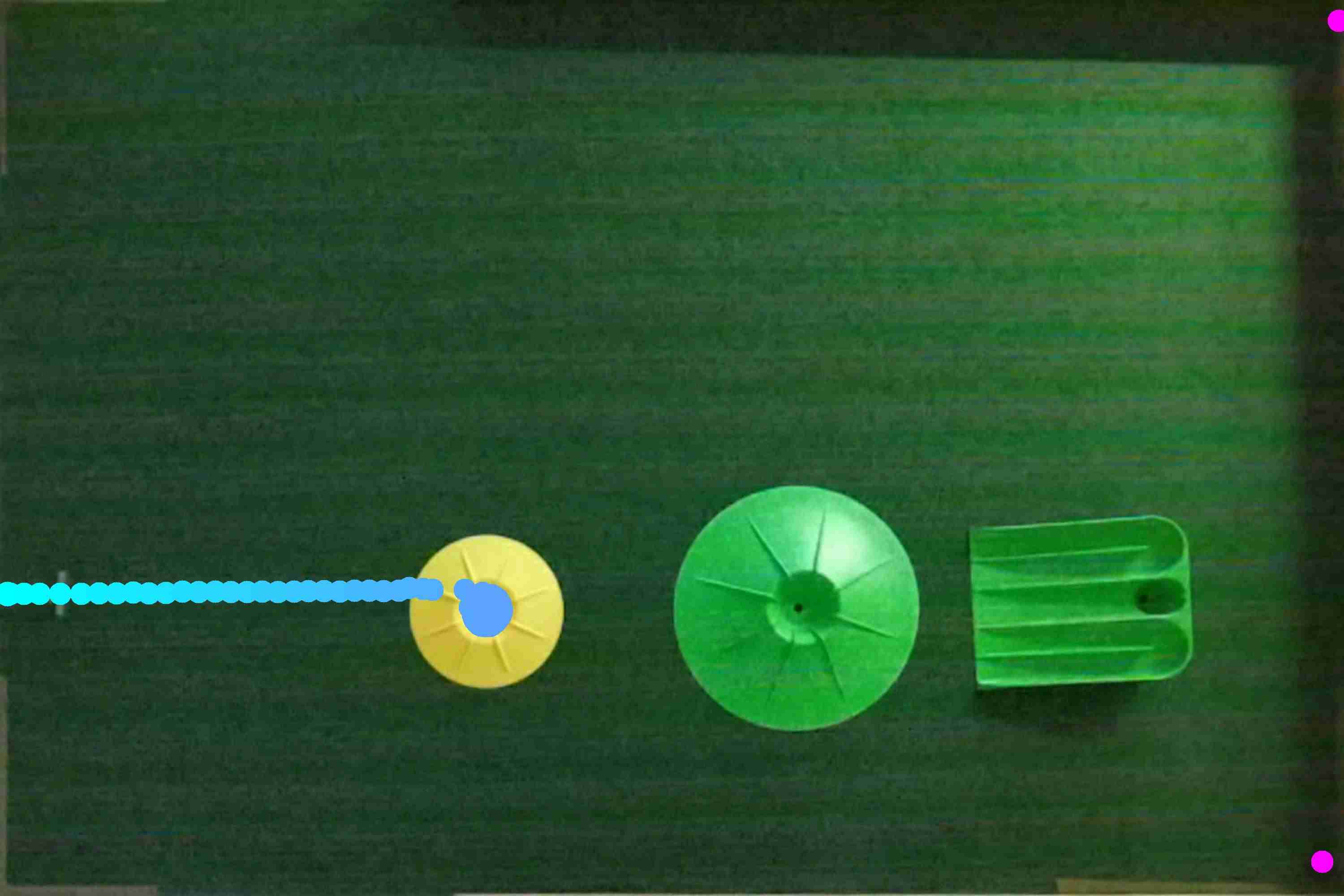}
    \end{tabular}
    \caption{\textbf{Hitting trajectories in medium courts.}}
    \label{appfig:mediumcourts}
\end{figure}

\begin{figure}[H]
    \centering
    \setlength{\tabcolsep}{2pt}  
    \begin{tabular}{cccc}
        \includegraphics[width=0.24\textwidth]{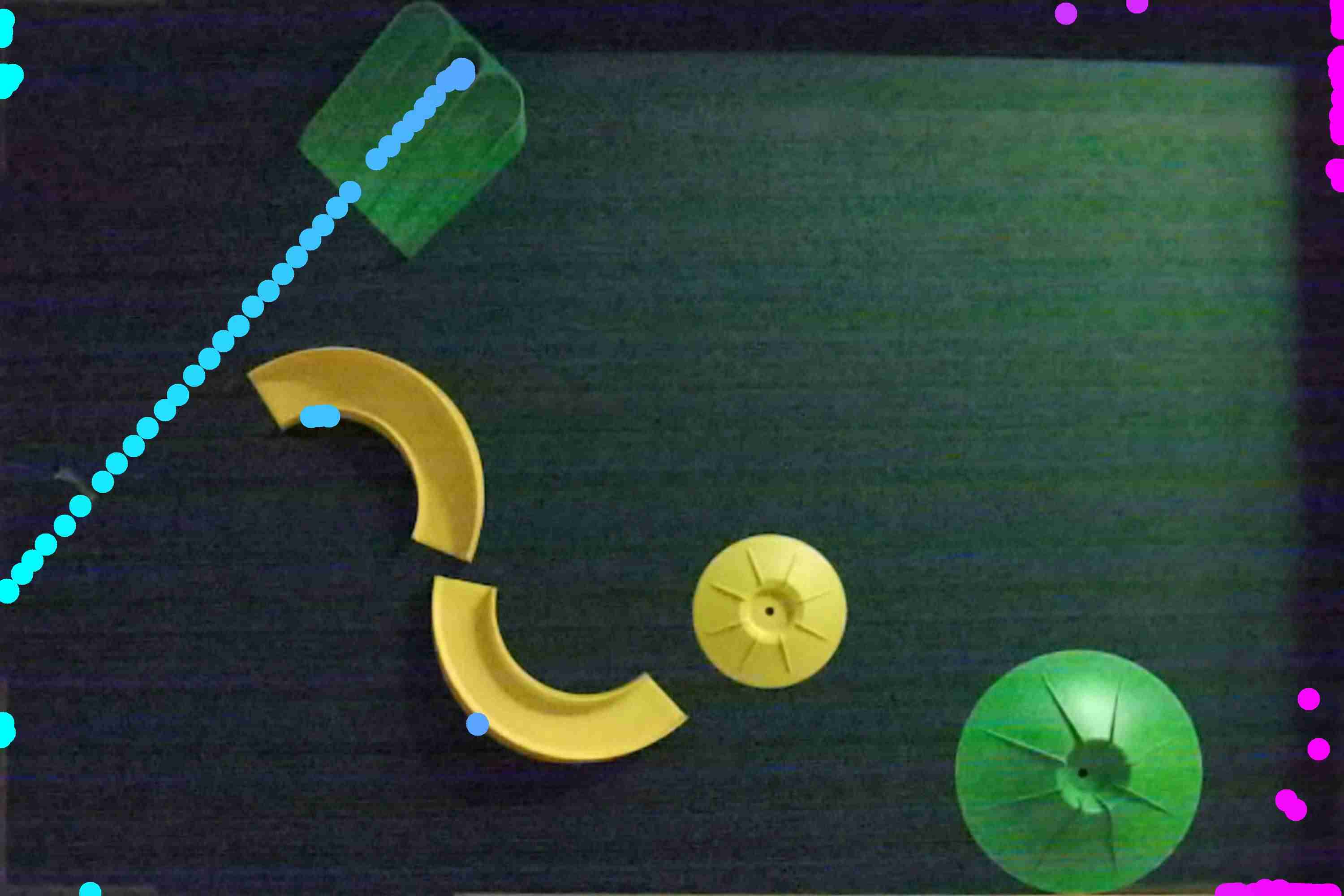} &
        \includegraphics[width=0.24\textwidth]{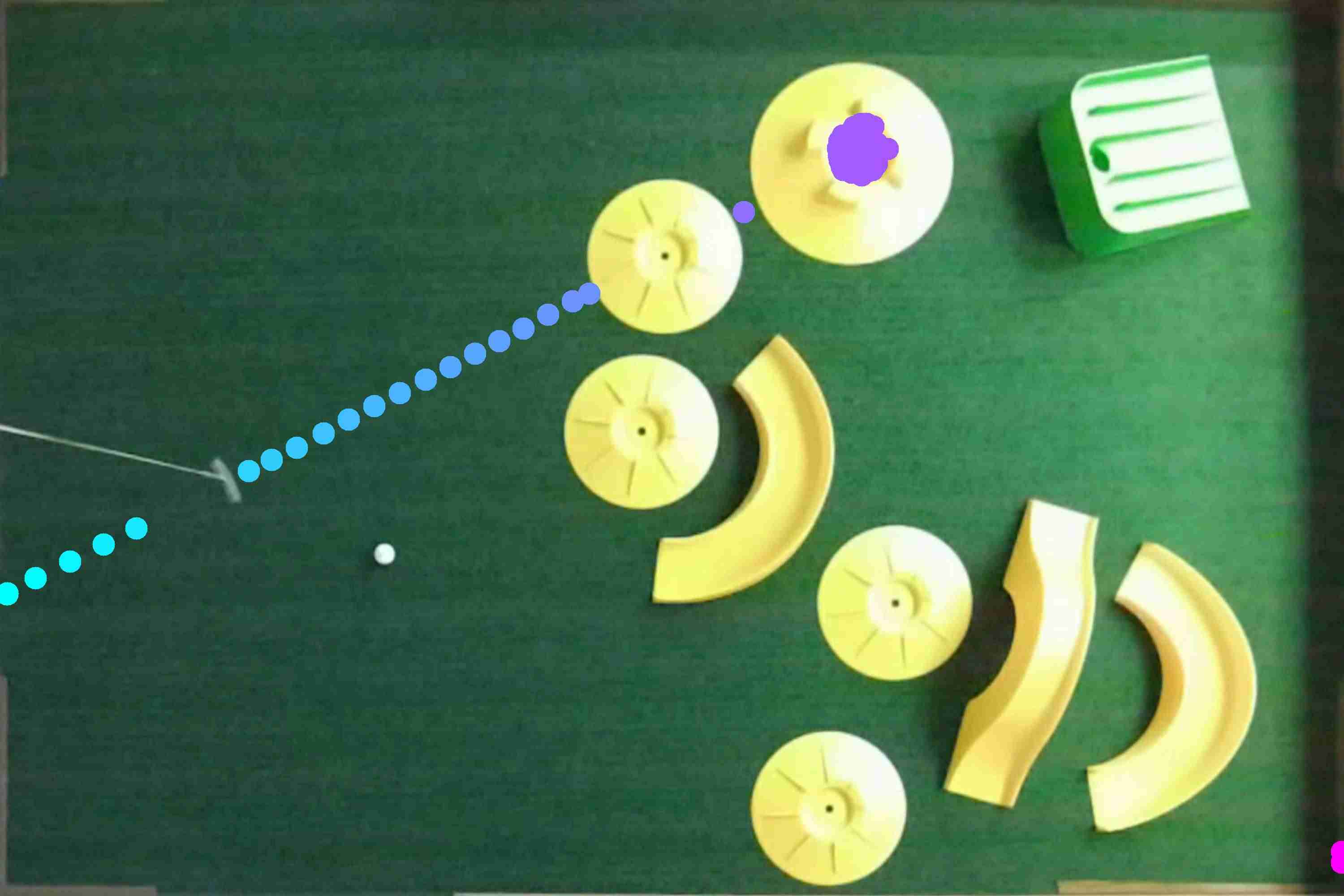} &
        \includegraphics[width=0.24\textwidth]{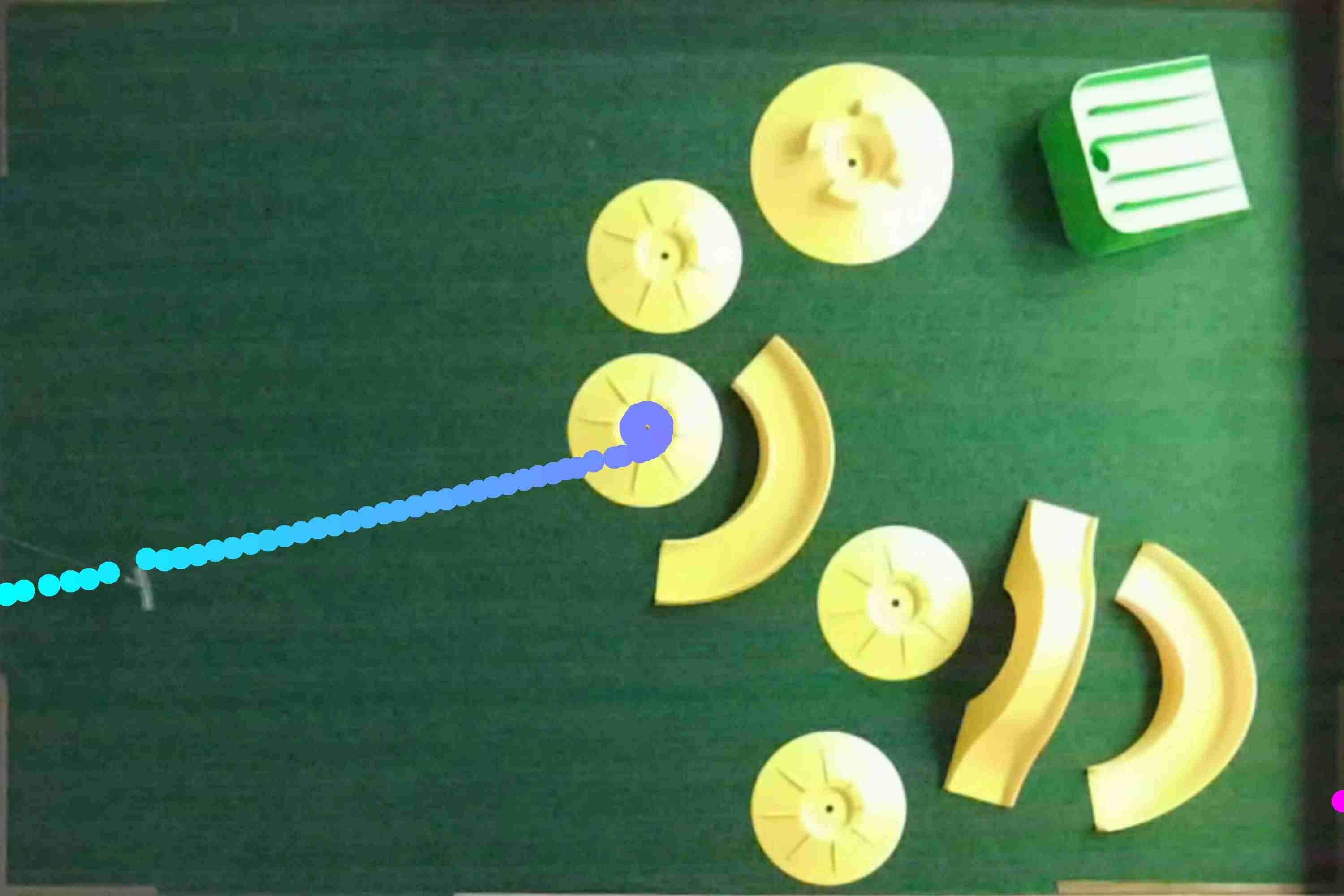} &
        \includegraphics[width=0.24\textwidth]{Figures/hard/19_f56fa4a2-3474-4acf-abf5-a6e69c90d8c9_transformed_rgb_trajectory_.jpg} \\
        \includegraphics[width=0.24\textwidth]{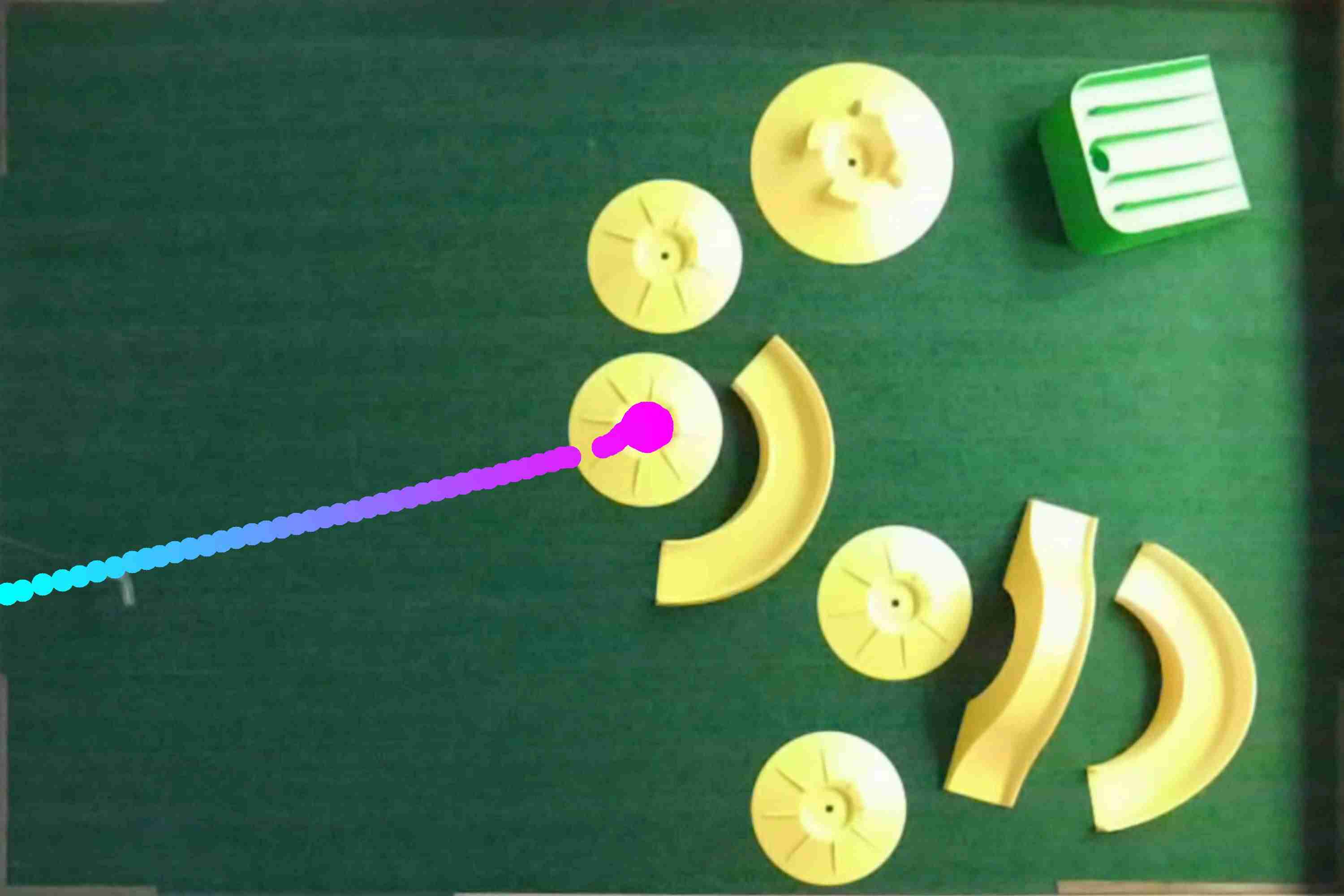} &
        \includegraphics[width=0.24\textwidth]{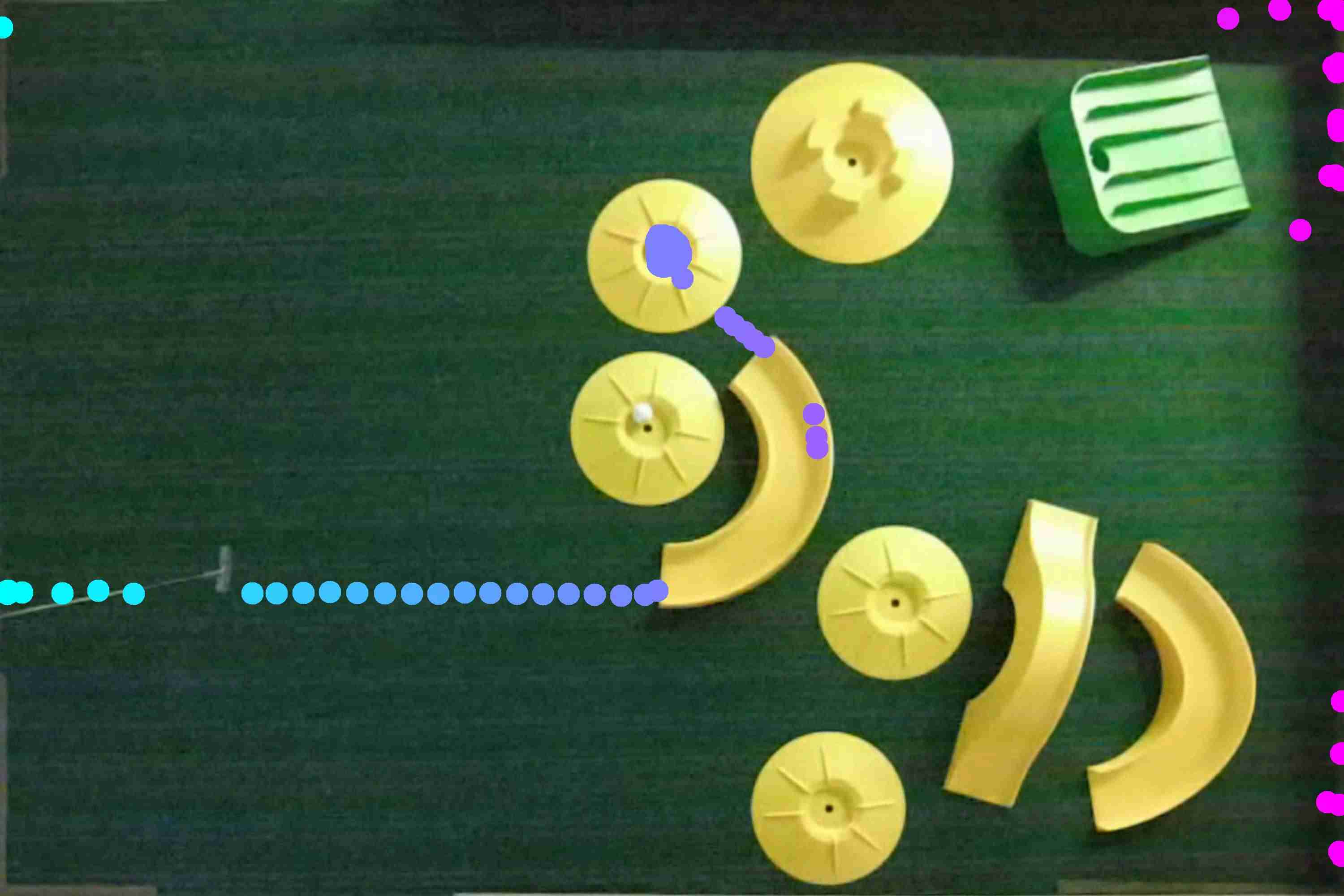} &
        \includegraphics[width=0.24\textwidth]{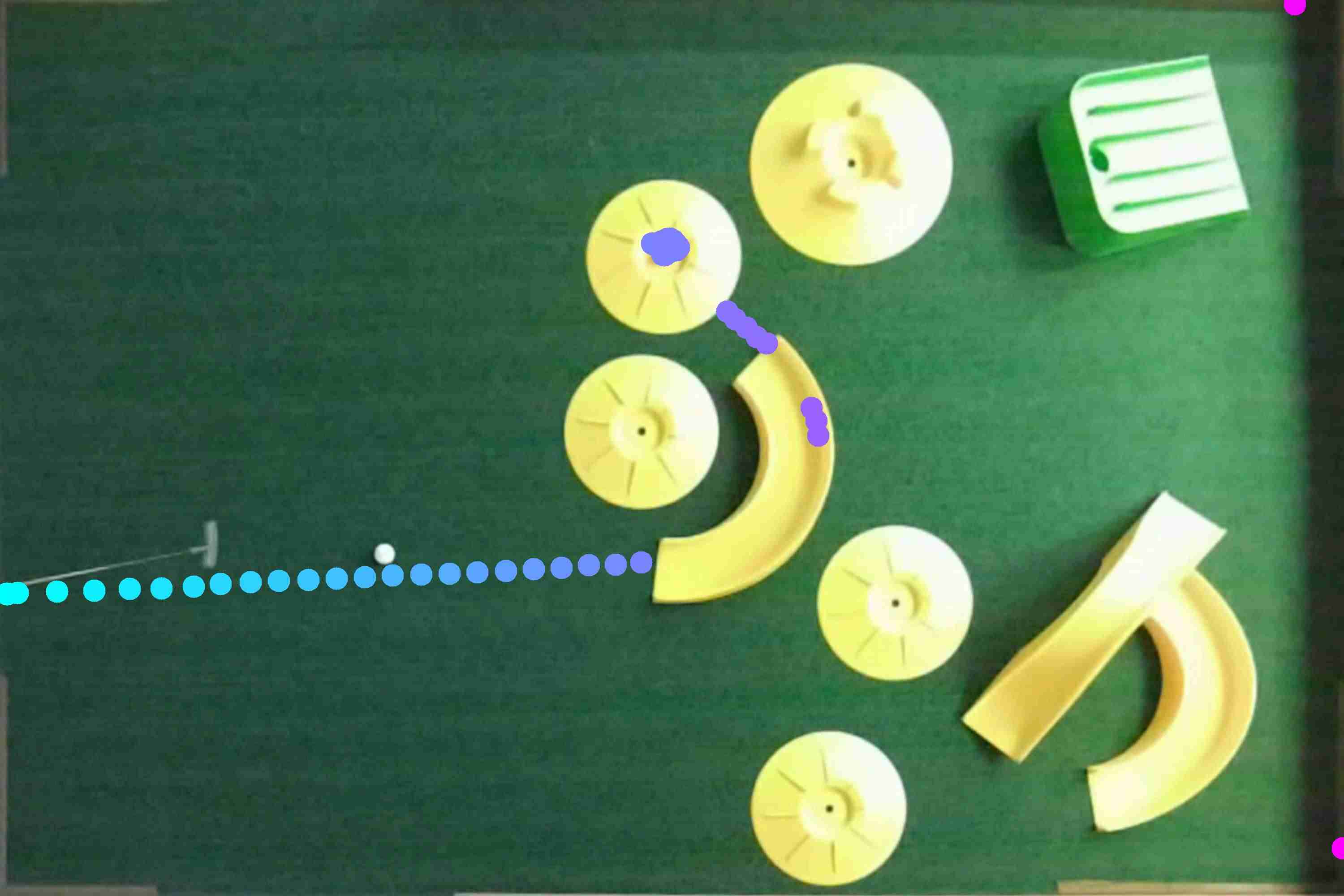} &
        \includegraphics[width=0.24\textwidth]{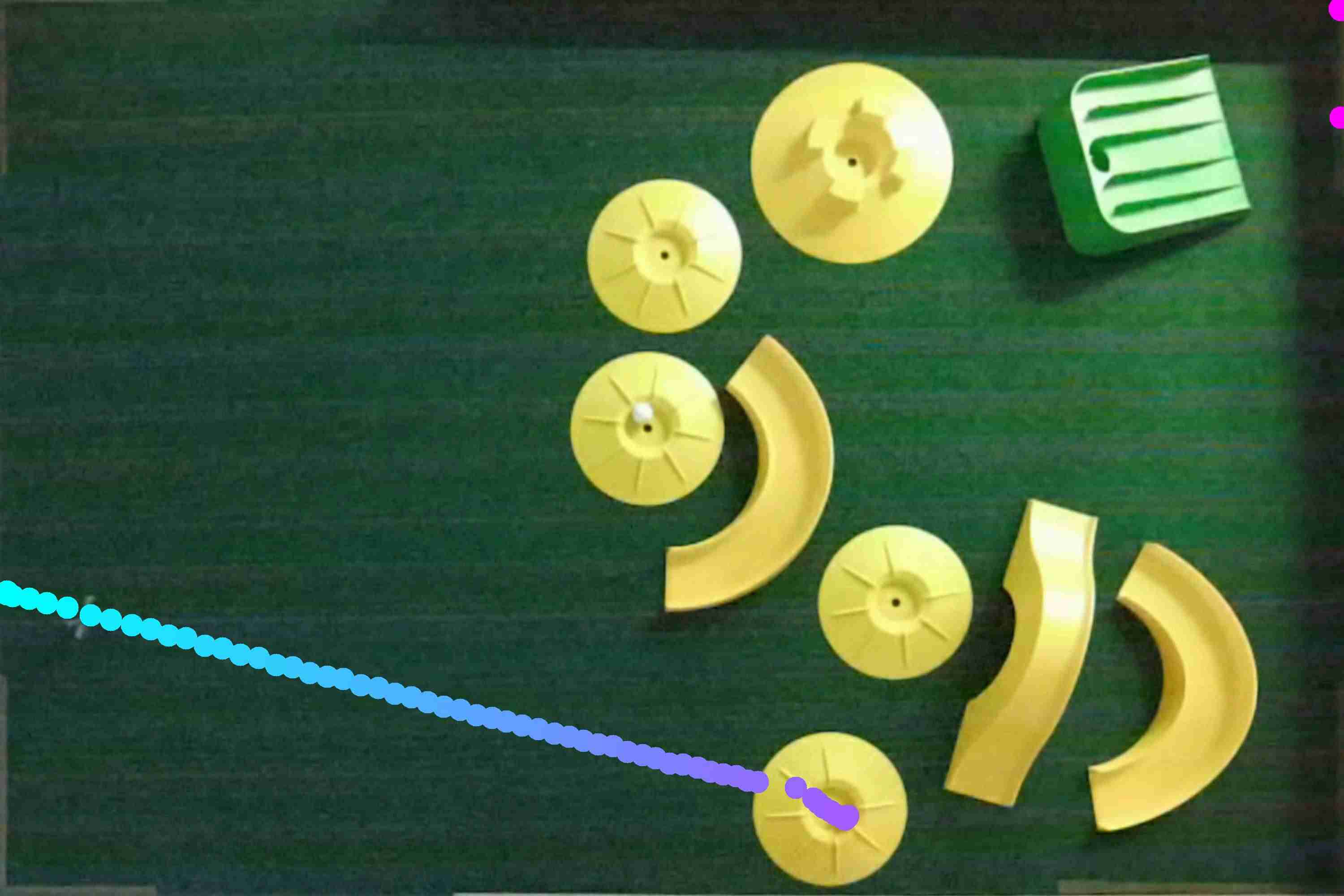} \\
        \includegraphics[width=0.24\textwidth]{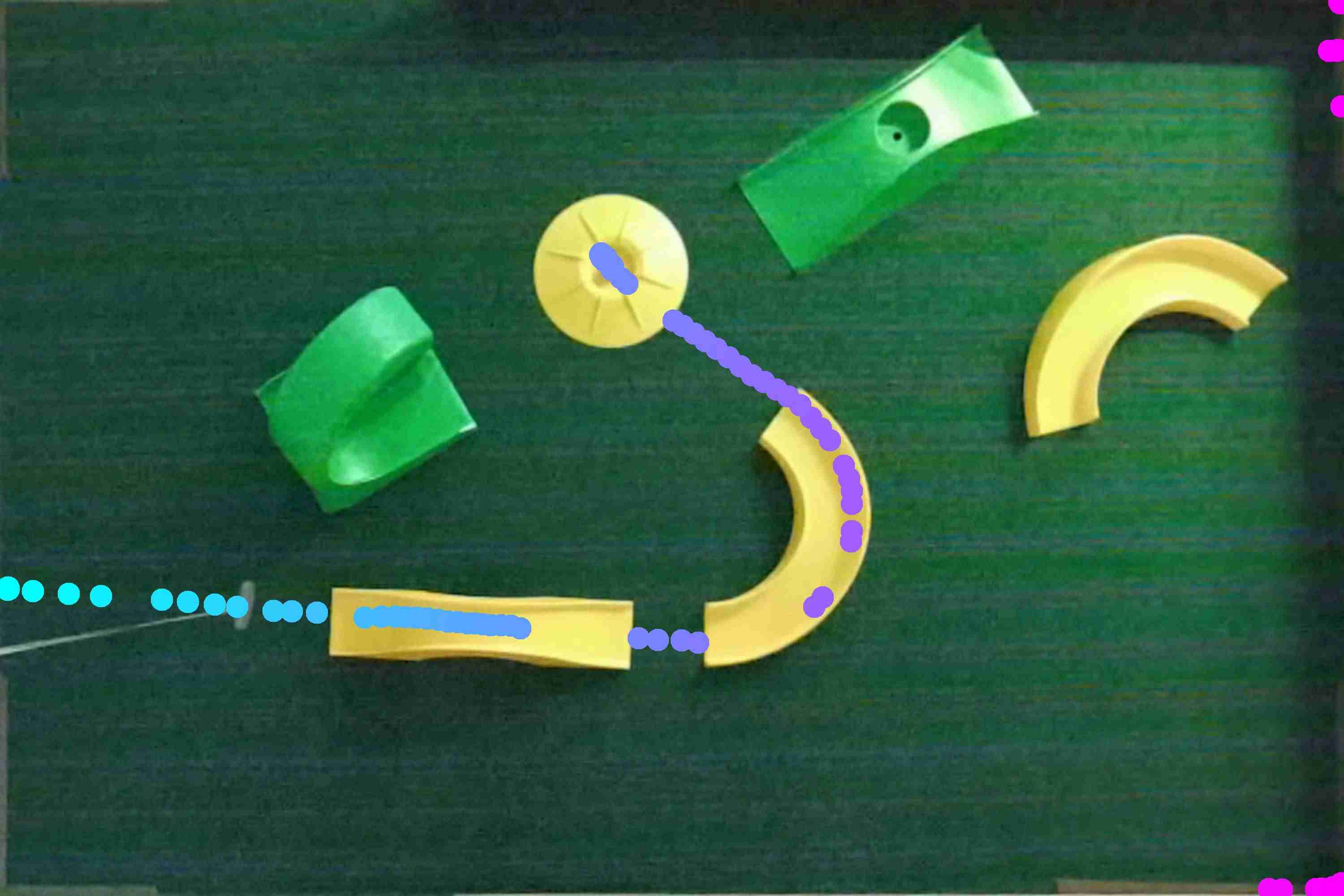} &
        \includegraphics[width=0.24\textwidth]{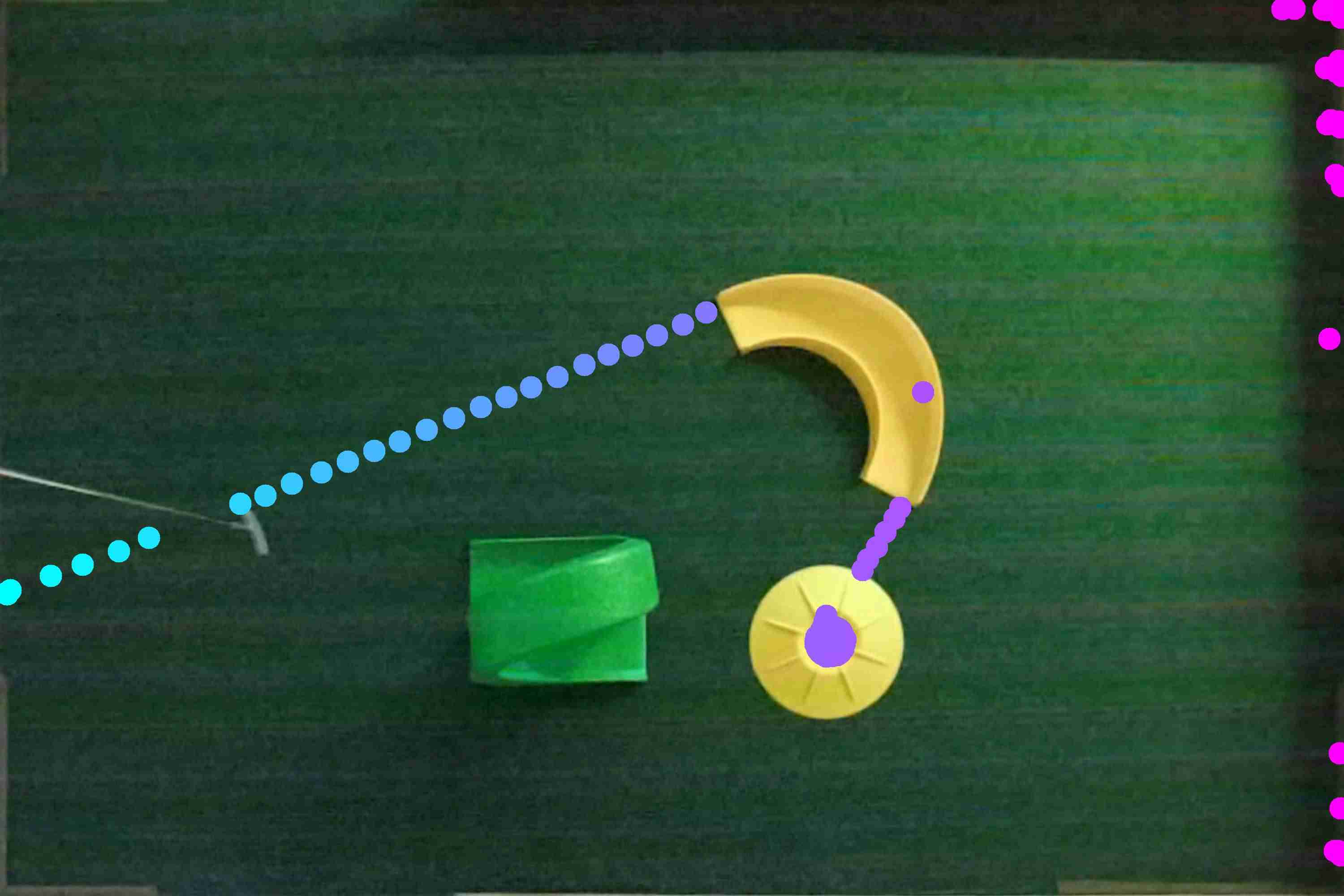} &
        \includegraphics[width=0.24\textwidth]{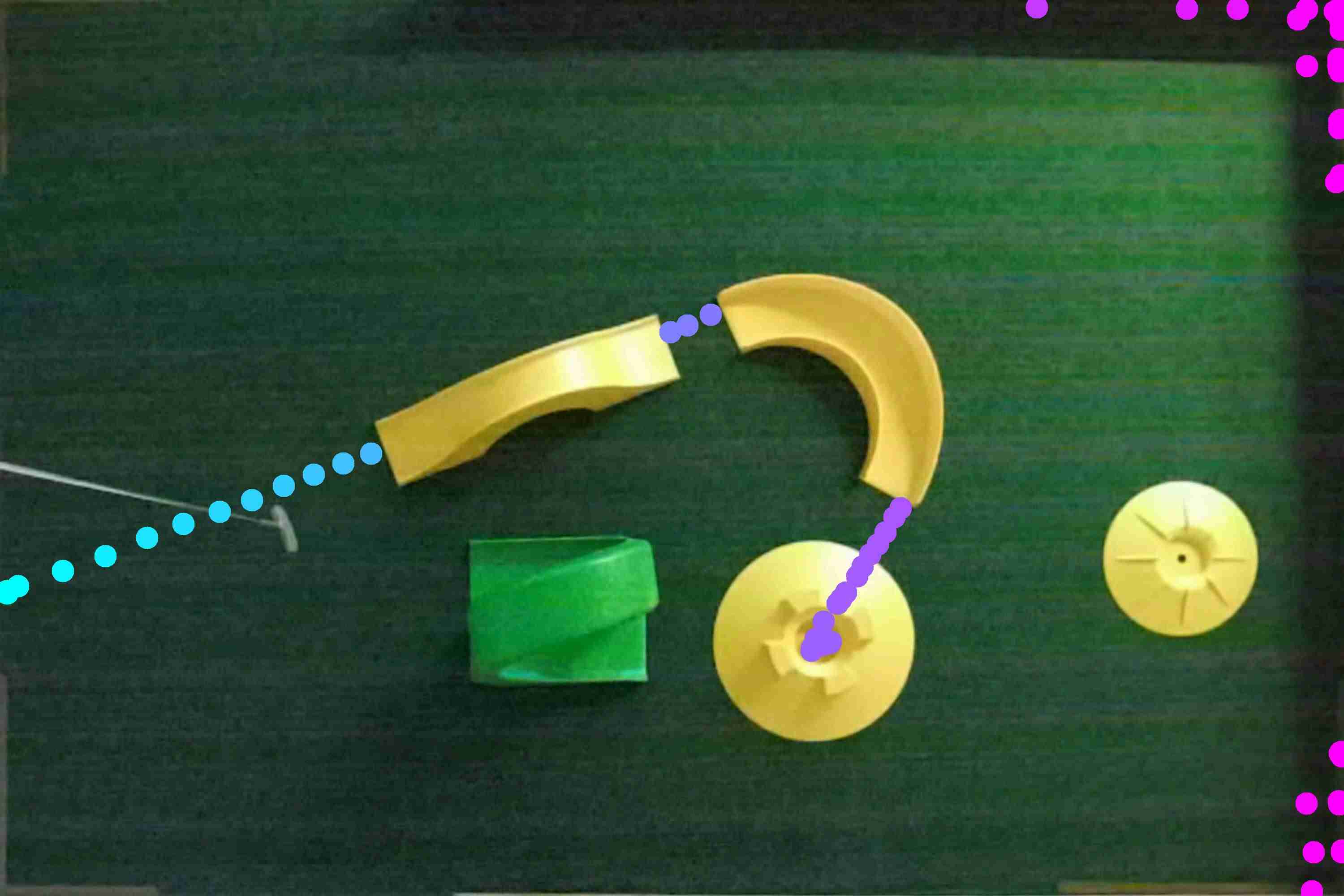} &
        \includegraphics[width=0.24\textwidth]{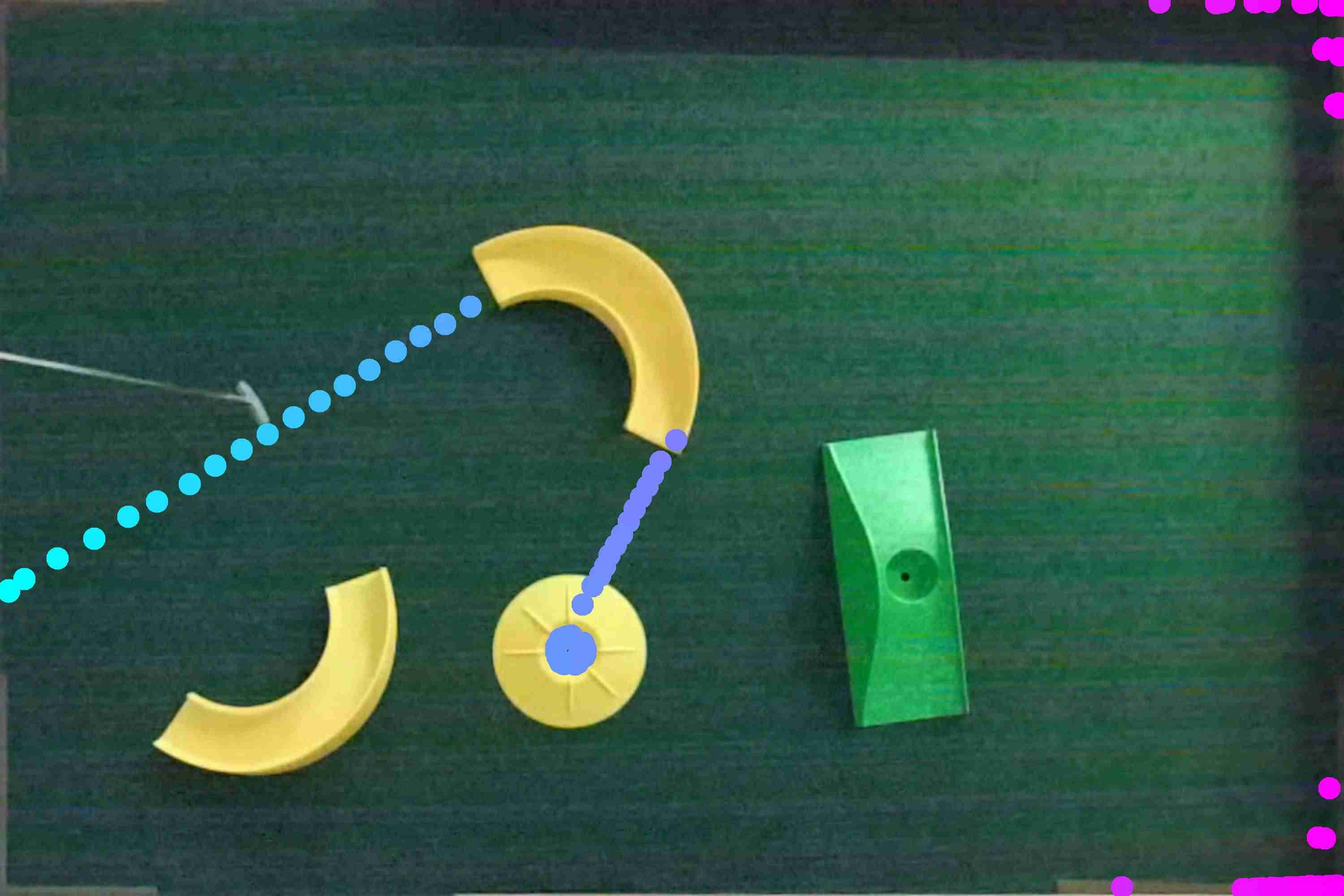} \\
        \includegraphics[width=0.24\textwidth]{Figures/hard/65_9d44ec42-bd9e-49fa-bee6-ee3d02675f03_transformed_rgb_trajectory_.jpg} &
        \includegraphics[width=0.24\textwidth]{Figures/hard/65_dbdc8c9c-ab44-41eb-ae72-d65c262126ba_transformed_rgb_trajectory_.jpg} &
        \includegraphics[width=0.24\textwidth]{Figures/hard/67_9a47b3fb-7d54-43f4-a2aa-629b1ab8c53f_transformed_rgb_trajectory_.jpg} &
        \includegraphics[width=0.24\textwidth]{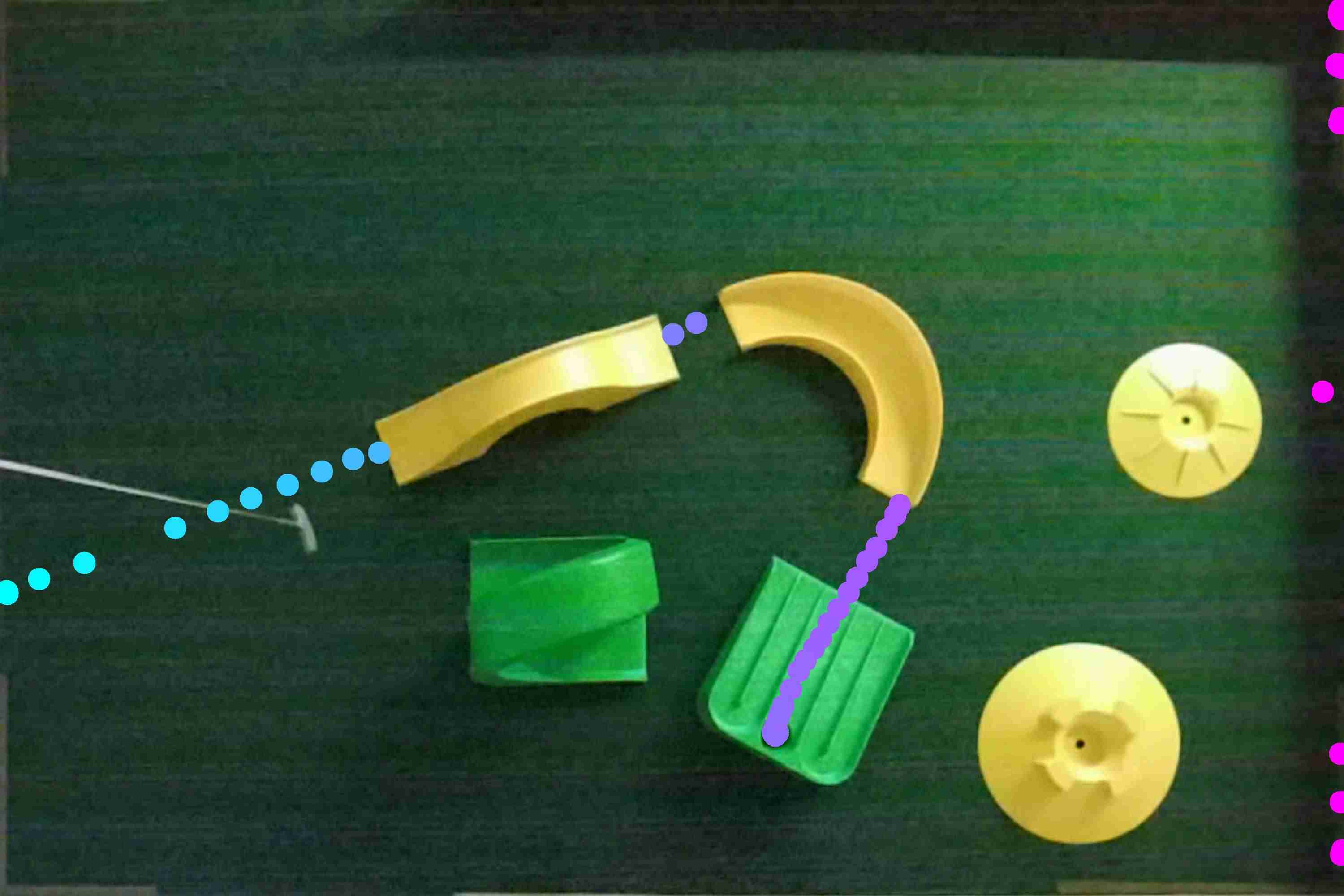} \\
        \includegraphics[width=0.24\textwidth]{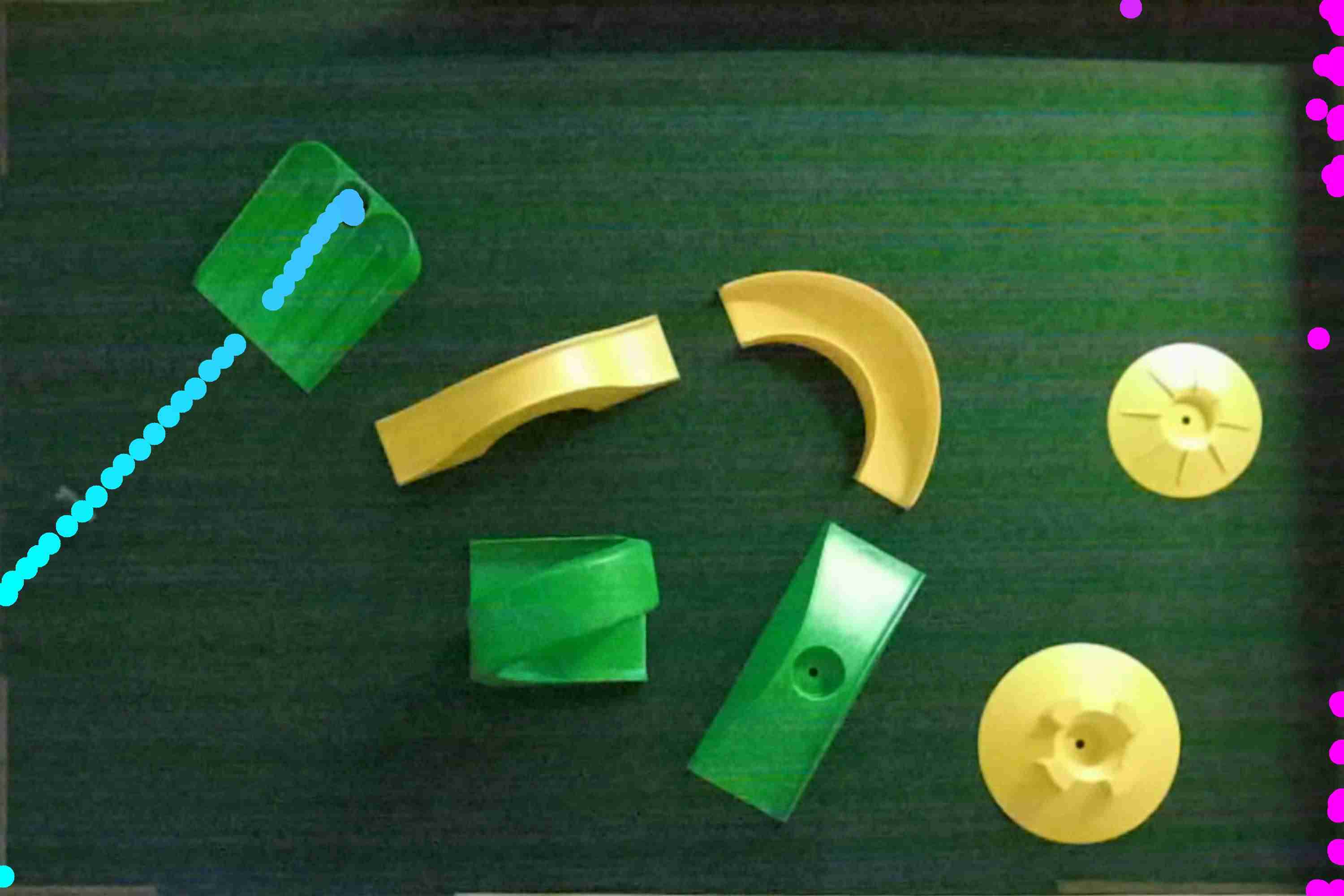} &
        \includegraphics[width=0.24\textwidth]{Figures/hard/95_fa29ba9d-afb9-4bd1-b8d0-f126d8b4593f_transformed_rgb_trajectory_.jpg} &
        \includegraphics[width=0.24\textwidth]{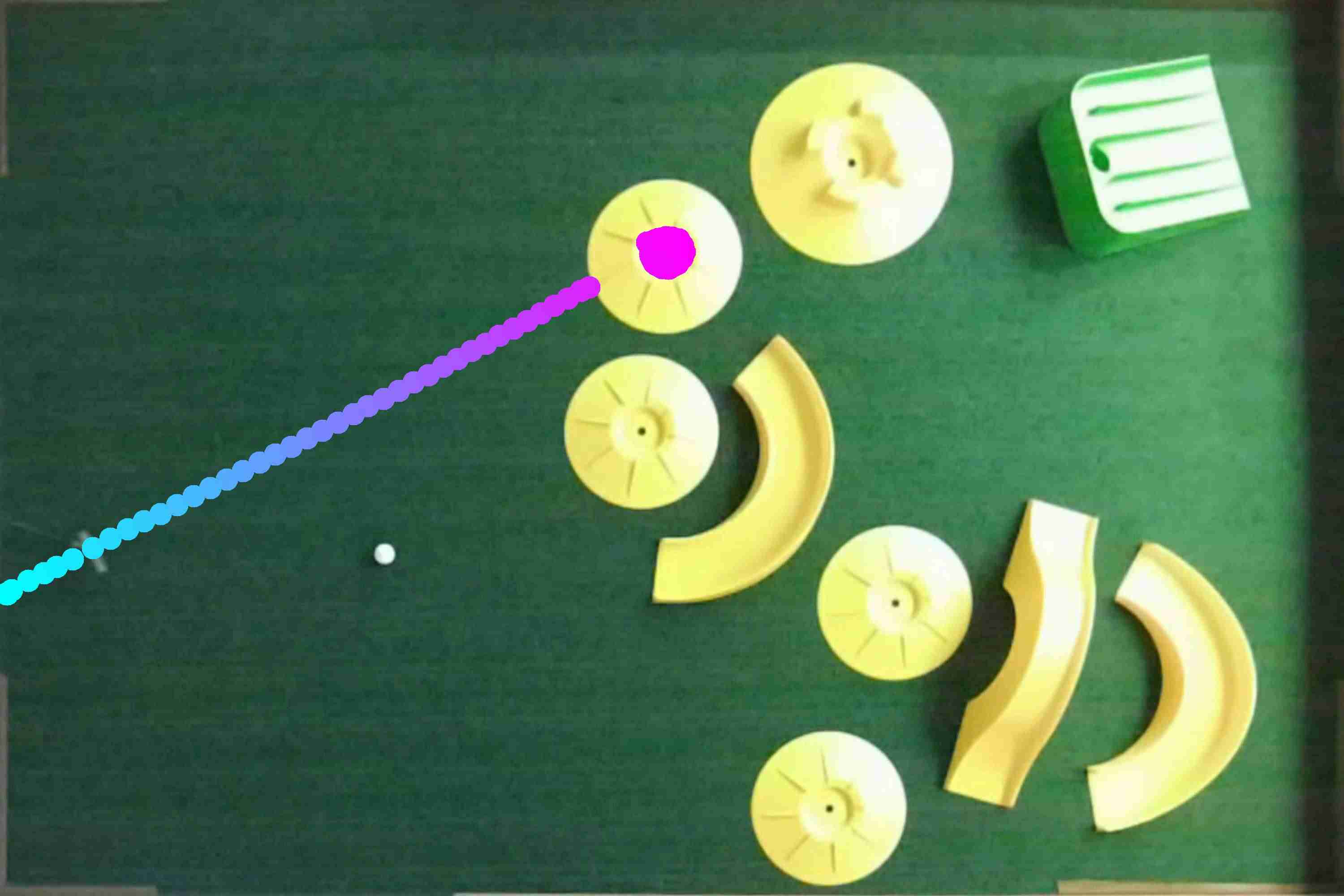} &
    \end{tabular}
    \caption{\textbf{Hitting trajectories in complex courts.}}
    \label{appfig:hardcourt}
\end{figure}

We conduct extensive experiments on RoboGolf across a variety of court complexities, ranging from simple to medium to complex. Figure~\ref{appfig:simplecourts}, Figure~\ref{appfig:mediumcourts} and Figure~\ref{appfig:hardcourt} illustrate the successful hitting trajectories observed.

\subsection{Experimental results in simple courts.}

\subsection{Experimental results in medium courts.}

\subsection{Experimental results in complex courts.}
\section{Ablation Studies}
We conduct comprehensive ablation studies to validate the effectiveness of the core designs of RoboGolf. Our analysis focuses on three key aspects: 1) the nested closed-loops design; 2) the fine-tuned VLMs on kinodynamic reasoning and counterfactual reasoning; and 3) the integration of topography information.

\subsection{Effects of nested closed-loops design.}
One of our key contributions, the nested closed-loops design, significantly enhances task completion in the real-world minigolf game. Both the outer and inner closed-loops are centered around providing feedback and making adaptations on top of the historical outcomes.

\paragraph{Inner action refinement.}
The goal of the inner loop is to iteratively refine the hitting actions until succeeding. 
Without this loop, the system lacks evaluation and feedback, preventing iterative improvements and information sharing between attempts.
Although we could repeatedly invoke the reasoning and planning components, the absence of a closed-loop means each trial operates independently, rendering each trial's outcome essentially an initial output without significant differences. In fact, even fine-tuned VLMs struggle to generate accurate hitting actions in a single shot, which could be observed in the examples in Table~\ref{apptab:typicalcourts}. 
Further, the noise from the unevenness of the grassland and the execution accuracy of the robot arm across each trial cannot be ignored. Even if the initial attempt succeeds, without sharing information between trials, subsequent attempts are likely to fail.

\paragraph{Outer closed-loop reflective equilibrium.}

Without the outer closed-loop, even equipped with the inner refinement loop, the robot is unable to solve those courts that are infeasible due to design errors. In such cases, the inner refinement loop will enter an infinite loop for impossible courts and cannot reflect and offer alteration suggestions on the current court.

\subsection{Finetuned VLM}
We replace the fine-tuned VLM with GPT-4V to demonstrate the superiority of our fine-tuned VLM. Using GPT-4V as the evaluating VLM often results in vague and unsubstantiated outcomes. 

\paragraph{Ablation on the kinodynamic reasoning VLM.}
To ensure a fair comparison between our fine-tuned VLM and GPT-4V, we provide them with identical inputs. We evaluate two aspects: 1) compare the initial reasoning and planning output based on the court information and assistant line; 2) compare the ability to recover from failures, given the historical hitting trajectories.

The kinodynamic reasoning is complex for minigolf, as even slight variations in court layout could induce significant deviations in hitting trajectories, hence posing challenges for a vanilla GPT-4V. As shown in Table~\ref{apptable:ablation-kino}, our fine-tuned kinodynamic reasoning VLM shines while GPT-4V fails to generate reasonable hitting actions or refinement suggestions.

\begin{table}[t]
    \centering
    \setlength{\tabcolsep}{2pt} 
    \renewcommand{\arraystretch}{3} 
    \begin{tabular}{|>{\centering\arraybackslash}m{0.2\textwidth}|>{\arraybackslash}m{0.38\textwidth}|>{\arraybackslash}m{0.38\textwidth}|}
        \hline
        \textbf{Input} &\textbf{Kinodynamic Reasoning VLM} & \textbf{GPT-4V} \\
        \hline
        {\small Assistant line}
        \includegraphics[width=0.8\linewidth]{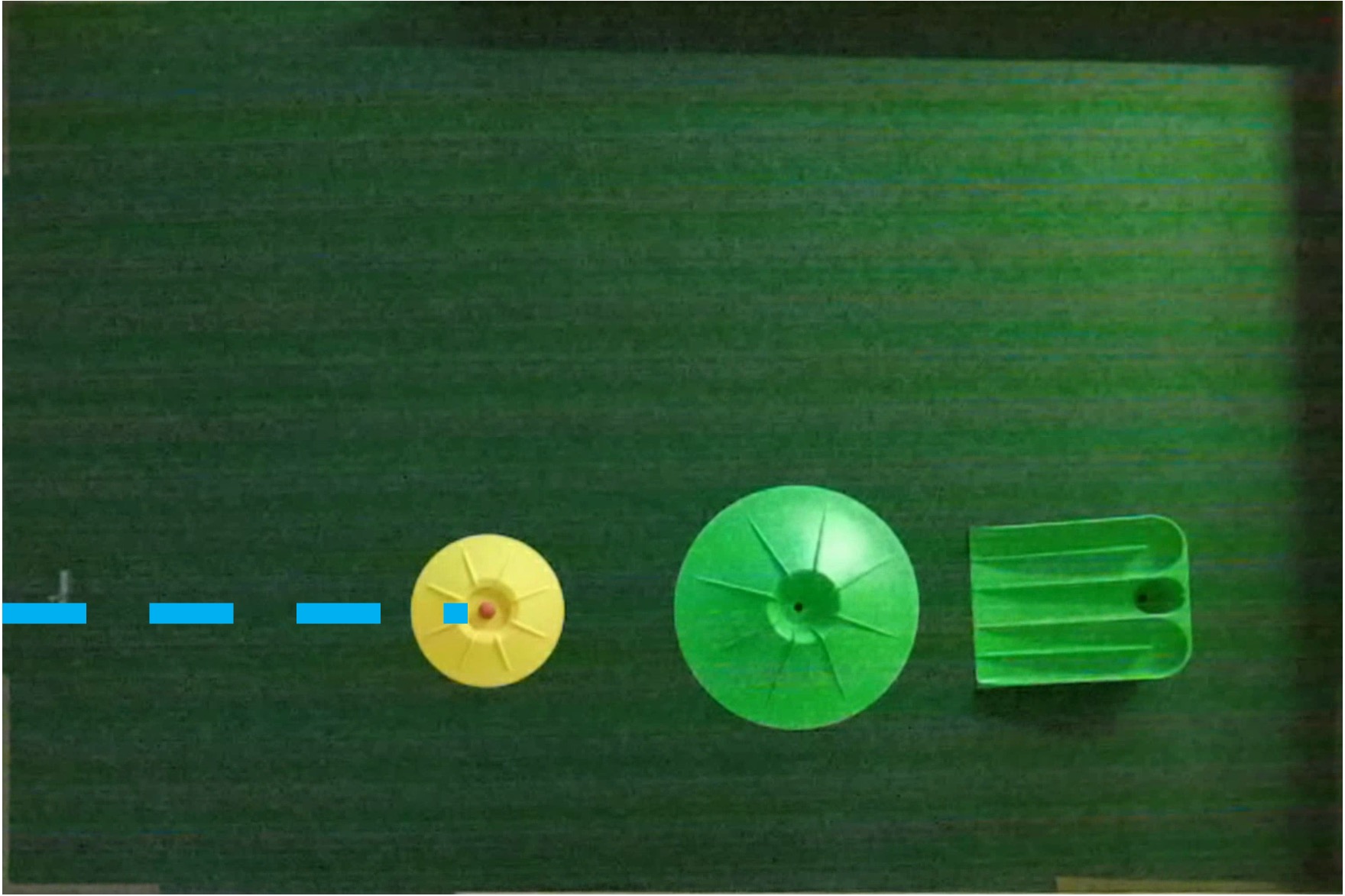}
        & 
         {\small Given the endpoint's distance and the possible obstacles on the route, \textcolor{mygreen}{the hitting force should be roughly 0.7. 
         Given the trajectory starting angle, the hitting angle should roughly be 2 degrees.}}
        & 
        {\small Glad to help! Considering the distance to the endpoint and the obstacles identified along the predicted route, \textcolor{myred}{the required hitting force is estimated at approximately 1.0. Given that the initial trajectory angle is nearly aligned with 0 degrees but slightly offset to the positive side}.}
        \\
        \hline
        {\small Historical trajectory}
        \includegraphics[width=0.8\linewidth]{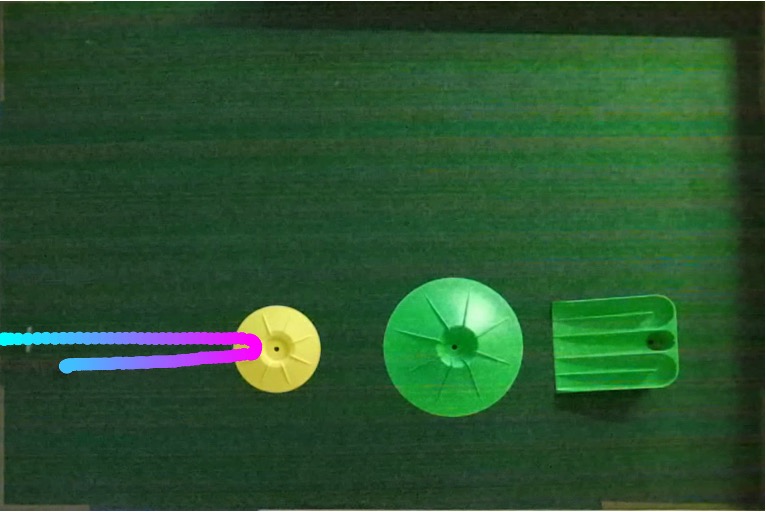}
        & 
         {\small Given the trajectory's evaluation, the golf ball reached the yellow plate, yet rolled backward, this is possibly due to \textcolor{mygreen}{insufficient hitting speed.}}
        & 
        {\small Based on the trajectory evaluation, the golf ball reached the yellow plate but did not stop at the endpoint. This could be attributed to several factors: \textcolor{myred}{1. the ball's path may have been altered by an obstacle; 2. the hitting parameters may not have been correctly set...}}
        \\
        \hline
    \end{tabular}
    \vspace{2mm}
    \caption{\textbf{Comparison on our kinodynamic reasoning VLM and GPT-4V.}}
    \label{apptable:ablation-kino}
\end{table}

\paragraph{Ablation on the Counterfactual Reasoning VLM}

The counterfactual reasoning VLM, fine-tuned on data tailored for reasoning and court modifications, exhibits a distinct advantage over the GPT-4V in providing accurate analysis of current failures, as shown in Table~\ref{apptable:ablation-counterfactual-part1} and Table~\ref{apptable:ablation-counterfactual-part2}. When presented with prompts, GPT-4V generates an abundance of suggestions, albeit nearly redundant, encompassing almost all possible solutions. In contrast, the significance of fine-tuning the VLM for counterfactual reasoning lies in its ability to provide exact suggestions based on specific circumstances, rather than vague and general recommendations.

Furthermore, the value of fine-tuning the VLM for this task is also evident in its capacity to discern intricate differences causing failures, as exemplified in the first case of Table~\ref{apptable:modifyingcourt}. The two concatenated curves, with small and subtle differences in relative positions, lead to substantial differences in outcomes. In contrast, GPT-4V struggles to identify these nuances, often attributing failures to the robot's limited power and speed.

\begin{table}[t]
    \centering
    \setlength{\tabcolsep}{2pt} 
    \renewcommand{\arraystretch}{3} 
    \begin{tabular}{|>{\centering\arraybackslash}m{0.2\textwidth}|>{\arraybackslash}m{0.38\textwidth}|>{\arraybackslash}m{0.38\textwidth}|}
        \hline
        \textbf{Input} &\textbf{Counterfactual Reasoning VLM} & \textbf{GPT-4V} \\
        \hline
        \raisebox{-0.25\height}{\includegraphics[width=0.85\linewidth]{Figures/comparison/5_23ce7b87-282f-49eb-b883-67b80f22d5af_transformed_rgb_trajectory_.jpg}}
        & 
        {\small To address the issue of the golf ball passing too quickly past the first obstacle with a pitfall in the middle and subsequently overshooting the second obstacle, here are some improved and detailed suggestions for modifications to the mini golf course:

Modify the First Obstacle: \textcolor{mygreen}{Consider replacing the first obstacle with one that requires less speed to overcome.} A ramp with a gentler incline or a series of smaller, less severe bumps could reduce the need for high initial speed, allowing for better control over the ball's momentum.

Adjust the Second Obstacle: \textcolor{mygreen}{Change the second obstacle to have a steeper incline or introduce a curved design that naturally slows down the ball as it ascends.} This change will help absorb some of the ball's kinetic energy, making it easier to stop at the desired endpoint.}
        & 
        {\small Modifications to the Court
Adjusting Surface Material: Using a surface material with higher friction could help in reducing the ball's momentum more effectively as it nears the endpoint.

Creating Dimples or Depressions: Introducing small depressions or dimples around the endpoint area can act as a natural barrier to stop the ball or slow it down significantly.

Strategic Obstacles: Placing additional obstacles strategically before the endpoint can help in reducing the ball's speed. These could be soft barriers that absorb some of the momentum without redirecting the ball too much.

Slope Adjustments: If the court allows for it, adjusting the slope so that it slightly rises towards the endpoint could help in naturally decelerating the ball.

Endpoint Design: Redesigning the endpoint to include a shallow cup or bowl-shaped area where the ball can settle would prevent it from moving past the target.

Hitting Parameters: \textcolor{myred}{Adjusting the recommended hitting parameters (angle and force) for players can also be an effective way to ensure the ball stops at the desired location.} Providing guidelines or marks on the hitting area could help players better judge their shots.}
        \\
        \hline
    \end{tabular}
    \vspace{2mm}
    \caption{\textbf{Comparison on counterfactual reasoning VLM and GPT-4V (Part 1).}}
    \label{apptable:ablation-counterfactual-part1}
\end{table}

\begin{table}[t]
    \centering
    \setlength{\tabcolsep}{2pt} 
    \renewcommand{\arraystretch}{3} 
    \begin{tabular}{|>{\centering\arraybackslash}m{0.2\textwidth}|>{\arraybackslash}m{0.38\textwidth}|>{\arraybackslash}m{0.38\textwidth}|}
        \hline
        \textbf{Input} &\textbf{Counterfactual Reasoning VLM} & \textbf{GPT-4V} \\
        \hline
        \raisebox{-0.25\height}{\includegraphics[width=0.85\linewidth]{Figures/comparison/61_d9739d4b-4918-446f-ab56-7f2e45b63da5_transformed_rgb_trajectory_.jpg}}
        & 
        {\small To further enhance the effectiveness of the course modifications and ensure consistent success in hitting the endpoint, here are additional suggestions that can be implemented:

\textcolor{mygreen}{Variety in Endpoint Designs}: Introduce different types of endpoints, such as a maze configuration or a tiered series of mini-bowls. These variations can add challenge and interest to the game, as well as provide natural stopping points for the golf ball. A maze design, for instance, would require precise navigation, slowing the ball as it makes its way to the center. Tiered bowls could catch the ball at different levels, depending on the initial force of the hit.

\textcolor{mygreen}{Incorporate Ramps into the Path}: Adding ramps at strategic points along the path can help in controlling the ball's speed. A ramp placed shortly before the endpoint could slow down the ball as it climbs the incline, reducing its momentum before it reaches the goal. Conversely, a downward ramp could be used to correctly align the ball towards a strategically challenging endpoint, such as one positioned in a depression or at a lower elevation.

Adjustable Obstacle Positions: Implement modular obstacles that can be moved or adjusted in position and difficulty. This flexibility allows for the course to be customized based on player skill level or for different gameplay experiences. It can also help in fine-tuning the course based on observed player interactions and ball trajectories.}
        & 
        {\small Suggestions for Court Modifications
Modify the Crescent Obstacle: Adjusting the angle or position of the crescent-shaped obstacle could help in better directing the ball towards the center of the endpoint. It could also be designed to decelerate the ball as it redirects it.

Enhance the Endpoint: Incorporating a slight depression or a textured surface within the endpoint can help to physically stop the ball. Another option is to make the endpoint slightly concave, so the ball naturally settles towards the center rather than rolling off.

\textcolor{myred}{Surface Texture Adjustments: Adding rougher textures or patches before the endpoint could help decrease the ball's speed more effectively, ensuring it stops upon reaching the target.}

Strategic Additional Obstacles: Placing additional small obstacles or buffers before the endpoint can slow down the ball. These should be designed so they don’t completely block the ball but rather reduce its speed.

Visual Aids for Players: Providing visual cues on the playing surface to guide players on where to aim and how hard to hit can improve accuracy and control over the ball’s speed and direction.}
        \\
        \hline
    \end{tabular}
    \vspace{2mm}
    \caption{\textbf{Comparison on counterfactual reasoning VLM and GPT-4V (Part 2).}}
    \label{apptable:ablation-counterfactual-part2}
\end{table}

\subsection{Topography Information}

The incorporation of topography information is crucial for providing the VLM with auxiliary data. This includes object segmentation masks, feature matching after semantic segmentation, and assistive lines to predict the optimal route. Without such information, relying solely on RGB images with shapes and colors, GPT-4V lacks the necessary context to construct accurate routes. This results in noisy hitting parameters.

In simple scenarios (refer to Figure~\ref{fig:easyscenarios}), the system may still provide adequate predictions for the possible route. This yields roughly correct hitting parameters and guides the direction towards the correct endpoint. The inner-closed loop's feedback and fine-tuned VLM can iteratively refine the hitting parameters to their correct values.

However, in medium cases (refer to Figure~\ref{fig:medium-scenarios}) and complex cases (refer to Figure~\ref{fig:hard-scenario}), the detection of the correct route becomes increasingly challenging for the framework without topography information assistance. This observation is consistent with the ablation results reported in MOKA~\citep{liu2024moka}. Without explanatory information, GPT-4V fails to establish correct connections between key points, thereby forming dynamic trajectories.

\end{document}